\documentclass[twoside,11pt,colorlinks,linktoc=all]{article}

\usepackage[nohyperref,abbrvbib]{jmlr2e}
\usepackage{lastpage}
\usepackage[dvipsnames,table,]{xcolor}  \usepackage[colorlinks,linktoc=all]{hyperref}  \usepackage[all]{hypcap}
\hypersetup{citecolor=MidnightBlue}
\hypersetup{linkcolor=MidnightBlue}
\hypersetup{urlcolor=MidnightBlue}
\hypersetup{ hidelinks }

\usepackage{scalefnt,letltxmacro}
\LetLtxMacro{\oldtextsc}{\textsc}
\renewcommand{\textsc}[1]{\oldtextsc{\scalefont{1.10}#1}}
\usepackage[acronym,smallcaps,nowarn]{glossaries}
\glsdisablehyper
\makeglossaries
\usepackage{xspace}
\usepackage{float}
\usepackage{physics}

\usepackage{amssymb}
\usepackage{mathtools}
\usepackage{amsfonts}
\usepackage{amsmath}
\usepackage{booktabs}
\usepackage[ruled,vlined]{algorithm2e}  \usepackage[]{microtype}
\usepackage{multirow}

\usepackage{enumitem}

\usepackage[capitalize,nameinlink,noabbrev]{cleveref}
\crefname{section}{\S}{\S\S}
\Crefname{section}{\S}{\S\S}

\usepackage[font=small,labelfont=bf,tableposition=top]{caption}
\usepackage{tikz}
\usepackage{graphicx}
\usepackage[]{subcaption}   
\newcommand{\rulesep}{\unskip\ \hrule\ }

\usepackage{pgfplots}
\pgfplotsset{compat=1.6}
\usepgfplotslibrary{groupplots}
\tikzstyle{every picture}+=[font=\sffamily]
\tikzstyle{optimized} = [circle,fill=white,draw=black, dashed,inner sep=1pt, minimum size=20pt, font=\fontsize{10}{10}\selectfont, node distance=1]
\pgfkeys{/pgf/number format/.cd,1000 sep={}}
\pgfplotsset{
	tick label style = {font=\sffamily},
	every axis label/.append style={font=\sffamily},
	typeset ticklabels with strut,
}
\pgfplotsset{every axis/.append style={
			every x tick label/.append style={font=\fontsize{6pt}{6pt}\sffamily, yshift=.5ex,},
			every y tick label/.append style={font=\fontsize{6pt}{6pt}\sffamily, xshift=.5ex},
			every y label/.append style={xshift=10ex, font=\sffamily},
			every x label/.append style={yshift=3ex, font=\sffamily},
			every title/.append style={font=\sffamily},
			},
}
\pgfplotsset{
	xticklabel={$\mathsf{\pgfmathprintnumber{\tick}}$},
	yticklabel={$\mathsf{\pgfmathprintnumber{\tick}}$},
}
\pgfplotsset{every axis title/.append style={yshift=-1ex}}
\newlength\figureheight
\newlength\figurewidth

\usepackage[colorinlistoftodos,
	textsize=scriptsize,
	linecolor=red!30,
	bordercolor=red!30,
	backgroundcolor=red!10]{todonotes}
\renewcommand{\todo}[2][]{\@todo[#1]{#2}}

\usepackage[most]{tcolorbox}

\tcbset{boxsep=4pt,left=2pt,right=2pt,top=-0pt,bottom=0pt}

\usepackage{url}

\usepackage{xr}
\newcommand{\redquestionmark}{\textbf{\textcolor{red!60!black}{?}}}

\newcommand{\mathbold}[1]{\ensuremath{\boldsymbol{\mathbf{#1}}}}

\newcommand{\g}{\,|\,}
\renewcommand{\d}[1]{\ensuremath{\operatorname{d}\!{#1}}}
\newcommand{\nestedmathbold}[1]{{\mathbold{#1}}}

\newcommand{\mbf}{\nestedmathbold{f}}

\newcommand{\mbr}{\nestedmathbold{r}}

\newcommand{\mbu}{\nestedmathbold{u}}
\newcommand{\mbv}{\nestedmathbold{v}}
\newcommand{\mbw}{\nestedmathbold{w}}
\newcommand{\mbx}{\nestedmathbold{x}}
\newcommand{\mby}{\nestedmathbold{y}}

\newcommand{\mbB}{\nestedmathbold{B}}
\newcommand{\mbC}{\nestedmathbold{C}}

\newcommand{\mbI}{\nestedmathbold{I}}

\newcommand{\mbK}{\nestedmathbold{K}}

\newcommand{\mbM}{\nestedmathbold{M}}

\newcommand{\mbX}{\nestedmathbold{X}}

\newcommand{\mbmu}{\nestedmathbold{\mu}}

\newcommand{\mbpsi}{\nestedmathbold{\psi}}

\newcommand{\mbsigma}{\nestedmathbold{\sigma}}

\newcommand{\mbtheta}{\nestedmathbold{\theta}}

\newcommand{\mbSigma}{\nestedmathbold{\Sigma}}

\DeclareRobustCommand{\KL}[2]{\ensuremath{\textsc{kl}\left[#1\;\|\;#2\right]}}
\DeclarePairedDelimiterX{\infdivx}[2]{[}{]}{#1\;\delimsize\|\;#2}

\DeclareMathOperator*{\argmax}{arg\,max}
\DeclareMathOperator*{\argmin}{arg\,min}

\newcommand{\cD}{\mathcal{D}}
\newcommand{\cL}{\mathcal{L}}
\newcommand{\cN}{\mathcal{N}}

\newcommand{\cF}{\mathcal{F}}

\newcommand{\cT}{\mathcal{T}}

\newcommand{\cX}{\mathcal{X}}
\newcommand{\cY}{\mathcal{Y}}

\newcommand{\E}{\mathbb{E}}

\newcommand{\bbR}{\mathbb{R}}

\newcommand{\defeq}{\stackrel{\text{\tiny def}}{=}}

\newcommand{\sub}[1]{{\texttt{\textit{\scriptsize {#1}}}}}
\newacronym{MAP}{map}{maximum-a-posteriori}
\newacronym{MLE}{mle}{maximum likelihood estimation}
\newacronym{MNLL}{mnll}{mean negative loglikelihood}
\newacronym{NLL}{nll}{negative loglikelihood}
\newacronym{RMSE}{rmse}{root mean square error}
\newacronym{ECE}{ece}{expected calibration error}

\newacronym{VAE}{vae}{variational autoencoder}

\newacronym{MC}{mc}{Monte Carlo}
\newacronym{MCMC}{mcmc}{Markov chain Monte Carlo}
\newacronym{HMC}{hmc}{Hamiltonian Monte Carlo}
\newacronym{MH}{mh}{Metropolis-Hastings}
\newacronym{NUTS}{nuts}{no-u-turn}
\newacronym{SGHMC}{sghmc}{stochastic gradient Hamiltonian Monte Carlo}
\newacronym{GGN}{ggn}{generalized Gauss-Newton}

\newacronym{SSGE}{ssge}{spectral Stein gradient estimator}
\newacronym{MMD}{mmd}{maximum mean discrepancy}

\newacronym{DGP}{dgp}{deep Gaussian process} \newacronym{GPLVM}{gplvm}{Gaussian process latent variable model}

\newacronym{VFE}{vfe}{variational free energy}

\newacronym[firstplural=Gaussian Processes]{GP}{gp}{Gaussian Process}

\newacronym{VI}{vi}{variational inference}
\newacronym{LA}{la}{Laplace approximation}

\newacronym{ELBO}{elbo}{evidence lower bound}
\newacronym{NELBO}{nelbo}{negative evidence lower bound}
\newacronym{ELL}{ell}{expected log likelihood}
\newacronym{KL}{kl}{Kullback-Leibler}
\newacronym{AUC}{auc}{area under the curve}
\newacronym{CV}{cv}{cross-validation}

\newacronym[firstplural=Bayesian neural networks]{BNN}{bnn}{Bayesian neural network}
\newacronym[firstplural=deep neural networks]{DNN}{dnn}{deep neural network}
\newacronym[]{CNN}{cnn}{convolutional neural network}
\newacronym{MLP}{mlp}{multilayer perceptron}
\newacronym{NN}{nn}{neural network}
\newacronym{RELU}{ReLU}{rectified linear unit}

\newacronym{NF}{nf}{normalizing flow}

\newacronym{RBF}{rbf}{radial basis function}
\newacronym{ARD}{ard}{automatic relevance determination}

\newacronym{RKHS}{rkhs}{reproducing kernel Hilbert space}

\newacronym{NKN}{nkn}{Neural Kernel Network}

\newcommand{\fbnn}{f\textsc{bnn}\xspace}
\newcommand{\lamarglik}{\textsc{la}-\textsc{m}arg\textsc{l}ik\xspace}
\newcommand{\laplace}{\textsc{l}aplace\xspace}

\newcommand{\lagnn}{\textsc{la}-\textsc{ggn}\xspace}
\newcommand{\sghmc}{\textsc{sghmc}\xspace}
\newcommand{\ggn}{\textsc{ggn}\xspace}

\newcommand{\fg}{\textsc{fg}\xspace}
\newcommand{\fh}{\textsc{fh}\xspace}
\newcommand{\fgts}{\textsc{fg+ts}\xspace}
\newcommand{\gpig}{\textsc{gp}i-\textsc{g}\xspace}
\newcommand{\gpih}{\textsc{gp}i-\textsc{h}\xspace}
\newcommand{\gpinf}{\textsc{gp}i-\textsc{nf}\xspace}

\newcommand{\name}[1]{{\textsc{#1}}\xspace}

\newcommand{\lenet}{\name{lenet5}}
\newcommand{\vgg}{\name{vgg16}}
\newcommand{\preresnet}{\name{preresnet20}}

\newcommand{\cifar}{\name{cifar10}}
\newcommand{\cifarc}{\name{cifar10c}}
\newcommand{\mnist}{\name{mnist}}
\newcommand{\notmnist}{\name{not-mnist}}
\newcommand{\uci}{\name{uci}}

\newcommand{\boston}{\name{boston}}
\newcommand{\concrete}{\name{concrete}}
\newcommand{\energy}{\name{energy}}
\newcommand{\kinm}{\name{kin8nm}}
\newcommand{\naval}{\name{naval}}
\newcommand{\power}{\name{power}}
\newcommand{\protein}{\name{protein}}
\newcommand{\wine}{\name{wine}}

\newcommand{\eeg}{\name{eeg}}
\newcommand{\htru}{\name{htru2}}
\newcommand{\magic}{\name{magic}}
\newcommand{\miniboo}{\name{miniboo}}
\newcommand{\letter}{\name{letter}}
\newcommand{\drive}{\name{drive}}
\newcommand{\mocap}{\name{mocap}}

\newcommand{\banana}{\name{banana}}

\jmlrheading{23}{2022}{1-\pageref{LastPage}}{11/20; Revised 10/21}{3/22}{20-1340}{Ba-Hien Tran, Simone Rossi, Dimitrios Milios, Maurizio Filippone}

\ShortHeadings{All You Need is a Good Functional Prior for Bayesian Deep Learning}{Tran, Rossi, Milios, and Filippone}
\firstpageno{1}

\title{All You Need is a Good Functional Prior\\for Bayesian Deep Learning}

\author{\name Ba-Hien Tran \email ba-hien.tran@eurecom.fr \\[.5ex]
	\name Simone Rossi \email simone.rossi@eurecom.fr \\[.5ex]
	\name Dimitrios Milios \email dimitrios.milios@eurecom.fr \\[.5ex]
	\name Maurizio Filippone \email maurizio.filippone@eurecom.fr \\
	\addr Data Science Department\\
	EURECOM\\
	Sophia Antipolis, FR
}

\editor{Mohammad Emtiyaz Khan}

\begin{document}

\maketitle

\begin{abstract}
The Bayesian treatment of neural networks dictates that a prior distribution is specified over their weight and bias parameters.
This poses a challenge because modern neural networks are characterized by a large number of parameters, and the choice of these priors has an uncontrolled effect on the induced functional prior, which is the distribution of the functions obtained by sampling the parameters from their prior distribution.
We argue that this is a hugely limiting aspect of Bayesian deep learning, and this work tackles this limitation in a practical and effective way.
Our proposal is to reason in terms of functional priors, which are easier to elicit, and to ``tune'' the priors of neural network parameters in a way that they reflect such functional priors.
Gaussian processes offer a rigorous framework to define prior distributions over functions, and we propose a novel and robust framework to match their prior with the functional prior of neural networks based on the minimization of their Wasserstein distance.
We provide vast experimental evidence that coupling these priors with scalable Markov chain Monte Carlo sampling offers systematically large performance improvements over alternative choices of priors and state-of-the-art approximate Bayesian deep learning approaches.
We consider this work a considerable step in the direction of making the long-standing challenge of carrying out a fully Bayesian treatment of neural networks, including convolutional neural networks, a concrete possibility. 

\end{abstract}


\section{Introduction} \label{sec:introduction}

The majority of tasks in machine learning, including classical ones such as classification and regression, can be reduced to estimation of functional representations, and neural networks offer a powerful framework to describe functions of high complexity.
In this work, we focus on the Bayesian treatment of neural networks, which results in a natural form of regularization and allows one to reason about uncertainty in predictions \citep{Tishby1989,neal1996bayesian,MacKay03}.
Despite the lack of conjugate priors for any \glspl{BNN} of interest, it is possible to generate samples from the posterior distributions over their parameters by means of Markov chain Monte Carlo algorithms \citep{neal1996bayesian,Cheni2014}.

The concept of prior distribution in Bayesian inference allows us to describe the family of solutions that we consider acceptable, \textit{before} having seen any data.
While in some cases selecting an appropriate prior is easy or intuitive given the context \citep{Hagan1991, Ghahramani2003,Srinivas2010,Cockayne2019, Briol2019, Tran2021}, for nonlinear parametric models with thousands (or millions) of parameters, like \glspl{DNN} and \glspl{CNN}, this choice is not straightforward.
As these models are nowadays accepted as the \textit{de facto standard} in machine learning \citep{LeCun2015}, the community has been actively proposing ways to enable the possibility to reason about the uncertainty in their predictions, with the Bayesian machinery being at the core of many contributions \citep{Graves2011,Cheni2014, Gal2016,Liu2016}.

Despite many advances in the field \citep{Kendall2017,Rossi2018,Osawa2019,Rossi2020}, it is reported that in some cases the predictive posteriors are not competitive to non-Bayesian alternatives, making these models---and Bayesian deep learning, in general---less than ideal solutions for a number of applications.
For example, \cite{Wenzel2020} have raised concerns about the quality of \gls{BNN} posteriors, where it is found that tempering the posterior distribution improves the performance of some deep models.
We argue that observations of this kind should not be really surprising.
Bayesian inference is a recipe with exactly three ingredients: the {\it prior distribution}, the {\it likelihood} and the {\it Bayes' rule}.
Regarding the Bayes' rule, that is simply a consequence of the axioms of probability.
The fact that the posterior might not be useful in some cases should never be attributed to the Bayesian method itself. In fact, it is very easy to construct Bayesian models with poor priors and/or likelihoods, which result in poor predictive posteriors.
One should therefore turn to the other two components, which encode model assumptions. 
In this work, we focus our discussion and analysis on the prior distribution of \glspl{BNN}.
For such models, the common practice is to define a prior distribution on the network weights and biases, which is often chosen to be Gaussian.
A prior over the parameters induces a prior on the functions generated by the model, which also depends on the network architecture.
However, due to the nonlinear nature of the model, the effect of this prior on the functional output is not obvious to characterize and control.

\begin{figure}[]
	\centering
	\begin{subfigure}{0.33\textwidth}
		\scriptsize\includegraphics{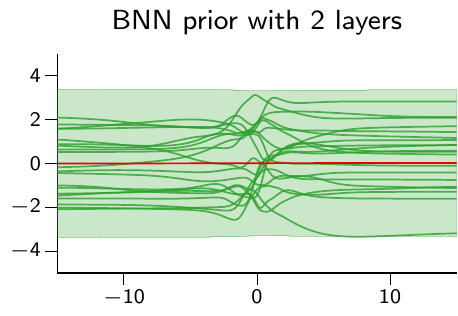}
	\end{subfigure}\begin{subfigure}{0.33\textwidth}
		\scriptsize\includegraphics{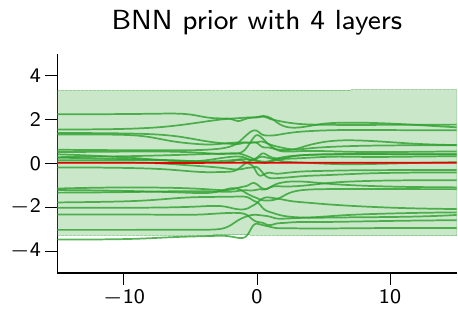}
	\end{subfigure}\begin{subfigure}{0.33\textwidth}
		\scriptsize\includegraphics{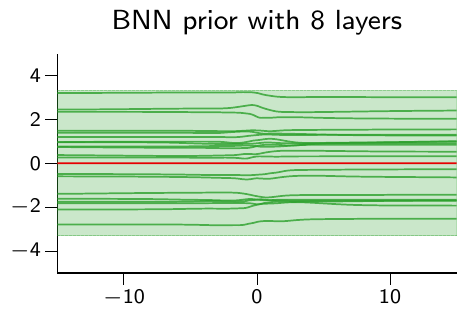}
	\end{subfigure}\\
	\vskip 0.05in
	\hrule\vskip 0.05in
	\centering
	\begin{subfigure}{0.33\textwidth}
		\scriptsize \includegraphics{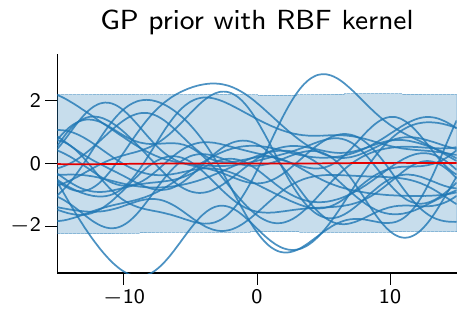}
	\end{subfigure}\begin{subfigure}{0.33\textwidth}
		\scriptsize \includegraphics{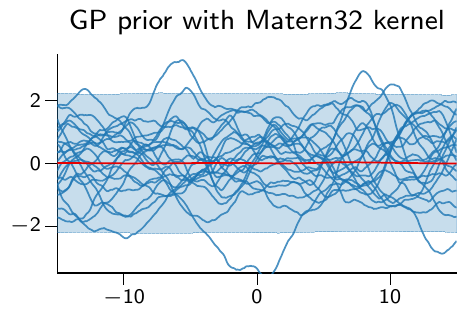}
	\end{subfigure}
	\caption{{(\textbf{Top})} Sample functions of a fully-connected \gls{BNN} with 2, 4 and 8 layers obtained by placing a Gaussian prior on the weights.
			{(\textbf{Bottom})} Samples from a \gls{GP} prior with two different kernels.}\label{fig:prior_example}
\end{figure}

Consider the example in \cref{fig:prior_example}, where we show the functions generated by sampling the weights of \glspl{BNN} with a $\mathrm{tanh}$ activation from their Gaussian prior $\cN(0, 1)$.
We see that as depth is increased, the samples tend to form straight horizontal lines, which is a well-known pathology stemming from increasing model's depth \citep{neal1996bayesian,Duvenaud14,Matthews2018}. We stress that a fixed Gaussian prior on the parameters is not always problematic, but it can be, especially for deeper architectures.
Nonetheless, this kind of generative priors on the functions is very different from shallow Bayesian models, such as \glspl{GP}, where the selection of an appropriate prior typically reflects certain attributes that we expect from the generated functions.
A \gls{GP} defines a distribution over functions which is characterized by a mean and a kernel function $\kappa$.
The \gls{GP} prior specification can be more \emph{interpretable} than the one induced by the prior over the weights of a \gls{BNN}, in the sense that the kernel effectively governs the properties of prior functions, such as shape, variability and smoothness.
For example, shift-invariant kernels may impose a certain characteristic length-scale on the functions that can be drawn from the prior distribution.

\pgfplotsset{every y tick label/.append style={font=\fontsize{1}{1}\selectfont}}
\pgfplotsset{every x tick label/.append style={font=\fontsize{1}{1}\selectfont}}
\setlength\figureheight{.25\textwidth}
\setlength\figurewidth{.37\textwidth}

\subsection*{Contributions}
The main research question that we investigate in this work is how to impose functional priors on \glspl{BNN}.
We seek to tune the prior distributions over \glspl{BNN} parameters so that the induced functional priors exhibit interpretable properties, similar to shallow \glspl{GP}.
While \gls{BNN} priors induce a regularization effect that penalizes large values for the network weights, a \gls{GP}-adjusted prior induces regularization directly on the space of functions.

We consider the \emph{Wasserstein distance} between the distribution of \gls{BNN} functions induced by a prior over their parameters, and a target \gls{GP} prior.
We propose an algorithm that optimizes such a distance with respect to the \gls{BNN} prior parameters and hyper-parameters.
An attractive property of our proposal is that estimating the Wasserstein distance relies exclusively on samples from both distributions, which are easy to generate.
We demonstrate empirically that for a wide range of \gls{BNN} architectures with smooth activations, it is possible to sufficiently capture the function distribution induced by popular \gls{GP} kernels.

We then explore the effect of \gls{GP}-induced priors on the predictive posterior distribution of \glspl{BNN} by means of an extensive experimental campaign.
We do this by carrying out fully Bayesian inference of neural network models with these priors through the use of scalable \gls{MCMC} sampling \citep{Cheni2014}.
We demonstrate systematic performance improvements over alternative choices of priors and state-of-the-art approximate Bayesian deep learning approaches on a wide range of regression and classification problems, as well as a wide range of network architectures including convolutional neural networks; we consider this a significant advancement in Bayesian deep learning.

\section{Related Work} \label{sec:related-work}

In the field of \glspl{BNN}, it is common practice to consider a diagonal Gaussian prior distribution for the network weights \citep{neal1996bayesian, Bishop06}.
Certain issues of these kind of \gls{BNN} priors have been recently exposed by \cite{Wenzel2020}, who show that standard Gaussian priors exhibit poor performance, especially in the case of deep architectures.
The authors address this issue by considering a temperate version of the posterior, which effectively reduces the strength of the regularization induced by the prior.
Many recent works \citep{Cheni2014,Springenberg2016} consider a hierarchical structure for the prior, where the variance of the normally-distributed \gls{BNN} weights is governed by a Gamma distribution. 
This setting introduces additional flexibility on the space of functions, but it still does not provide much intuition regarding the properties of the prior.
A different approach is proposed by \cite{Karaletsos2019,Karaletsos2020}, who consider a \gls{GP} model for the network parameters that can capture weight correlations.

{

Bayesian model selection constitutes a principled approach to select an appropriate prior distribution.
Model selection is based on the marginal likelihood -- the normalizing constant of the posterior distribution -- which may be estimated from the training data.
This practice is usually used to select hyperparameters of a \gls{GP} as its marginal likelihood is available in closed form \citep{Rasmussen06}.
However, the marginal likelihood of \glspl{BNN} is generally intractable, and lower bounds are difficult to obtain.
\citet{Graves2011} first and \cite{blundell2015weight} later used the variational lower bound of the marginal likelihood for optimizing the parameters of a prior, yielding in some cases worse results.
Recently, \cite{ImmerBFRK21} extended the Mackay's original proposal \citep{mackay1995probable} of using the Laplace's method to approximate the marginal likelihood.
In this way, one can obtain an estimate of the marginal likelihood which is scalable and differentiable with respect to the prior hyperparameters, such that they can be optimized together with the \gls{BNN} posterior.

}

Many recent attempts in the literature have turned their attention towards defining priors in the space of functions, rather than the space of weights.
For example, \cite{nalisnick2020} consider a family of priors that penalize the complexity of predictive functions.
\cite{Hafner2019} propose a prior that is imposed on training inputs, as well as out-of-distribution inputs.
This is achieved by creating pseudo-data by means of perturbing the training inputs; the posterior is then approximated by a variational scheme.
\cite{Yang2019} present a methodology to induce prior knowledge by specifying certain constraints on the network output.
\cite{Pearce2019} explore \gls{DNN} architectures that recreate the effect of certain kernel combinations for \glspl{GP}.
This result in an expressive family of network priors that converge to \glspl{GP} in the infinite-width limit.

{

A similar direction of research focuses not only on priors but also inference in the space of functions for \glspl{BNN}.
For example, \cite{MaLH19} consider a \gls{BNN} as an implicit prior in function space and then use \glspl{GP} for inference.
Conversely, \cite{Sun2019} propose a functional variational inference which employs a \gls{GP} prior to regularize \glspl{BNN} directly in the function space by estimating the \gls{KL} divergence between these two stochastic processes.
However, this method relies on a gradient estimator which can be inaccurate in high dimensions.
\cite{KhanIAK19} follow an alternative route by deriving a \gls{GP} posterior approximation for neural networks by means of the Laplace and \gls{GGN} approximations, leading to an implicit linearization.
\cite{ImmerKB21} make this linearization explicit and apply it to improve the performance of \gls{BNN} predictions.
In general, these approaches either heavily rely on non-standard inference methods or are constrained to use a certain approximate inference algorithm, such as variational inference or Laplace approximation.

}

A different line of work focuses on meta-learning by adjusting priors based on the performance of previous tasks \citep{Amit2018}.
In contrast to these approaches, we aim to define a suitable prior distribution entirely \emph{a priori}.
We acknowledge that our choice to impose \gls{GP} (or hierarchical \gls{GP}) priors on neural networks is essentially heuristic: there is no particular theory that necessarily claims superiority for this kind of prior distributions.
In some applications, it could be preferable to use priors that are tailored to certain kinds of data or architectures, such the \emph{deep weight prior} \citep{Atanov19}.
However, we are encouraged by the empirical success and the interpretability of \gls{GP} models, and we seek to investigate their suitability as \gls{BNN} priors on a wide range of regression and classification problems.

Our work is most closely related to a family of works that attempt to map \gls{GP} priors to \glspl{BNN}.
\cite{Flam2017} propose to minimize the \gls{KL} between the \gls{BNN} prior and some desired \gls{GP}.
As there is no analytical form for this \gls{KL}, the authors rely on approximations based on moment matching and projections on the observation space.
This limitation was later addressed \citep{Flam2018} by means of a hypernetwork \citep{Ha2017}, which generates the weight parameters of the original \gls{BNN}; the hypernetwork parameters were trained so that a \gls{BNN} fits the samples of a \gls{GP}.
In our work, we also pursue the minimization of a sample-based distance between the \gls{BNN} prior and some desired \gls{GP}, but we avoid the difficulties in working with the \gls{KL} divergence, as its evaluation is challenging due to the empirical entropy term.
To the best of our knowledge, the Wasserstein distance scheme we propose is novel, and it demonstrates satisfactory convergence for compatible classes of \glspl{GP} and \glspl{BNN}.

Concurrently to the release of this paper, we have come across another work advocating for the use of \gls{GP} priors to determine priors for \glspl{BNN}. \cite{Matsubara2020} rely on the \emph{ridgelet transform} to approximate the covariance function of a \gls{GP}.
Our work is methodologically different, as our focus is to propose a practical framework to impose sensible priors.
Most importantly, we present an extensive experimental campaign that demonstrates the impact of functional priors on deep models.

\section{Preliminaries} \label{sec:preliminaries}

In this section, we establish some basic notation on \glspl{BNN} that we follow throughout the paper,
and we review \gls{SGHMC}, which is the main sampling algorithm that we use in our experiments.
Finally, we give a brief introduction to the concept of Wasserstein distance, which is the central element of our methodology to impose functional \gls{GP} priors on \glspl{BNN}.

\subsection{Bayesian Neural Networks} \label{ssec:bayesian_neural_networks}

We consider a \gls{DNN} consisting of $L$ layers, where the output of the $l$-th layer $f_l(\mbx)$ is a function of the previous layer outputs $f_{l-1}(\mbx)$, as follows:
\begin{equation}
	f_{l}(\mbx) = \frac{1}{\sqrt{D_{l-1}}} \bigg( W_{l} \varphi(f_{l-1}(\mbx)) \bigg) + b_{l}, \quad l \in \{1, ..., L\},
\end{equation}
where $\varphi$ is a nonlinearity, $b_{l} \in \mathbb{R}^{D_{l}}$ is a vector containing the bias parameters for layer $l$, and $W_{l} \in \mathbb{R}^{D_{l} \times D_{l-1}}$ is the corresponding matrix of weights.
We shall refer to the union of weight and bias parameters of a layer $l$ as $\mbw_{l}=\{W_{l}, b_{l} \}$, while the entirety of trainable network parameters will be
denoted as $\mbw = \{\mbw_{l}\}_{l=1}^L$.
In order to simplify the presentation, we focus on fully-connected \glspl{DNN}; the weight and bias parameters of \glspl{CNN} are treated in a similar way, unless stated otherwise.

The scheme that involves dividing by $\sqrt{D_{l-1}}$ is known as the \textit{NTK parameterization} \citep{jacot2018neural,lee2020finite}, and it ensures that the asymptotic variance neither explodes nor vanishes.
For fully-connected layers, $D_{l-1}$ is the dimension of the input, while for convolutional layers $D_{l-1}$ is replaced with the filter size multiplied by the number of input channels.

\paragraph{Inference}
The Bayesian treatment of neural networks \citep{mackay1992information, neal1996bayesian} dictates that a prior distribution $p(\mbw)$ is placed over the parameters.
The learning problem is formulated as a transformation of a prior belief into a posterior distribution  by means of Bayes' theorem.
Given a dataset with $N$ input-target pairs $\cD=\{\mbX, \mby\} \defeq \{ (\mathbf{x}_i, y_i) \}_{i=1}^{N}$, the posterior over $\mbw$ is:
\begin{equation}
	p(\mbw \g \cD) = \frac{p(\cD \g \mbw) p(\mbw)}{p(\cD)}.
	\label{eq:bnn_posterior}
\end{equation}
Although the posterior for most nonlinear models, such as neural networks, is analytically intractable, it can be approximated by \gls{MCMC} methods, as they only require an unnormalized version of the target density.
Regarding \cref{eq:bnn_posterior}, the unnormalized posterior density is given by the joint probability in the numerator, which can be readily evaluated if the prior and likelihood densities are known.

\gls{HMC} \citep{Duane1987} considers the joint log-likelihood as a potential energy function $U(\mbw) = -\log p(\cD \g \mbw) - \log p(\mbw)$, and introduces a set of auxiliary momentum variables $\mbr$.
Samples are generated from the joint distribution $p(\mbw, \mbr)$ based on the Hamiltonian dynamics:
\begin{align}
	\mathrm{d} \mbw & = \mbM^{-1} \mbr \mathrm{d}t,   \\
	\mathrm{d} \mbr & = -\nabla U(\mbw) \mathrm{d} t,
\end{align}
where, $\mbM$ is an arbitrary mass matrix that plays the role of a preconditioner.
In practice, this continuous system is approximated by means of a $\varepsilon$-discretized numerical integration and followed by Metropolis steps to accommodate numerical errors stemming from the integration.

However, \gls{HMC} is not practical for large datasets due to the cost of computing the gradient $\nabla U(\mbw) = \nabla \log p(\cD| \mbw)$ on the entire dataset.
To mitigate this issue, \cite{Cheni2014} proposed \gls{SGHMC}, which considers a noisy, unbiased estimate of the gradient $\nabla \tilde{U}(\mbw)$ computed from a mini-batch of the data.
The discretized Hamiltonian dynamics equations are then updated as follows
\begin{align}
	\Delta \mbw & = \varepsilon \mbM^{-1} \mbr,  \label{eq:discr_hamil_weight}                                                                                       \\
	\Delta \mbr & = -\varepsilon \nabla \tilde{U}(\mbw) -\varepsilon \mbC\mbM^{-1}\mbr + \cN(0, 2\varepsilon(\mbC - \tilde{\mbB})),  \label{eq:discr_hamil_momentum}
\end{align}
where $\varepsilon$ is an step size, $\mbC$ is an user-defined friction matrix, $\tilde{\mbB}$ is the estimate for the noise of the gradient evaluation.

In this work, we employ the \gls{SGHMC} algorithm to generate posterior samples for all the models and datasets considered.
The step size $\varepsilon$ as well as the matrices $\mbM$, $\mbC$ and $\tilde{\mbB}$ constitute additional parameters that require careful tuning to guarantee the quality of samples produced by the algorithm.
We adopt the tuning strategy of \cite{Springenberg2016}, which involves a burn-in period during which the matrices $\mbM$ and $\tilde{\mbB}$ are adjusted by monitoring certain statistics of the dynamics.
The only parameters that we manually define are the integration interval and the step size.

\subsection{Gaussian Process Priors} \label{ssec:gp-priors}

\glspl{GP} constitute a popular modeling choice in the field of Bayesian machine learning \citep{Rasmussen06}, as they allow one to associate a certain class of functional representations with a probability measure.
A \gls{GP} is a stochastic process that is uniquely characterized by a mean function $\mu(\mbx)$ and a covariance function $\kappa(\mbx, \mbx')$.
The latter is also known as a kernel function, and it determines the covariance between the realization of the function at pairs of inputs $\mbx$ and $\mbx'$.
For a finite set of inputs $\mbX$, a \gls{GP} yields a multivariate Gaussian distribution with mean vector $\mbmu = \mu(\mbX)$ and covariance matrix $\mbK = \kappa(\mbX, \mbX)$.

There is a significant body of research whose objective is to perform inference for \gls{GP} models; see \cite{Liu2020} for an extensive review.
However, in this work we only treat \glspl{GP} as a means to define meaningful specifications of priors over functions.
Different choices for the kernel result in different priors in the space of functions.
A popular choice in the literature is the \gls{RBF} kernel:
\begin{equation}
	\kappa_{\alpha, l}(\mbx, \mbx') = \alpha^2 \exp(-\frac{(\mbx-\mbx')^\top (\mbx-\mbx')}{l^2}),
\end{equation}
which induces functions that are infinitely differentiable, as in \cref{fig:prior_example}.
The subscripts $\alpha, l$ denote the dependency on hyper-parameters: $\alpha$ is the \emph{amplitude}, which controls the prior marginal standard deviation, and $l$ is known as the \emph{lengthscale}, as it controls how rapidly sample functions can vary.

\paragraph{Hierarchical GP Priors}

The most common practice in \gls{GP} literature is to select values for the hyper-parameters that optimize the marginal log-likelihood.
We do not recommend such an approach in our setting however, as it introduces additional complexity from a computational perspective.
Instead, we opt to consider a hierarchical form for the target prior.
Assuming a shift-invariant kernel $\kappa_{\alpha,l}(\mbx, \mbx^\prime)$ with hyper-parameters $\alpha$ and $l$, we have:
\begin{equation}
	\alpha, l \sim \text{LogNormal}(m, s^2), \qquad f \sim \mathcal{N}(0, \kappa_{\alpha,l}(\mbx, \mbx^\prime))
	\label{eq:hgp_sample}
\end{equation}
where $m$ and $s$ are user-defined parameters.
Samples of the target prior are generated by means of a Gibbs sampling scheme: we first sample the hyper-parameters from a log-normal distribution, and then we sample from the corresponding \gls{GP}.
This form of hierarchical \gls{GP} priors is adopted in the majority of experiments of \cref{sec:experiments}, unless otherwise specified.

\subsection{Wasserstein Distance} \label{ssec:wasserstein_dist}

The concept of distance between probability measures is central to this work, as we frame the problem of imposing a \gls{GP} prior on a \gls{BNN} as a distance minimization problem.
We present some known results on the Wasserstein distance that will be used in the sections that follow.
Given two {\it Borel's probability measures} $\pi(\mbx)$ and $\nu(\mby)$ defined on the {\it Polish space} $\cX$ and $\cY$ (i.e. any complete separable metric space), the generic formulation of the {\bf $p$-Wasserstein distance} is defined as follows:
\begin{align}
	W_p(\pi, \nu) = \left(\inf_{\gamma \in \Gamma(\pi, \nu)} \int_{\cX\times\cY}D(\mbx,\mby)^p \gamma(\mbx,\mby) \d{\mbx}\d{\mby}  \right)^{1/p}\,,
\end{align}
where $D(\mbx,\mby)$ is a proper distance metric between two points $\mbx$ and $\mby$ in the space $\cX\times\cY$ and $\Gamma(\pi, \nu)$ is the set of functionals of all possible joint densities $\gamma$ whose marginals are $\pi$ and $\nu$.

When the spaces of $\mbx$ and $\mby$ coincide (i.e. $\mbx,\mby\in\cX\subseteq\bbR^d$), with $D(\mbx,\mby)$ being the Euclidian norm distance, the Wasserstein-1 distance (also known in the literature as Earth-Mover distance) takes the following shape,
\begin{align}
	\label{eq:wasserstein-1}
	W_{1}(\pi, \nu) & = \inf_{\gamma \in \Gamma(\pi, \nu)} \int_{\cX\times\cX}\|\mbx - \mby\| \gamma(\mbx,\mby) \d{\mbx}\d{\mby} \,.
	\end{align}
With the exception of few cases where the solution is available analytically (e.g. $\pi$ and $\nu$ being Gaussians), solving \cref{eq:wasserstein-1} directly or via optimization is intractable.
On the other hand, the Wasserstein distance defined in \cref{eq:wasserstein-1} admits the following dual form \citep{Kantorovich1942,Kantorovich1948},
\begin{align}
	\label{eq:theo-wasserstein-dual}
	W_{1}(\pi, \nu) & = \sup_{\|\phi\|_{L}\leq1} \left[\int \phi(\mbx)\pi(\mbx)d\mbx - \int \phi(\mby)\nu(\mby)d\mby\right] \nonumber \\
	                & = \sup_{\|\phi\|_{L}\leq1} \E_{\pi} \phi(\mbx) - \E_{\nu} \phi(\mbx) \,,
\end{align}
where $\phi$ is a 1-Lipschitz continuous function defined on $\cX\rightarrow\bbR$.
This is effectively a functional maximization over $\phi$ on the difference two expectations of $\phi$ under $\pi$ and $\nu$.
A revised proof of this dual form by \citet{Villani2003} is available in the Supplement.

\section{Imposing Gaussian Process Priors on Bayesian Neural Networks} \label{sec:imposing-priors}

The equivalence between function-space view and weight-space view of linear models, like Bayesian linear regression and \glspl{GP} \citep{Rasmussen06}, is a straightforward application of Gaussian identities, but it allows us to seamlessly switch point of view accordingly to which characteristics of the model we are willing to observe or impose.
We would like to leverage this equivalence also for \glspl{BNN} but the nonlinear nature of such models makes it analytically intractable (or impossible, for non-invertible activation functions).
We argue that for \glspl{BNN}---and Bayesian deep learning models, in general---starting from a prior over the weights is not ideal, given the impossibility of interpreting its effect on the family of functions that the model can represent.
We therefore rely on an optimization-based procedure to impose functional priors on \glspl{BNN} using the  Wasserstein distance as a similarity metric between such distributions, as described next.

\subsection{Wasserstein Distance Optimization} \label{sec:wasserstein_dist_optim}

Assume a prior distribution $p(\mbw; \mbpsi)$ on the weights of a \gls{BNN}, where $\mbpsi$ is a set of parameters that determine the prior (e.g., $\mbpsi = \{\mu, \sigma\}$ for a Gaussian prior; we discuss more options on the parametrization of \gls{BNN} priors in the section that follows).
This prior over weights induces a prior distribution over functions:
\begin{equation}
	p_\sub{nn}(\mbf ; \mbpsi) = \int p(\mbf\g\mbw) p(\mbw;\mbpsi) \d\mbw,
	\label{eq:nn_prior}
\end{equation}
where $p(\mbf\g\mbw)$ is deterministically defined by the network architecture.

In order to keep the notation simple, we consider non-hierarchical \gls{GP} priors.
Hierarchical \glspl{GP} are treated in the same way, except that samples are generated by the Gibbs sampling scheme of \cref{eq:hgp_sample}.
Our target \gls{GP} prior is $p_\sub{gp}(\mbf \g\mathbf{0}, \mbK)$, where $\mbK$ is the covariance matrix obtained by computing the kernel function $\kappa$ for each pair of $\{\mbx_i, \mbx_j\}$ in the training set.
We aim at matching these two stochastic processes at a finite number of measurement points $\mbX_{\mathcal{M}} \defeq [\mbx_1, ..., \mbx_M]^{\top}$ sampled from a distribution $q(\mbx)$.
To achieve this, we propose a sample-based approach using the 1-Wasserstein distance in \cref{eq:theo-wasserstein-dual} as objective:
\begin{align}
	\min_\mbpsi \max_\mbtheta\mathbb{E}_{q} \Big[ \underbrace{\mathbb{E}_{p_\sub{gp}} [\phi_{\mbtheta} (\mbf_{\mathcal{M}})] - \mathbb{E}_{p_{\sub{nn}}} [\phi_{\mbtheta} (\mbf_{\mathcal{M}})]}_{\cL(\mbpsi, \mbtheta)} \Big] \,,
	\label{eq:objective_function}
\end{align}
where $\mbf_{\mathcal{M}}$ denotes the set of random variables associated with the inputs at $\mbX_{\mathcal{M}}$, and $\phi_{\mbtheta}$ is a 1-Lipschitz function.
Following recent literature \citep{goodfellow2014generative, arjovsky2017wasserstein}, we parameterize the Lipschitz function by a neural network\footnotemark with parameters $\mbtheta$.
\footnotetext{Details on the 1-Lipschitz function: we used a \gls{MLP} with two hidden layers, each with 200 units; the activation function is softplus, which is defined as: $\mathrm{softplus}(x) = 1 / (1+\exp(-x))$.}

\begin{figure}[]
	\includegraphics[width=1.0\textwidth]{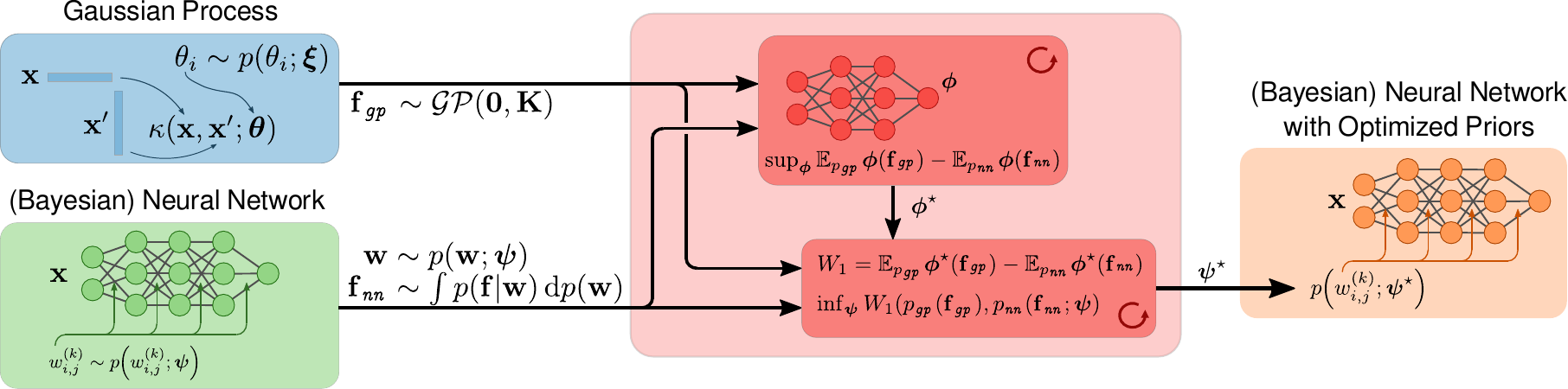}
	
	\caption{Schematic representation of the process of imposing \gls{GP} priors on \glspl{BNN} via Wasserstein distance minimization.
		\label{fig:schematic}
	}
\end{figure}

{
	Regarding the optimization of the $\mbtheta$ and $\mbpsi$ parameters we alternate between $n_{\mathrm{Lipschitz}}$ steps of maximizing $\cL$ with respect to the Lipschitz function's parameters $\mbtheta$ and one step of minimizing the Wasserstein distance with respect to the prior's parameters $\mbpsi$.
	We therefore use two independent optimizers \citep[RMSprop--see, for example,][]{Tieleman2012} for $\mbtheta$ and $\mbpsi$.
	\Cref{fig:schematic} offers a high-level schematic representation of the proposed procedure.
	Given samples from two stochastic processes, the Wasserstein distance is estimated by considering the inner maximization of \cref{eq:objective_function}, resulting in an optimal $\phi^*$.
	This inner optimization step is repeated for every step of the outer optimization loop.
	Notice that the objective is fully sample-based.
	As a result, it is not necessary to know the  closed-form of the marginal density $p_{\sub{nn}}(\mbf ; \mbpsi)$.
	One may consider any stochastic process as a target prior over functions, as long as we can draw samples from it (e.g., a hierarchical \gls{GP}).
}
Finally, we acknowledge that the two training steps could have been optimized jointly in a single loop, as \cref{eq:theo-wasserstein-dual} defines a minimax problem.
However, this choice allows $\phi_{\mbtheta}$ to converge enough before a single Wasserstein minimization step takes place.
In fact, this is a common trick to make convergence more stable (see e.g., the original \citet{goodfellow2014generative} paper, which suggests to allow more training of the discriminator for each step of the generator).
In \cref{ssec:wd_opt} we further discuss this choice and we show qualitatively the convergence improvements.

\paragraph{Lipschitz constraint.}
In order to enforce the Lipschitz constraint on $\phi_{\mbtheta}$, \citet{arjovsky2017wasserstein} propose to clip the weights $\mbtheta$ to lie within a compact space $[-c, c]$ such that all functions $\phi_\mbtheta$ are K-Lipschitz.
This approach usually biases the resulting $\phi_{\mbtheta}$ towards a simple function.
Based on the fact that a differentiable function is 1-Lipschitz if and only if the norm of its gradient is at most one everywhere, \citet{gulrajani2017improved} propose to constrain the gradient norm of the output of the Lipschitz function $\phi_{\mbtheta}$ with respect to its input.
More specifically, the loss of the Lipschitz function is augmented by a regularization term
\begin{align}
	\cL_R(\mbpsi, \mbtheta) = \cL(\mbpsi, \mbtheta) +
	\underbrace{\lambda \mathbb{E}_{p_{\hat{\mbf}}}\left[\Big(\norm{\grad_{\hat{\mbf}}\phi(\hat\mbf)}_{2} -1\Big)^{2}\right]}_{\text{Gradient penalty}}.
	\label{eq:gradient_penalty}
\end{align}
Here $p_{\hat{\mbf}}$ is the distribution of ${\hat{\mbf}} = \varepsilon {\mbf_{\sub{nn}}} + (1-\varepsilon) {\mbf_{\sub{gp}}}$ for $\varepsilon \sim \mathcal{U}[0,1]$ and $\mbf_{\sub{nn}} \sim p_{\sub{nn}}$, $\mbf_{\sub{gp}} \sim p_{\sub{gp}}$ being the sample functions from \gls{BNN} and \gls{GP} priors, respectively; $\lambda$ is a penalty coefficient.

\paragraph{Choice of the measurement set.}
In our formulation, we consider finite measurement sets to have a practical and well-defined optimization strategy.
	{
		As discussed by \cite{ShiK019}, there are several approaches to define the measurement set for functional-space inference \citep{Hafner2019, Sun2019}.
		For low-dimensional problems, one can simply use a regular grid or apply uniform sampling in the input domain.
		For high-dimensional problems, one can sample from the training set, possibly with augmentation, where noise is injected into the data.
		In applications where we know the input region of the test data points, we can set $q(\mbx)$ to include it.
		We follow a combination of the two approaches: we use the training inputs (or a subset of thereof) as well as additional points that are randomly sampled (uniformly) from the input domain.
	}

\subsection{Prior Parameterization for Neural Networks}
\label{ssec:parameterization}

In the previous section, we have treated the parameters of a \gls{BNN} prior $p_\sub{nn}(\mbf ; \mbpsi)$ in a rather abstract manner.
Now we explore three different parametrizations of increasing complexity.
The only two requirements needed to design a new parametrization are (1) to be able to generate samples and (2) to compute the log-density at any point; the latter is required to be able to draw samples from the posterior over model parameters using most \gls{MCMC} sampling methods, such as \gls{SGHMC} which we employ in this work.

\paragraph{Gaussian prior on weights.}
We consider a layer-wise factorization with two independent zero-mean Gaussian distributions for weights and biases.
The parameters to adjust are $\mbpsi = \{\sigma_{l_{\text{w}}}^2, \sigma_{l_{\text{b}}}^2\}_{l=1}^L$, where $\sigma_{l_{\text{w}}}^2$ is the prior variance shared across all weights in layer $l$, and $\sigma_{l_{\text{b}}}^2$ is the respective variance for the bias parameters.
For any weight and bias entries $w_{l}, b_{l} \in \mbw_{l}$ of the $l$-th layer, the prior is:
\begin{equation*}
	p(w_{l}) = \cN\left(w_{l};\; 0, \sigma_{l_{\text{w}}}^2\right) \quad \text{and} \quad p(b_{l}) = \cN\left(b_{l};\; 0, \sigma_{l_{\text{b}}}^2\right).
	\end{equation*}
In the experimental section, we refer to this parametrization as the \emph{\gls{GP}-induced \gls{BNN} prior with Gaussian weights} (\gpig).
Although this simple approach assumes a Gaussian prior on the parameters, in many cases it is sufficient to capture the target \gls{GP}-based functional priors.

Regarding the implementation of this scheme, there are a few technical choices to discuss.
In order to maintain positivity for the standard deviation $\sigma$ and perform unconstrained optimization, we optimize $\rho$ such that $\sigma = \log(1 + e^{\rho})$, which guarantees that $\sigma$ is always positive.
Also, we have to use gradient backpropagation through stochastic variables such as $w_{l}$.
Thus, in order to treat the parameter $w_{l}$ in a deterministic manner, instead of sampling the prior distribution directly $w_{l} \sim \cN\left(w_{l};\; 0, \sigma_{l_{\text{w}}}^2\right)$, we use the reparameterization trick \citep{rezende2014stochastic, Kingma14}, and sample from the noise distribution instead,
\begin{equation}
	w_{l} := \sigma_{l_{\text{w}}} \varepsilon, \quad \varepsilon \sim \cN(0, 1).
	\label{eq:reparam_trick}
\end{equation}

\paragraph{Hierarchical prior.}
A more flexible family of priors for \glspl{BNN} considers a hierarchical structure where the network parameters follow a conditionally Gaussian distribution, and the prior variance for each layer follows an Inverse-Gamma distribution.
For the weight and bias variances we have:
\begin{equation*}
	\sigma_{l_{\text{w}}}^2 \sim \Gamma^{-1}(\alpha_{l_{\text{w}}}, \beta_{l_{\text{w}}})
	\quad \text{and} \quad
	\sigma_{l_{\text{b}}}^2 \sim \Gamma^{-1}(\alpha_{l_{\text{b}}}, \beta_{l_{\text{b}}})
\end{equation*}
In this case, we have $\mbpsi = \{ \alpha_{l_{\text{w}}}, \beta_{l_{\text{w}}}, \alpha_{l_{\text{b}}}, \beta_{l_{\text{b}}} \}_{l=1}^L$, where $\alpha_{l_{\text{w}}}, \beta_{l_{\text{w}}}, \alpha_{l_{\text{b}}}, \beta_{l_{\text{b}}}$ denote the shape and rate parameters of the Inverse-Gamma distribution for the weight and biases correspondingly for layer $l$.
The conditionally Gaussian prior over the network parameters is given as in the previous section.
In the experiments, we refer to this parametrization as the \emph{\gls{GP}-induced \gls{BNN} prior with Hierarchically-distributed weights} (\gpih).

Similar to the Gaussian prior, we impose positivity constraints on the shape and rate of the Inverse-Gamma distribution.
In addition, we apply the reparameterization trick proposed by \cite{jankowiak2018pathwise} for the Inverse-Gamma distribution.
This method computes an implicit reparameterization using a closed-form approximation of the CDF derivative.
We used the corresponding original PyTorch \citep{paszke2019pytorch} implementation of the method in our experiments.

\paragraph{Beyond Gaussians with Normalizing flows.}

Finally, we also consider \glspl{NF} as a family of much more flexible distributions.
By considering an invertible, continuous and differentiable function $t$
$:\bbR^{D_l}\rightarrow\bbR^{D_l}$, where $D_l$ is the number of parameters for $l$-th layer,
a \gls{NF} is constructed as a sequence of $K$ of such transformations $\cT_K = \{t_1, \dots, t_K\}$ of a simple known distribution (e.g., Gaussian).
Sampling from such distribution is as simple as sampling from the initial distribution and then apply the set of transformation $\cT_K$.
Given an initial distribution $p_0(\mbw_l)$, by denoting $p (\cT_K(\mbw_l))$ the final distribution, its log-density can be analytically computed by taking into account to Jacobian of the transformations as follows,
\begin{align}
	\log p (\cT_K(\mbw_l)) = \log p_0(\mbw_l) - \sum_{k=1}^K \log \left| \det \partialderivative{t_k(\mbw_{l_{k-1}})}{\mbw_{l_{k-1}}} \right|,
\end{align}
where $\mbw_{l_{k-1}} = (t_{k-1} \circ ... \circ t_2 \circ t_1) (\mbw_l)$ for $k > 1$, and $\mbw_{l_{0}} = \mbw_l$.

	{
		We shall refer to this class of \gls{BNN} priors as the \emph{\gls{GP}-induced \gls{BNN} prior, parametrized by normalizing flows} (\gpinf).
		We note that \glspl{NF} are typically used differently in the literature; while previous works showed how to use this distributions for better approximation of the posterior in variational inference \citep{Rezende2015,Kingma2016,Louizos2017} or for parametric density estimation \citep[e.g.,][]{Grover2018}, or for enlarging the flexibility of a prior for \glspl{VAE} \citep[e.g.,][]{Chen2017VariationalLA}, as far as we are aware this is the first time that \glspl{NF} are used to characterize a prior distribution for \glspl{BNN}.
	}

In our experiments, we set the initial distribution $p_{0}(\mbw_l)$ to a fully-factorized Gaussian
$\cN(\mbw_l \g \mathbf{0}, \mbsigma_l^{2} \mbI)$.
We then employ a sequence of four \textit{planar flows} \citep{Rezende2015}, each defined as
\begin{align}
	t_k(\mbw_{l_{k-1}}) = \mbw_{l_{k-1}} + \mbu_{l_{k}} h(\mbtheta_{l_{k}}^{^\top} \mbw_{l_{k-1}} + b_{l_{k}}),
	\label{eq:planar_flow}
\end{align}
where $\mbu_{l_{k}} \in \mathbb{R}^{D_l}, \mbtheta_{l_{k}} \in \mathbb{R}^{D_l}, b_{l_{k}} \in \mathbb{R}$ are trainable parameters, and $h(\cdot) = \mathrm{tanh}(\cdot)$.
The log-determinant of the Jacobian of $t_k$ is
\begin{align}
	\log \left| \det \frac{\partial t_k(\mbw_{l_{k-1}})}{\partial \mbw_{l_{k-1}}} \right| & =  \log \left| 1 + \mbu_{l_{k}}^{\top} \mbtheta_{l_{k}} h'(\mbtheta_{l_{k}}^{\top}\mbw_{l_{k-1}} + b_{l_{k}}) \right|.
\end{align}
Thus for the $l$-th \gls{BNN} layer, the parameters to optimize are $\mbpsi_l = \{\mbsigma_l^2\} \bigcup \{ \mbu_{l_{k}}, \mbtheta_{l_{k}}, b_{l_{k}} \}_{k=1}^{K}$.

\begin{algorithm}
	\caption{{Wasserstein Distance Optimization}} \label{alg:main}
	\SetAlgoLined
	\small
	\textbf{Requires}: {$N_s$, number of stochastic process samples;
	$q(\mbx)$, sampling distribution for measurement set; $n_{\mathrm{Lipschitz}}$, number of iterations of Lipschitz function per prior iteration}\;
	\While{$\mbpsi$ has not converged}{
	draw $\mbX_{\mathcal{M}}$ from $q(\mbx)$ \texttt{// Sample measurement set}  \;

	\For{$t=1, ..., n_{\mathrm{Lipschitz}}$}{
	
	draw GP functions $\{\mbf_{\sub{gp}}^{(i)}\}_{i=1}^{N_s} \sim p_{\sub{gp}}(\mbf;\kappa)$ at $\mbX_{\mathcal{M}}$\;

	draw NN functions $\{\mbf_{\sub{nn}}^{(i)}\}_{i=1}^{N_s} \sim p_{\sub{nn}}(\mbf;\mbpsi)$ at $\mbX_{\mathcal{M}}$\;

	$\cL_R = N_s^{-1}\sum_{i=1}^{N_s}\cL_R^{(i)}$ \texttt{// Compute Lipschitz objective $\cL_R$ using \cref{eq:gradient_penalty}}  \;

	$\mbtheta \leftarrow \mathrm{Optimizer}(\mbtheta, \grad_{\mbtheta}\cL_R)$\texttt{ // Update Lipschitz function $\phi_{\mbtheta}$} \;

	}

	draw GP functions $\{\mbf_{\sub{gp}}^{(i)}\}_{i=1}^{N_s} \sim p_{\sub{gp}}(\mbf;\kappa)$ at $\mbX_{\mathcal{M}}$\;

	draw NN functions $\{\mbf_{\sub{nn}}^{(i)}\}_{i=1}^{N_s} \sim p_{\sub{nn}}(\mbf;\mbpsi)$ at $\mbX_{\mathcal{M}}$\;

	$\widetilde{W}_1 = N_s^{-1}\sum_{i=1}^{N_s} \phi_{\mbtheta} \big( \mbf_{\sub{gp}}^{(i)} \big) - \phi_{\mbtheta} \big( \mbf_{\sub{nn}}^{(i)} \big)$ \texttt{// Compute Wasserstein-1 distance using \cref{eq:objective_function}}  \; 
	$\mbpsi \leftarrow \mathrm{Optimizer}(\mbpsi, \grad_{\mbpsi} \widetilde{W}_1)$ \texttt{// Update prior $p_{\sub{nn}}$} \; 
	}

\end{algorithm}

\subsection{Algorithm and Complexity} \label{sec:algo_complexity}
\Cref{alg:main} summarizes our proposed method in pseudocode.
The outer loop is essentially a gradient descent scheme that updates the $\mbpsi$ parameters that control the \gls{BNN} prior.
The inner loop is responsible for the optimization of the Lipschitz function $\phi_{\mbtheta}$, which is necessary to estimate the Wasserstein distance.
The computational complexity is dominated by the number of stochastic process samples $N_s$ used for the calculation of the Wasserstein distance, and the size $N_{\mathcal{M}}$ of the measurement set $\mbX_{\mathcal{M}}$.

Sampling from a \gls{BNN} prior does not pose any challenges; $N_s$ samples can be generated in $\mathcal{O}(N_s)$ time.
However, sampling from a \gls{GP} is of cubic complexity, as it requires linear algebra operations such as the Cholesky decomposition.
The total complexity of sampling from a hierarchical \gls{GP} target is $\mathcal{O}(N_s^2 N_{\mathcal{M}}^3)$, as the Cholesky decomposition should be repeated for every sample.
For a single step of the outer loop in \cref{alg:main}, we have to account the $n_{\mathrm{Lipschitz}}$ steps required for the calculation of the distance, resulting in
complexity of $\mathcal{O}(n_{\mathrm{Lipschitz}} N_s^2 N_{\mathcal{M}}^3)$ per step.
Although our approach introduces an extra computational burden, we note that this is not directly connected to the size of the dataset.
We argue that it is worthwhile to invest this additional cost before the actual posterior sampling phase (via \gls{SGHMC}), and this is supported by our extensive experimental campaign.

The complexity also depends on the number of parameters in $\psi$, whose size is a function of the network architecture and the prior parameterization.
For the Gaussian and hierarchical parameterizations discussed in \cref{ssec:parameterization} (i.e.~ \gpig and \gpih), the set $\mbpsi$ grows sub-lineraly with the number of network parameters, as we consider a single weight/bias distribution per layer.
The obvious advantage of this arrangement is that our approach can be easily scaled to deep architectures, such as \preresnet and \vgg, as we demonstrate in the experiments.

In the case where \gls{BNN} weight and bias distributions are represented by normalizing flows, the size of $\mbpsi$ grows linearly with the total number of \gls{BNN} parameters $N_{\gls{BNN}}$.
More formally, for a sequence of $K$ transformations, the number of prior parameters that we need to optimize is of order $\mathcal{O}(K N_{\gls{BNN}})$.
This might be an issue for more complex architectures; in our experiments we apply the \gpinf configuration for fully connected \glspl{BNN} only.
A more efficient prior parameterization that relies on normalizing flows requires some kind of sparification, which is subject of future work.

\section{Examples and Practical Considerations} \label{sec:practical-considerations}

We shall now elaborate on some of the design choices that we have made in this work.
First, we visually show the prior one can obtain by using our proposed procedure on a 1D regression (\cref{ssec:1d_regression}) and how the choice of \gls{GP} priors (in terms of kernel parameters) affects the \gls{BNN} posterior for 2D classification examples (\cref{ssec:2d_classification}).
We then empirically demonstrate that the proposed optimization scheme based on the Wasserstein distance produces a consistent convergence behavior when compared with a \gls{KL}-based approach (\cref{ssec:wasserstein_vs_kl}).
	
	{
		For these experiments and the rest of the empirical evaluation, we use \gls{SGHMC} \citep{Springenberg2016} for posterior inference.
		The likelihood for regression and classification are set to Gaussian and Bernoulli/multinomial, respectively.
		Unless otherwise specified, we run four parallel \gls{SGHMC} chains with a step size of $0.01$ and a momentum coefficient of $0.01$.
		We assess the convergence of the predictive posterior based on the $\hat{R}$-statistic \citep{Gelman92} over the four chains.
		In all our experiments, we obtain $\hat{R}$-statistics below $1.1$, which indicate convergence to the underlying distribution.
		To further validate the obtained samples from \gls{SGHMC}, for a selection of medium-sized datasets we also run a carefully tuned \gls{HMC} obtaining similar results (see \cref{table:uci_hmc} in the Appendix).
	}

\begin{figure}[t]
	\pgfplotsset{every axis title/.append style={yshift=-1ex}}
	\captionsetup[subfigure]{labelformat=empty,justification=centering}
	\centering
	\scriptsize
	\sffamily
	\setlength\figureheight{.215\textwidth}
	\setlength\figurewidth{.31\textwidth}  
	\begin{subfigure}[t]{.25\textwidth}
		\includegraphics{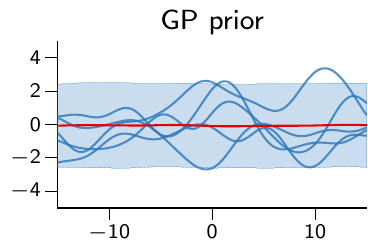}
	\end{subfigure}\begin{subfigure}[t]{.25\textwidth}
		\includegraphics{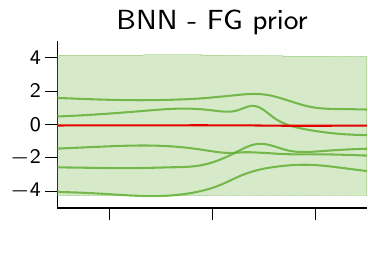}
	\end{subfigure}\begin{subfigure}[t]{.25\textwidth}
		\includegraphics{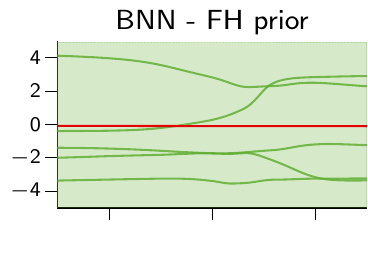}
	\end{subfigure}\begin{subfigure}[t]{.25\textwidth}
		\includegraphics{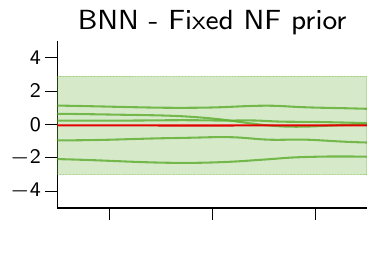}
	\end{subfigure}\\

	\begin{subfigure}[t]{.25\textwidth}
		\hfill
	\end{subfigure}\begin{subfigure}[t]{.25\textwidth}
		\includegraphics{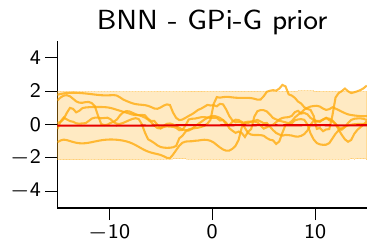}
	\end{subfigure}\begin{subfigure}[t]{.25\textwidth}
		\includegraphics{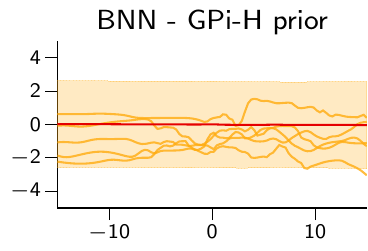}
	\end{subfigure}\begin{subfigure}[t]{.25\textwidth}
		\includegraphics{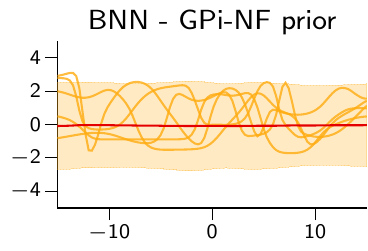}
	\end{subfigure}\\

	\hrule

	\begin{subfigure}[t]{.25\textwidth}
		\hfill
	\end{subfigure}\begin{subfigure}[t]{.25\textwidth}
		\includegraphics{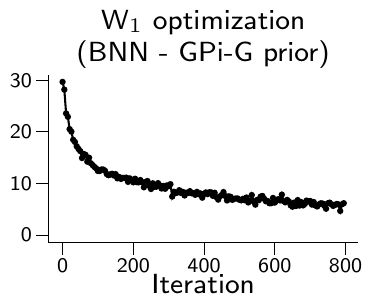}
	\end{subfigure}\begin{subfigure}[t]{.25\textwidth}
		\includegraphics{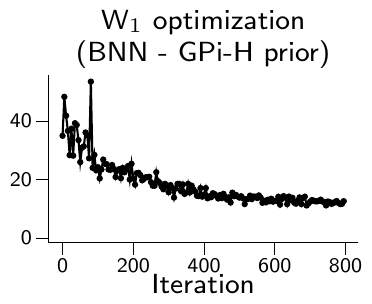}
	\end{subfigure}\begin{subfigure}[t]{.25\textwidth}
		\includegraphics{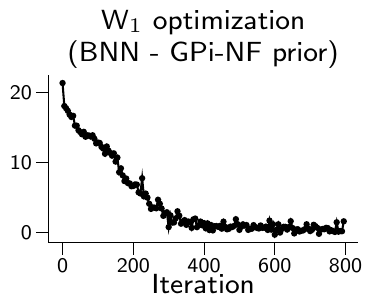}
	\end{subfigure} \\

	\hrule

	\begin{subfigure}[t]{.25\textwidth}
		\includegraphics{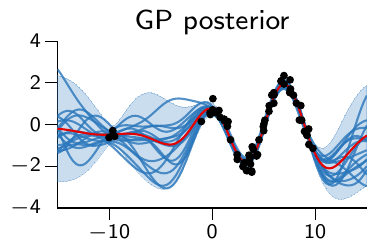}
	\end{subfigure}\begin{subfigure}[t]{.25\textwidth}
		\includegraphics{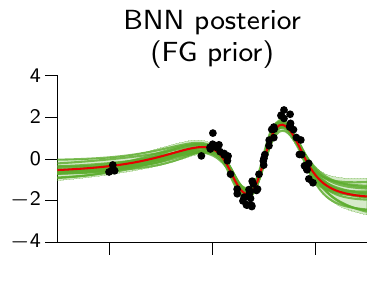}
	\end{subfigure}\begin{subfigure}[t]{.25\textwidth}
		\includegraphics{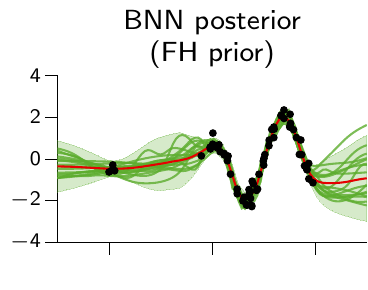}
	\end{subfigure}\begin{subfigure}[t]{.25\textwidth}
		\includegraphics{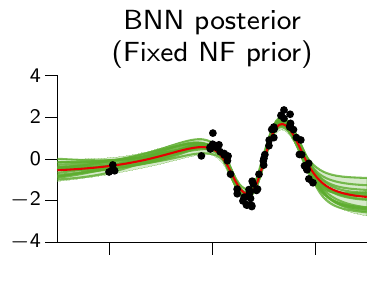}
	\end{subfigure} \\

	\begin{subfigure}[t]{.25\textwidth}
		\hfill
	\end{subfigure}\begin{subfigure}[t]{.25\textwidth}
		\includegraphics{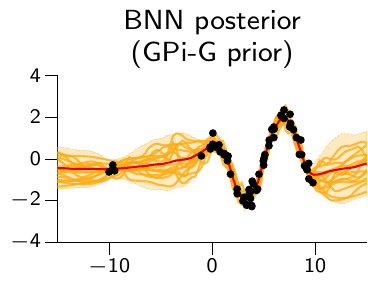}
	\end{subfigure}\begin{subfigure}[t]{.25\textwidth}
		\includegraphics{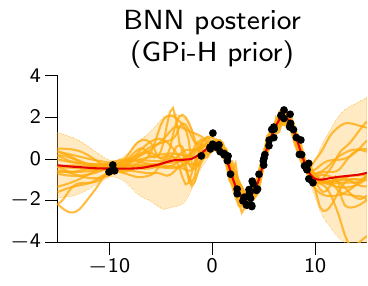}
	\end{subfigure}\begin{subfigure}[t]{.25\textwidth}
		\includegraphics{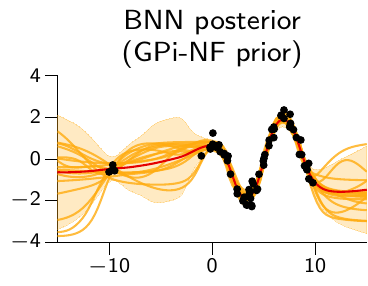}
	\end{subfigure}\\

	\caption{Visualization of one-dimensional regression example with a three-layer \gls{MLP}. The first two rows illustrate the prior sample and distributions, whereas the last two rows show the corresponding posterior distributions. The means and the 95\% credible intervals are represented by red lines and shaded areas, respectively. The middle row shows progressions of the prior optimization.
	}
	\label{fig:1d-prior-comparison}
\end{figure}
\subsection{Visualization on a 1D regression synthetic dataset}\label{ssec:1d_regression}
The dataset used is built as follows: (1) we uniformly sample 64 input locations $\mbx$ in the interval $[-10, 10]$; (2) we rearrange the locations on a defined interval to generate a gap in the dataset; (3) we sample a function $\mbf$ from the \gls{GP} prior ($l=0.6, \alpha=1$) computed at locations $\mbx$; (4) we corrupt the targets with i.i.d. Gaussian noise ($\sigma_{\epsilon}^2 = 0.1$).
In this example, we consider a three-layer \gls{MLP}.
\cref{fig:1d-prior-comparison} shows all the results.
The first two rows illustrate the different choice of priors.
For the Wasserstein-based functional priors (\gpig, \gpih, \gpinf), the third row shows the convergence of the optimization procedure.
Finally, the last two rows represent the posterior collected by running \gls{SGHMC} with the corresponding priors.

From the analysis of these plots, we clearly see the benefit of placing a prior on the functions rather than on the parameters.
First, the Wasserstein distance plots show satisfactory convergence, with the normalizing flow prior closely matching the \gls{GP} prior.
Second, as expected, the posteriors exhibit similar behavior according to the possible solutions realizable from the prior: classic priors tend to yield degenerate functions resulting in overconfidence in regions without data, while our \gls{GP}-based priors (\gpig, \gpih, \gpinf) retain information regarding lengthscale and amplitude.

\setlength\figureheight{.26\textwidth}
\setlength\figurewidth{.26\textwidth}
\begin{figure}[t]
	\begin{subfigure}[b]{0.79\textwidth}
		\centering
		\includegraphics{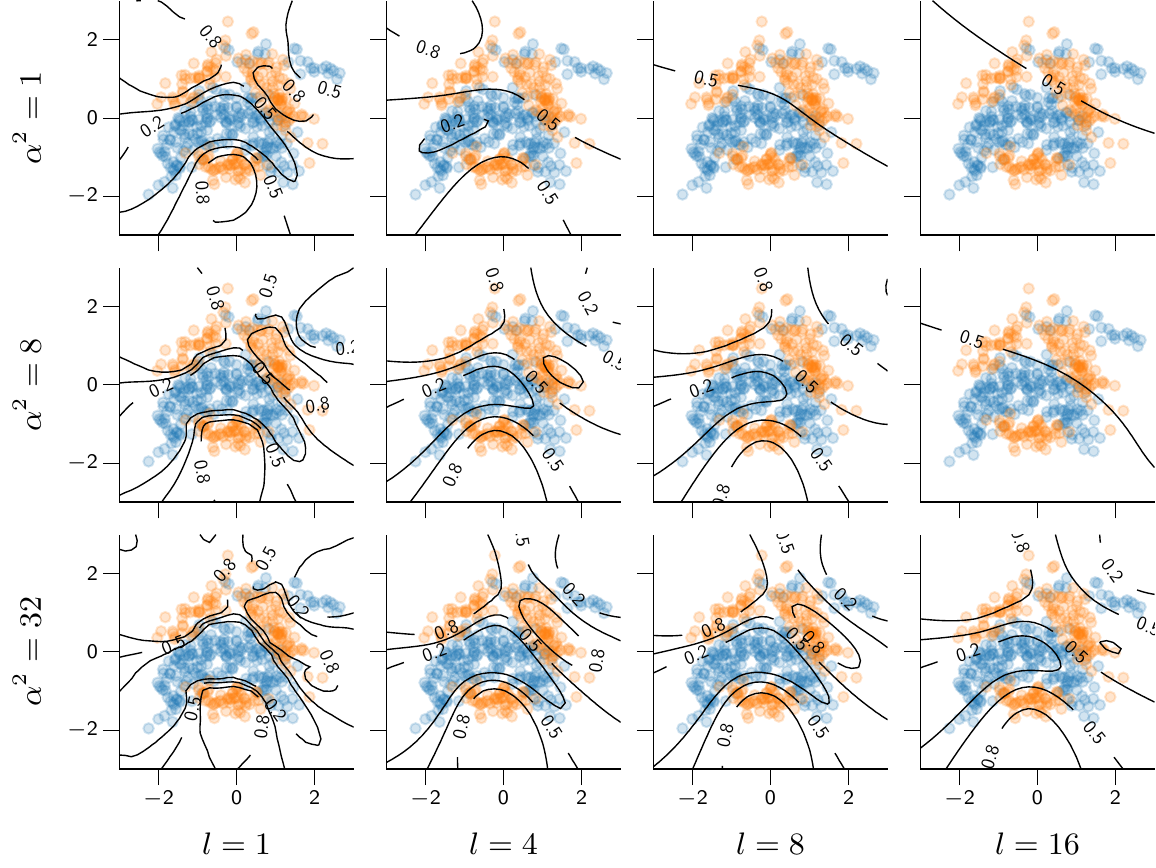}
	\end{subfigure}
	\vrule
	\begin{subfigure}[b]{0.19\textwidth}
		\centering
		\includegraphics{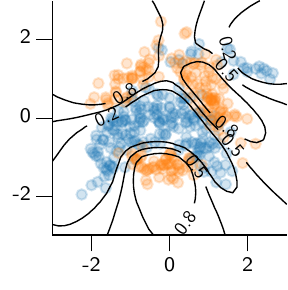}
	\end{subfigure}
	\caption{\textbf{(Left)} The effect of using different hyper-parameters of the RBF kernel of the target \gls{GP} prior to the predictive posterior.
		Rows depict increasing the amplitude $\alpha$, whist columns show increasing the lengthscale $l$.
		In each panel the orange and blue dots represent the training points from the two different classes, while the black lines represent decision boundaries at different confidence levels.
		\textbf{(Right)} The predictive posterior with respect to using a target hierarchical-\gls{GP} prior, in which hyper-priors $\mathrm{LogNormal}(\log\sqrt{2D}, 1)$ and $\mathrm{LogNormal}(\log 8, 0.3)$ are employed on the lengthscales $l$ and variance $\alpha^2$ respectively, where $D$ is the number of input dimensions.
		\label{fig:banana_full}
	}
\end{figure}

\subsection{The effects of the GP prior on the BNN posterior} \label{ssec:2d_classification}
In order to gain insights into the effect of the \gls{GP} prior (i.e., kernel parameters), we set up an intuitive analysis on the \banana dataset.
We can define the regularization strength of the prior in a sensible way by modifying the hyper-parameters of the RBF kernel.
\cref{fig:banana_full} (left) illustrates the predictive posterior of a two-layer \gls{BNN}, whose prior has been adapted to different target \gls{GP} priors, featuring different hyper-parameters.
We observe that the decision boundaries are more complex for smaller lengthscales $l$ and larger amplitudes $\alpha$, while in the opposite case, we obtain posterior distributions that are too smooth.
This behavior reflects the properties of the induced prior.

In a regular \gls{GP} context, it is possible to tune these hyper-parameters by means of marginal likelihood maximization.
This is \textit{not} the way we proceed, for two reasons: (1) the overhead to solve the \gls{GP} and (2) the uselessness of the overall procedure (solving the task with \glspl{GP}, so to then pick the converged \gls{GP} prior to solve the \gls{BNN} inference).
As discussed in \cref{ssec:gp-priors}, we approach this issue by means of hierarchical \glspl{GP}.
In the rightmost plot of \cref{fig:banana_full}, we include the \gls{BNN} posterior that was adapted to a hierarchical-\gls{GP} target.
Since samples from the target prior can be easily generated using a Gibbs sampling scheme, we can positively impact the expressiveness of the \gls{BNN} posterior without explicitly worrying which \gls{GP} prior works best.

\begin{figure}[]
	\captionsetup[subfigure]{labelformat=empty}
	\centering
	\setlength\figureheight{.245\textwidth}
	\setlength\figurewidth{.35\textwidth}
	\scriptsize
	\includegraphics{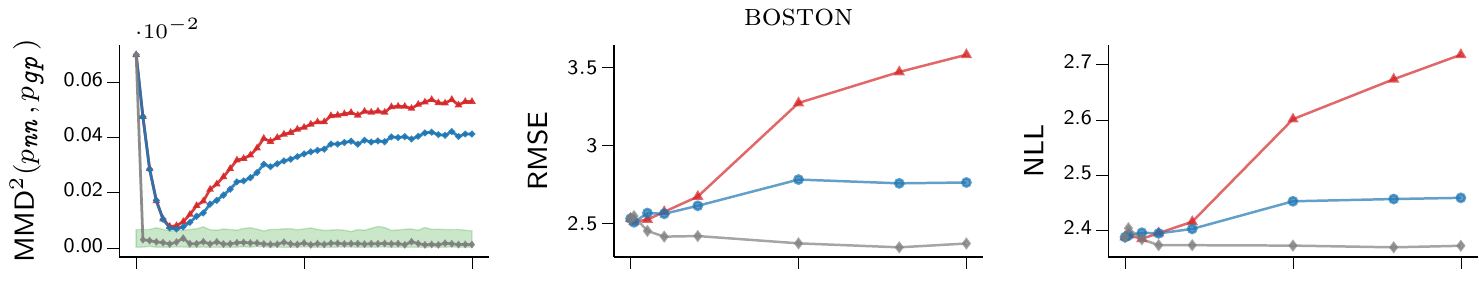}\\
	\vskip -0.1in
	\includegraphics{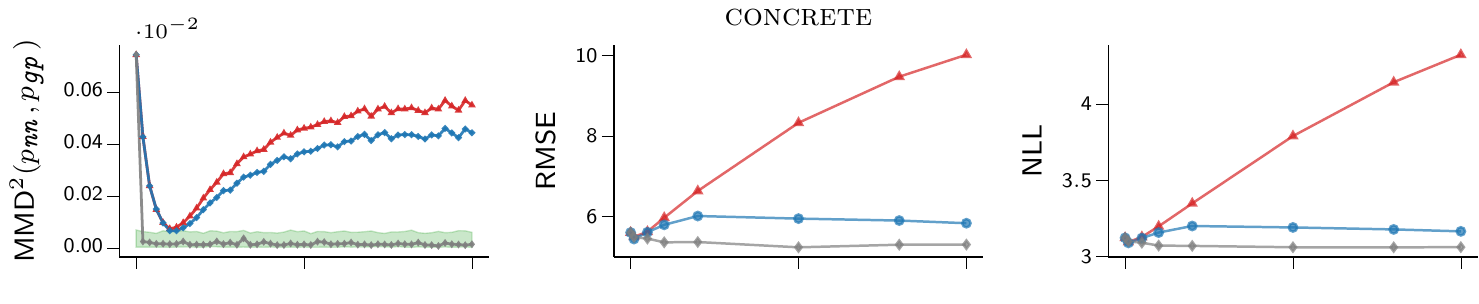}\\
	\vskip -0.1in
	\includegraphics{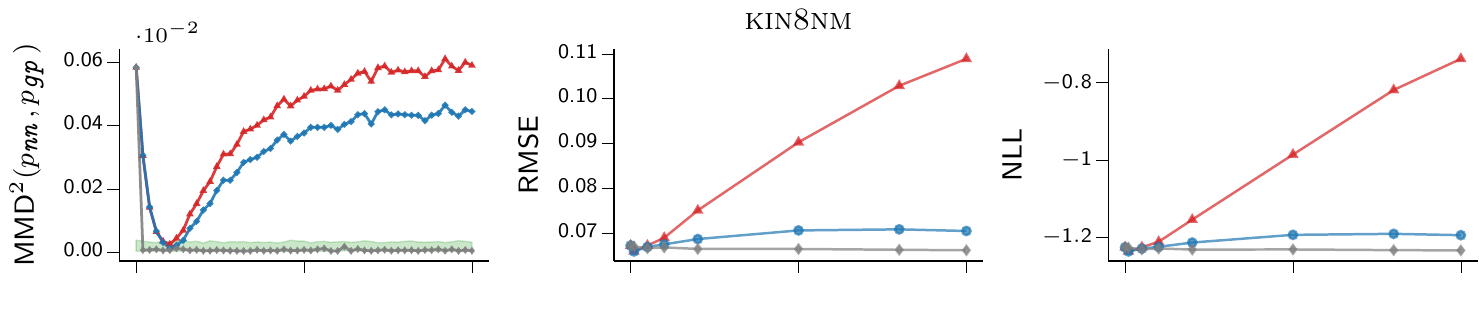}\\
	\vskip -0.1in
	\includegraphics{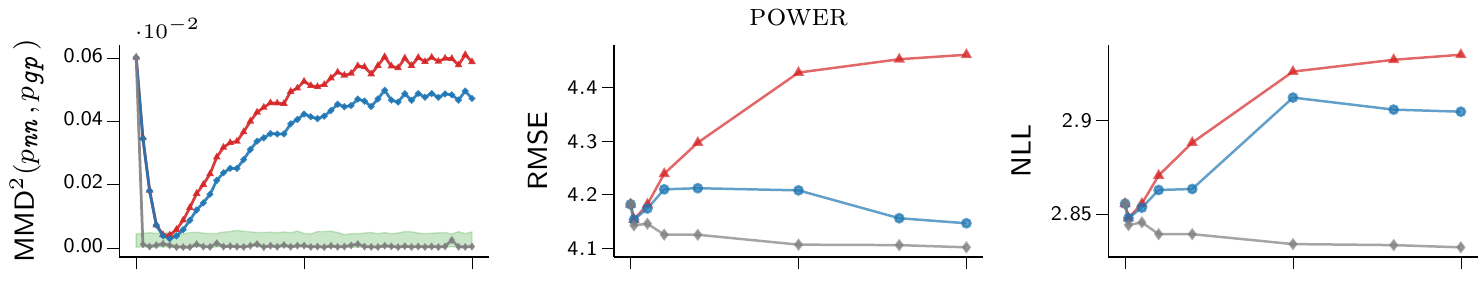}\\
	\vskip -0.1in
	\includegraphics{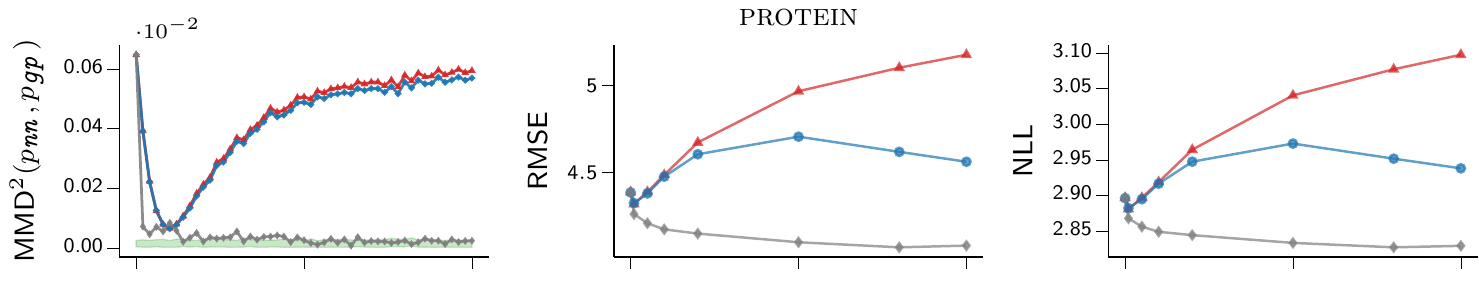}\\
	\vskip -0.1in
	\includegraphics{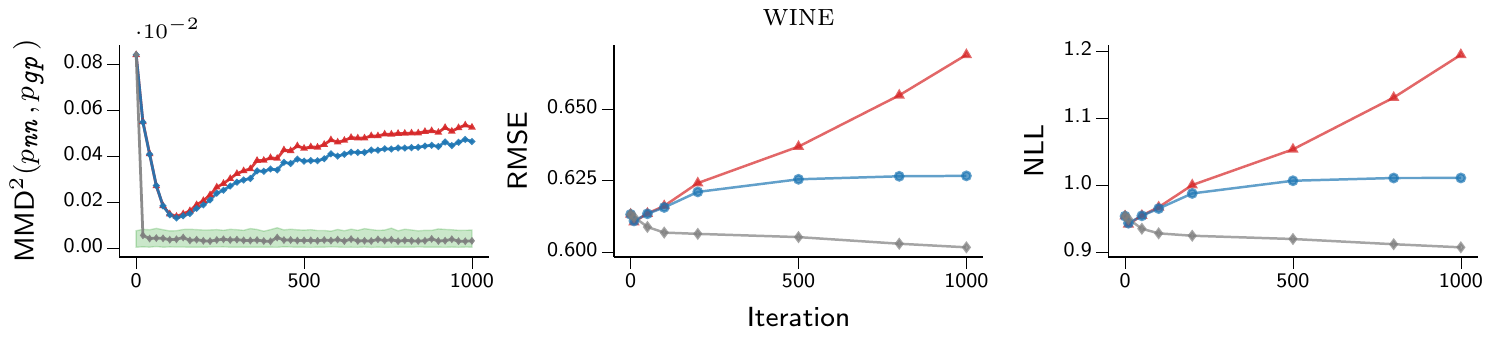}\\
	\vskip 0.1in
	\definecolor{kl}{rgb}{0.8359375,0.15234375,0.15625}
\definecolor{ssge}{rgb}{0.12,0.47,0.71}
\definecolor{wasserstein}{rgb}{0.49609375,0.49609375,0.49609375}
\begin{tabular}{lll}  {
    \protect\tikz[baseline=-1ex]\protect\draw[thick, color=kl, fill=kl, mark=triangle,  mark size=1.5pt, line width=1.5pt] plot[] (-.0, 0)--(.25,0)--(-.25,0);}  \textsf{KL-based optimization} &  {\protect\tikz[baseline=-1ex]\protect\draw[thick, color=ssge, fill=ssge, mark=*,  mark size=1.5pt, line width=1.5pt] plot[] (-.0, 0)--(.25,0)--(-.25,0);}  \textsf{KL-based + SSGE optimization} &  {\protect\tikz[baseline=-1ex]\protect\draw[thick, color=wasserstein, fill=wasserstein, mark=diamond,  mark size=1.5pt, line width=1.5pt] plot[] (-.0, 0)--(.25,0)--(-.25,0);}  \textsf{Wasserstein-based optimization}  \end{tabular}
	{\color{blue}
		\caption{Comparison between KL-based and Wasserstein-based optimization.
			The green shaded area is for calibration and denotes the difference between the squared \gls{MMD} of the target \gls{GP} to itself and to another \gls{GP} with a doubled lengthscale.
			\label{fig:compare_kldiv}}
	}
\end{figure}

\subsection{Wasserstein distance vs KL divergence} \label{ssec:wasserstein_vs_kl}

The \gls{KL} divergence is a popular criterion to measure the similarity between two distributions.
In our context, the \gls{KL} divergence could be used as follows:
\begin{align}
	\KL{p_\sub{nn}}{p_\sub{gp}} = - \int p_\sub{nn}(\mbf ; \mbpsi) \log{p_\sub{gp}(\mbf)} \d\mbf +
	\underbrace{\int p_\sub{nn}(\mbf ; \mbpsi) \log{p_\sub{nn}(\mbf;\mbpsi)} \d\mbf}_{\text{Entropy (intractable)}},
\end{align}
This is the form considered by \cite{Flam2017}, which propose to minimize the KL divergence between samples of a \gls{BNN} and a \gls{GP}.
This requires an empirical estimate of the entropy,
which is a challenging task for high-dimensional distributions \citep{Delattre2017}.
These issues were also reported by \cite{Flam2017}, where they propose an early stopping scheme to what is essentially an optimization of the cross-entropy term (i.e., -$\int p_\sub{nn}(\mbf ; \mbpsi) \log p_\sub{gp}(\mbf) \d\mbf$).
	{
		Instead of computing the entropy, another approach is to estimate its gradient as required by optimization algorithms.
		This can be carried out by using any methods estimating the log density derivative function of an implicit distribution.
		For example, \cite{Sun2019} use the \gls{SSGE} \citep{ShiS018} to obtain an estimate of the gradient of the entropy.

		In our experiments, we have found that a scheme based on the Wasserstein distance converges more consistently without the need for additional heuristics.
		We demonstrate the convergence properties of our scheme against the KL-divergence based optimization with early stopping \cite{Flam2017} and \gls{SSGE} in \cref{fig:compare_kldiv}.
		In this experiment, following \cite{Matthews2018}, we additionally use the kernel two-sample test based on the \gls{MMD} \citep{GrettonBRSS12} as an alternative assesment of the similarity between \glspl{BNN} and \glspl{GP}.
		A detailed description of estimating this discrepancy and experimental settings are available in \cref{ssec:mmd_appendix}.
		As done by \cite{Matthews2018}, we use a target \gls{GP} prior with a characteristic lengthscale of $l = \sqrt{2D}$, where $D$ is the input dimensionality.
		We monitor the evolution of squared \gls{MMD} from the target \gls{GP} prior and performance metrics for the \uci datasets (test \gls{NLL} and \gls{RMSE}).
		The \gls{KL}-based approaches offer improvements for the first few iterations, before degrading the quality of the approximation despite using the \gls{SSGE} for estimating the entropy gradient.
		Our approach, instead, consistently improves the quality of the approximation to the desired prior.
		In the \cref{ssec:wasserstein_convergence} we include a complete account on the convergence of Wasserstein distance for all experiments that follow in the next section.
	}

\section{Experimental Evaluation} \label{sec:experiments}

We shall now evaluate whether our scheme offers any competitive advantage in comparison to standard choices of priors. This section is organized as follows: we first summarize the baselines considered in our experimental campaign in \cref{ssec:baselines}.
We then investigate the effect of functional priors on classic \uci benchmark datasets for regression in \cref{ssec:uci_regression} and classification in \cref{ssec:uci_classification}. Bayesian \glspl{CNN} are explored in \cref{ssec:convolutional-neural-networks}, where we also study the benefits of functional priors for handling out-of-distribution data.
We next compare against some well-established alternatives to determine prior parameters, such as cross-validation and empirical Bayes in \cref{ssec:empirical_bayes}.
We then perform experiments on active learning (\cref{ssec:active_learning}), where having good and calibrated estimates of uncertainty is critical to achieve fast convergence.
Finally, we conclude in \cref{ssec:map} with a non-Bayesian experiment:
we explore the effect of functional priors on \gls{MAP} estimates, demonstrating that our scheme can also be beneficial as a regularization term in a purely optimization-based setting.

\subsection{Baselines} \label{ssec:baselines}
{
	
	In the following experiments, we consider two fixed priors: (1) fixed Gaussian (\fg) prior, $\cN(0, 1)$; (2) fixed hierarchical (\fh) prior where the prior variance for each layer is sampled from an Inverse-Gamma distribution, $\Gamma^{-1}(1, 1)$ \citep{Springenberg2016}; and three \gls{GP}-induced \gls{NN} priors, namely: (3) \gls{GP}-induced Gaussian (\gpig) prior, (4) \gls{GP}-induced hierarchical (\gpih) prior, and (5) \gls{GP}-induced normalizing flow (\gpinf) prior.
	Since the computational cost of the \gpinf prior is high, we only consider this prior in some of the regression experiments.
	For hierarchical priors, we resample the prior variances using a Gibbs step every $100$ iterations.

	Considering the aforementioned settings, we compare \glspl{BNN} against Deep Ensemble \citep{lakshminarayanan2017simple}, arguably one of the state-of-the-art approaches for uncertainty estimation in deep learning \citep{ashukhapitfalls, ovadia2019can}.
	This non-Bayesian method combines solutions that maximize the predictive log-likelihood for multiple neural networks trained with different initializations.
	We employ an ensemble of 5 neural networks in all experiments.
	Following \cite{lakshminarayanan2017simple}, we use Adam optimizer \citep{jlb2015adam} to train the individual networks.
	Furthermore, we compare the results obtained by sampling from the posterior obtained with \gls{GP}-induced priors against ``tempered'' posteriors \citep{Wenzel2020} that use the \fg prior and temperature scaling;  we refer to this approach as \fgts.
	In our experiments, the weight decay coefficient for Deep Ensemble and the temperature value for the ``tempered'' posterior are tuned by cross-validation.

	Additionally, we benchmark our approach against the state-of-the-art variational inference method in function space \citep{Sun2019}, referred to as \fbnn.
	We also evaluate our methodology of imposing priors against 
	an empirical Bayes approach \citep{ImmerBFRK21}, namely \lagnn, which optimizes the prior based on an approximation of the marginal likelihood by means of the Laplace and \gls{GGN} approximations.
	See the \cref{sec:implementation_details} for implementation details and more detailed hyper-parameter settings.
	\Cref{tab:methods} presents an overview of the methods considered in the experiments.
}

\begin{table*}
	\centering
	\caption{Glossary of methods used in the experimental campaign. Here, $p(\mbf) = \textstyle{\int p(\mbf\g\mbw)\d p(\mbw)}$ denotes the induced prior over functions; $\Gamma^{-1}(\alpha, \beta)$ denotes the Inverse-Gamma distribution with shape $\alpha$, and rate $\beta$; $\cN\cF(\cT_{K})$ indicates a normalizing flow distribution constructed from a sequence of $K$ invertible transformations $\cT$; $\widehat \sigma^2$, and ($\widehat \alpha$, $\widehat \beta$) denote the optimized parameters for the \gpig and \gpih priors, respectively.
	$\widehat{\kappa}$ corresponds to optimized kernel parameters, while $\widehat\sigma^2_{\text{LA}}$ shows that the parameters are optimized on the Laplace approximation of the marginal likelihood.
	References are [{a}] for \citet{Wenzel2020}, [{b}] for \citet{Springenberg2016}, [{c}] for \citet{lakshminarayanan2017simple}, [{d}] for \citet{Sun2019} and, finally, [{e}] for \citet{ImmerBFRK21}.
	}
	\label{tab:methods}
	\tiny
	\definecolor{neal}{rgb}{0.49609375,0.49609375,0.49609375}
\definecolor{temperature}{rgb}{0.12109375,0.46484375,0.703125}
\definecolor{optim_gaussian}{rgb}{0.171875,0.625,0.171875}
\definecolor{scale}{rgb}{0.99609375,0.49609375,0.0546875}
\definecolor{optim_scale}{rgb}{0.8359375,0.15234375,0.15625}
\definecolor{optim_nflow}{rgb}{0.08984375,0.7421875,0.80859375}
\definecolor{ensemble}{rgb}{0.546875,0.3359375,0.29296875}
\definecolor{optim_fbnn}{rgb}{0.421875,0.31640625,0.78125}
\definecolor{optim_laplace}{rgb}{0.81640625,0.36328125,0.3671875}

\newcommand{\drawline}[1]{{\protect\tikz[baseline=-.7ex]\protect\draw[thick, color=#1, fill=#1, line width=1.5pt] plot[] (-.0, 0)--(.15,0)--(-.15,0);}}

\renewcommand{\tabcolsep}{.5ex}

\begin{tabular}{rl ccccc c r}
	\toprule
	                            &                                                            & \multicolumn{5}{c}{Priors}                    & Inference                                                                                                                                                                  \\
	\cmidrule{3-7}
	& Name                                                       & $p(\sigma^2)$                                 &               & $p(\mbw\g\sigma^2)$                        &               & $p(\mbf)$                           &                           & Reference                   \\
	\midrule
	(\drawline{neal})           & \gls{BNN} w/ Fixed Gaussian (\fg) prior                    & --                                            &               & $\cN(0, \sigma^2\mbI)$                     & $\rightarrow$ & \redquestionmark                    & \acrshort{SGHMC}          &                             \\
	(\drawline{temperature})    & \gls{BNN} w/ Fixed Gaussian prior and TS (\fgts)           & --                                            &               & $\cN(0, \sigma^2\mbI)$                     & $\rightarrow$ & \redquestionmark                    & Tempered \acrshort{SGHMC} & [{a}]                  \\ (\drawline{scale})          & \gls{BNN} w/ Fixed hierarchical (\fh) prior                & $\Gamma^{-1}(\alpha,\beta)$                   & $\rightarrow$ & $\cN(0, \sigma^2\mbI)$                     & $\rightarrow$ & \redquestionmark                    & \acrshort{SGHMC} + Gibbs  & [{b}]                  \\ \midrule
	(\drawline{ensemble})       & Deep ensemble                                              & --                                            &               & \redquestionmark                           &               & \redquestionmark                    & Ensemble                  & [{c}]                  \\ \midrule
	(\drawline{optim_fbnn})          & Functional \gls{BNN} w/ variational inference (f\gls{BNN}) & --                                            &               & --                                         &               & $\mathcal{GP}(0, \widehat{\kappa})$ & Variational inference     & [{d}]                  \\
	(\drawline{optim_laplace})          & \gls{BNN} w/ Laplace GGN approximation (\textsc{la-ggn})   & --                                            &               & $\cN(0, \widehat\sigma_{\text{LA}}^2\mbI)$ & $\rightarrow$ & \redquestionmark                    & Laplace approximation     & [{e}]                  \\
	\midrule
	(\drawline{optim_gaussian}) & \gls{BNN} w/ \gls{GP}-induced Gaussian (\gpig) prior       & --                                            &               & $\cN(0, \widehat\sigma^2\mbI)$             & $\leftarrow$  & $\mathcal{GP}(0, \kappa)$           & \acrshort{SGHMC}          & [{\textbf{This work}}] \\
	(\drawline{optim_scale})    & \gls{BNN} w/ \gls{GP}-induced hierarchical (\gpih) prior   & $\Gamma^{-1}(\widehat \alpha,\widehat \beta)$ & $\leftarrow$  & $\cN(0, \sigma^2\mbI)$                     & $\leftarrow$  & $\mathcal{GP}(0, \kappa)$           & \acrshort{SGHMC} + Gibbs  & [{\textbf{This work}}] \\
	(\drawline{optim_nflow})    & \gls{BNN} w/ \gls{GP}-induced norm. flow (\gpinf) prior    & --                                            &               & $\cN\cF(\cT_{K})$                          & $\leftarrow$  & $\mathcal{GP}(0, \kappa)$           & \acrshort{SGHMC}          & [{\textbf{This work}}] \\
	\bottomrule
\end{tabular}

\end{table*}

\subsection{UCI regression benchmark}  \label{ssec:uci_regression}
We start our evaluation on real-world data by using regression datasets from the \uci collection \citep{asuncion2007uci}.
Each dataset is randomly split into training and test sets, comprising of $90$\% and $10$\% of the data, respectively.
This splitting process is repeated $10$ times except for the \protein dataset, which uses $5$ splits.
We use a two-layer \gls{MLP} with $\mathrm{tanh}$ activation function, containing $100$ units for smaller datasets and $200$ units for the \protein dataset.
We use a mini-batch size of $32$ for both the \gls{SGHMC} sampler and the Adam optimizer for Deep Ensemble.

We map a target hierarchical-GP prior to \gpig, \gpih, and \gpinf priors using our proposed Wasserstein optimization scheme with a mini-batch size of $N_s=128$.
We use an RBF kernel with dimension-wise lengthscales, also known as \gls{ARD} \citep{mackay1996bayesian}.
Hyper-priors $\mathrm{LogNormal}(\log\sqrt{2D}, 1)$ and $\mathrm{LogNormal}(0.1, 1)$ are placed on the lengthscales $l$ and the variance $\alpha^2$, respectively.
Here, $D$ is the number of input dimensions.
We use measurement sets having a size of $N_{\mathcal{M}}=100$, which include $70$\% random training samples and $30$\% uniformly random points from the input domain.

\begin{figure}[t]
	\centering
	\setlength\figurewidth{.16\textwidth}
	\setlength\figureheight{.25\textwidth}
	\scriptsize
	\hspace{-2ex}
	\begin{subfigure}[t]{0.9\textwidth}
		\includegraphics{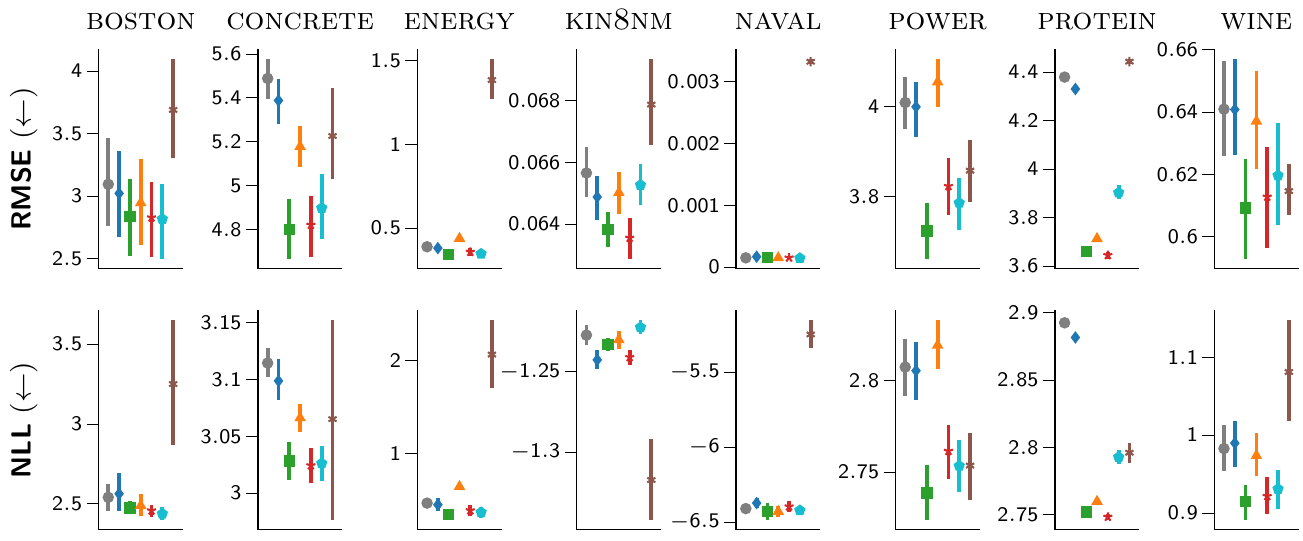}
	\end{subfigure}
	\begin{subfigure}[t]{0.1\textwidth}
		\includegraphics{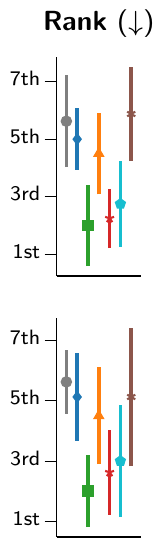}
	\end{subfigure} \\
	\definecolor{neal}{rgb}{0.49609375,0.49609375,0.49609375}
\definecolor{neal_temperature}{rgb}{0.12109375,0.46484375,0.703125}
\definecolor{optim_gaussian}{rgb}{0.171875,0.625,0.171875}
\definecolor{scale}{rgb}{0.99609375,0.49609375,0.0546875}
\definecolor{optim_scale}{rgb}{0.8359375,0.15234375,0.15625}
\definecolor{optim_nflow}{rgb}{0.08984375,0.7421875,0.80859375}
\definecolor{ensemble}{rgb}{0.546875,0.3359375,0.29296875}
\begin{tabular}{llll}
	{\protect\tikz[baseline=-1ex]\protect\draw[thick, color=neal, fill=neal, mark=*, mark size=1.2pt, line width=0.9pt] plot[] (-.0, 0)--(.25,0)--(-.25,0);} \fg prior &	{\protect\tikz[baseline=-1ex]\protect\draw[thick, color=neal_temperature, fill=neal_temperature, mark=diamond, mark size=1.2pt, line width=0.9pt] plot[] (-.0, 0)--(.25,0)--(-.25,0);} \fgts &	{\protect\tikz[baseline=-1ex]\protect\draw[thick, color=optim_gaussian, fill=optim_gaussian, mark=square, mark size=1.2pt, line width=0.9pt] plot[] (-.0, 0)--(.25,0)--(-.25,0);} \gpig prior \textbf{(ours)}\\ {\protect\tikz[baseline=-1ex]\protect\draw[thick, color=scale, fill=scale, mark=triangle, mark size=1.2pt, line width=0.9pt] plot[] (-.0, 0)--(.25,0)--(-.25,0);} \fh prior &	{\protect\tikz[baseline=-1ex]\protect\draw[thick, color=optim_scale, fill=optim_scale, mark=star, mark size=1.2pt, line width=0.9pt] plot[] (-.0, 0)--(.25,0)--(-.25,0);} \gpih prior \textbf{(ours)}&	{\protect\tikz[baseline=-1ex]\protect\draw[thick, color=optim_nflow, fill=optim_nflow, mark=pentagon, mark size=1.2pt, line width=0.9pt] plot[] (-.0, 0)--(.25,0)--(-.25,0);} \gpinf prior \textbf{(ours)}&	{\protect\tikz[baseline=-1ex]\protect\draw[thick, color=ensemble, fill=ensemble, mark=asterisk, mark size=1.2pt, line width=0.9pt] plot[] (-.0, 0)--(.25,0)--(-.25,0);} Deep Ensemble
\end{tabular}
	\caption{\uci regression benchmark results. The dots and error bars represent the means and standard errors over the test splits, respectively. Average ranks are computed across datasets. \label{fig:uci_regression}}
\end{figure}

\begin{figure}[t]
	\centering
	\setlength\figurewidth{.26\textwidth}
	\setlength\figureheight{.32\textwidth}
	\scriptsize
	\includegraphics{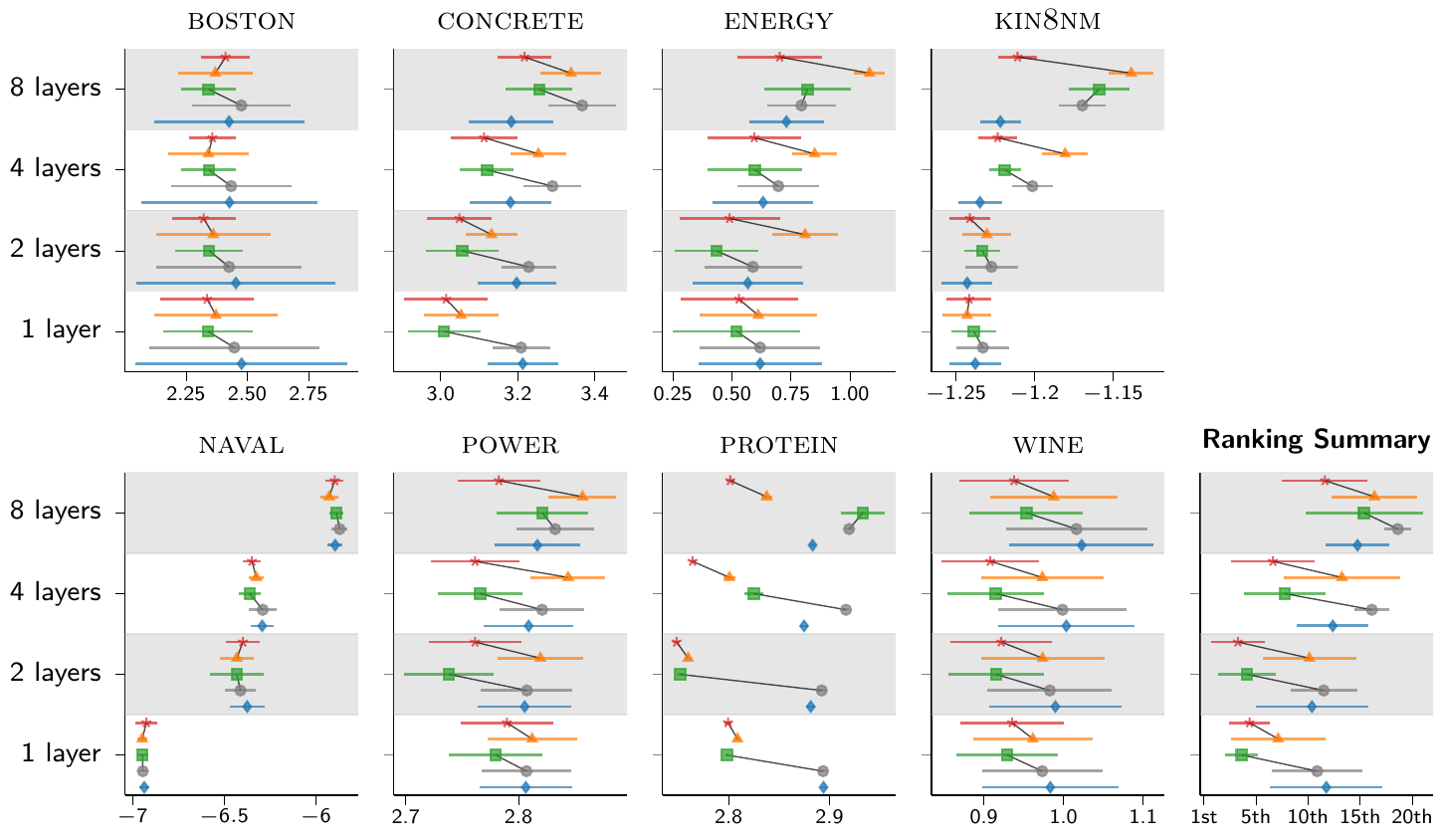}
	\vskip 0.1in
	\centering
	\definecolor{neal_temperature}{rgb}{0.12109375,0.46484375,0.703125}
\definecolor{neal}{rgb}{0.49609375,0.49609375,0.49609375}
\definecolor{optim_gaussian}{rgb}{0.171875,0.625,0.171875}
\definecolor{scale}{rgb}{0.99609375,0.49609375,0.0546875}
\definecolor{optim_scale}{rgb}{0.8359375,0.15234375,0.15625}
\begin{tabular}{lll}
	{\protect\tikz[baseline=-1ex]\protect\draw[thick, color=neal_temperature, fill=neal_temperature, mark=diamond, mark size=1.8pt, line width=1.3pt] plot[] (-.0, 0)--(.25,0)--(-.25,0);} \fgts & 	{\protect\tikz[baseline=-1ex]\protect\draw[thick, color=neal, fill=neal, mark=*, mark size=1.8pt, line width=1.3pt] plot[] (-.0, 0)--(.25,0)--(-.25,0);} \fg prior & 	{\protect\tikz[baseline=-1ex]\protect\draw[thick, color=optim_gaussian, fill=optim_gaussian, mark=square, mark size=1.8pt, line width=1.3pt] plot[] (-.0, 0)--(.25,0)--(-.25,0);} \gpig prior \textbf{(ours)} \\ 	{\protect\tikz[baseline=-1ex]\protect\draw[thick, color=scale, fill=scale, mark=triangle, mark size=1.8pt, line width=1.3pt] plot[] (-.0, 0)--(.25,0)--(-.25,0);} \fh prior & 	{\protect\tikz[baseline=-1ex]\protect\draw[thick, color=optim_scale, fill=optim_scale, mark=star, mark size=1.8pt, line width=1.3pt] plot[] (-.0, 0)--(.25,0)--(-.25,0);} \gpih prior \textbf{(ours)}
\end{tabular}
	\caption{Ablation study on the test \gls{NLL} based on the \uci regression benchmark for different number of hidden layers of \gls{MLP}. Error bars represent one standard deviation. We connect the fixed and \gls{GP}-induced priors with a thin black line as an aid for easier comparison. Further to the left is better. \label{fig:uci_regression_depth_nll}}
\end{figure}
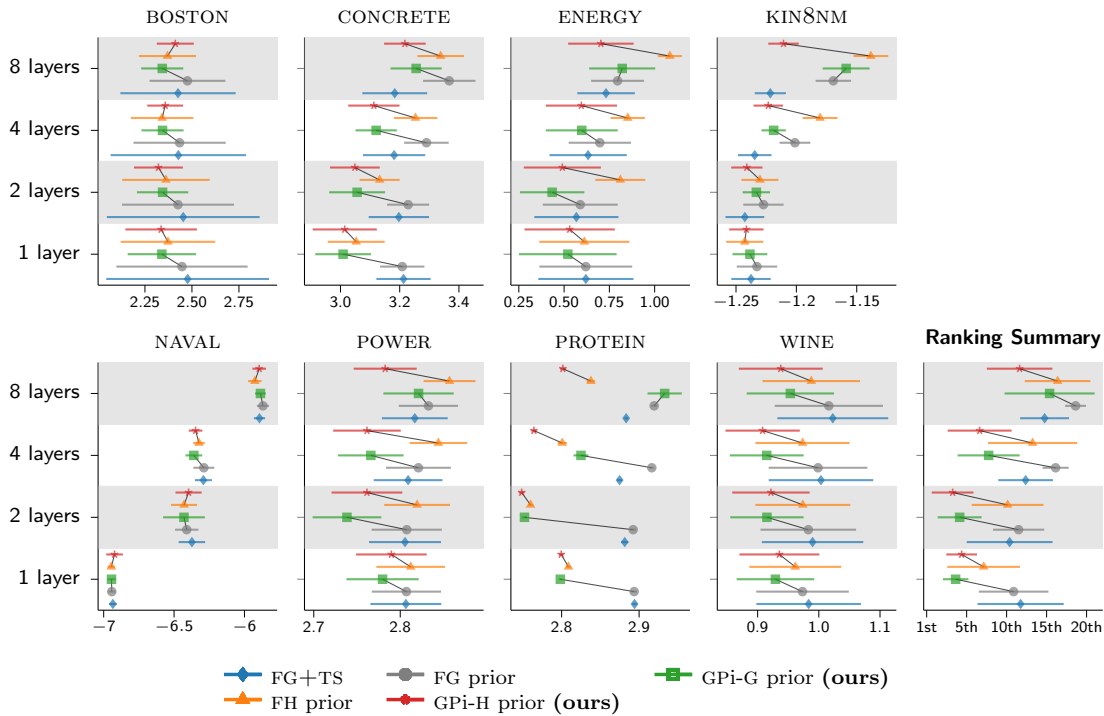

\cref{fig:uci_regression} illustrates the average test \gls{NLL} and \gls{RMSE}.
On the majority of datasets, our \gls{GP}-induced priors provide the best results.
They significantly outperform Deep Ensemble in terms of both \gls{RMSE} and \gls{NLL}, a metric that considers both uncertainty and accuracy.
We notice that tempering the posterior delivers only small improvements for the \fg prior.
Instead, by using the \gpig prior, the true posterior's predictive performance is improved significantly.

\paragraph{Ablation study on the model capacity.}
We further investigate the relation of the model capacity to the prior effect.
\cref{fig:uci_regression_depth_nll} illustrates the test \gls{NLL} on the \uci regression benchmark for different number of \gls{MLP} hidden layers.
For most datasets, the \gls{GP}-induced priors consistently outperform other approaches for all \gls{MLP} depths.
Remarkably, we observe that when increasing the model's capacity, the effect of temperature scaling becomes more prominent.
We argue that a tempered posterior is only beneficial for over-parameterized models, as evidenced by pathologically poor results for one-layer \glspl{MLP}.
We further elaborate on this hypothesis in \cref{ssec:convolutional-neural-networks} with much more complex models such as \glspl{CNN}.

\subsection{UCI classification benchmark} \label{ssec:uci_classification}
Next, we consider $7$ classification datasets from the \uci repository.
The chosen datasets have a wide variety in size, number of dimensions, and classes.
We use a two-layer \gls{MLP} with $\mathrm{tanh}$ activation function, containing $100$ units for small datasets (\eeg, \htru, \letter, and \magic), $200$ units for large datasets (\miniboo, \drive, and \mocap).
The experiments have been repeated for 10 random training/test splits.
We use a mini-batch size of $64$ examples for the \gls{SGHMC} sampler and the Adam optimizer.
Similarly to the previous experiment, we use a target hierarchical-\acrshort{GP} prior with hyper-priors for the lengthscales and the variance are $\mathrm{LogNormal}(\log\sqrt{2D}, 1)$ and $\mathrm{LogNormal}(\log 8, 0.3)$, respectively.
We use the same setup of the measurement set as used in the \uci regression experiments.

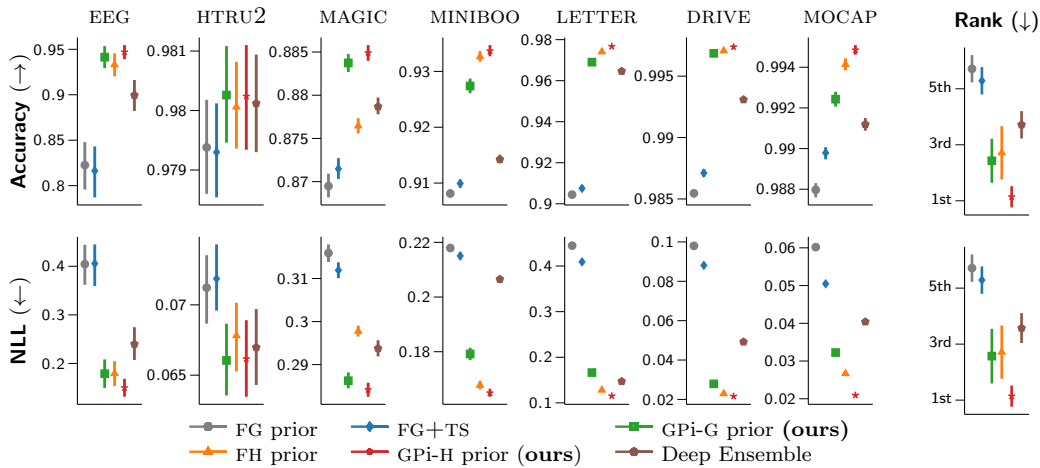
\begin{figure}
	\centering
	\setlength\figurewidth{.16\textwidth}
	\setlength\figureheight{.25\textwidth}
	\scriptsize
	\begin{subfigure}[t]{0.9\textwidth}
		\includegraphics{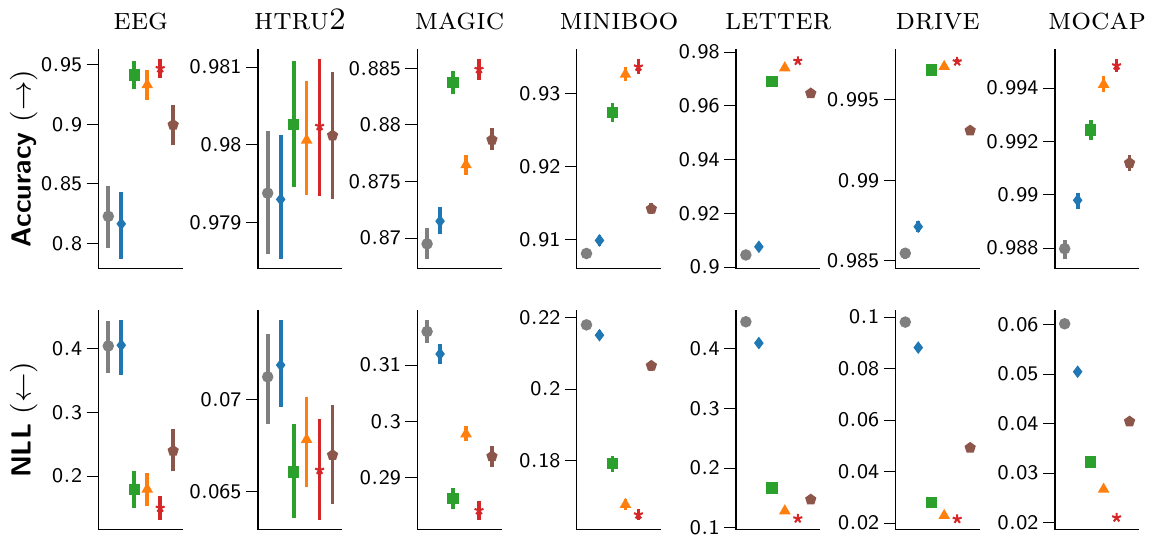}
	\end{subfigure}
	\hspace{-14ex}
	\begin{subfigure}[t]{0.1\textwidth}
		\includegraphics{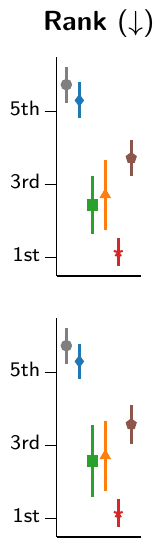}
	\end{subfigure} \\
	\definecolor{neal}{rgb}{0.49609375,0.49609375,0.49609375}
\definecolor{temperature}{rgb}{0.12109375,0.46484375,0.703125}
\definecolor{optim_gaussian}{rgb}{0.171875,0.625,0.171875}
\definecolor{scale}{rgb}{0.99609375,0.49609375,0.0546875}
\definecolor{optim_scale}{rgb}{0.8359375,0.15234375,0.15625}
\definecolor{ensemble}{rgb}{0.546875,0.3359375,0.29296875}
\begin{tabular}{llll}
	{\protect\tikz[baseline=-1ex]\protect\draw[thick, color=neal, fill=neal, mark=*, mark size=1.2pt, line width=0.9pt] plot[] (-.0, 0)--(.25,0)--(-.25,0);} \fg prior &	{\protect\tikz[baseline=-1ex]\protect\draw[thick, color=temperature, fill=temperature, mark=diamond, mark size=1.2pt, line width=0.9pt] plot[] (-.0, 0)--(.25,0)--(-.25,0);} \fgts &	{\protect\tikz[baseline=-1ex]\protect\draw[thick, color=optim_gaussian, fill=optim_gaussian, mark=square, mark size=1.2pt, line width=0.9pt] plot[] (-.0, 0)--(.25,0)--(-.25,0);} \gpig prior  \textbf{(ours)} \\	{\protect\tikz[baseline=-1ex]\protect\draw[thick, color=scale, fill=scale, mark=triangle, mark size=1.2pt, line width=0.9pt] plot[] (-.0, 0)--(.25,0)--(-.25,0);} \fh prior &	{\protect\tikz[baseline=-1ex]\protect\draw[thick, color=optim_scale, fill=optim_scale, mark=star, mark size=1.2pt, line width=0.9pt] plot[] (-.0, 0)--(.25,0)--(-.25,0);} \gpih prior (\textbf{ours}) &	{\protect\tikz[baseline=-1ex]\protect\draw[thick, color=ensemble, fill=ensemble, mark=pentagon, mark size=1.2pt, line width=0.9pt] plot[] (-.0, 0)--(.25,0)--(-.25,0);} Deep Ensemble
\end{tabular}
	\caption{\uci classification benchmark results. The dots and error bars represent the means and standard errors over the test splits, respectively. Average ranks are computed across datasets. \label{fig:uci_classification}}
\end{figure}

\cref{fig:uci_classification} reports the average test accuracy and \gls{NLL}.
The results for Deep Ensemble are significantly better than those of the \fg prior with and without using temperature scaling.
Similarly to the previous experiment, the \gpig prior outranks Deep Ensemble and is comparable with the \fh prior, which is a more flexible prior.
Once again, the \gpih prior consistently outperforms other priors across all datasets.

\subsection{Bayesian convolutional neural networks for image classification} \label{ssec:convolutional-neural-networks}
We proceed with the analysis of convolutional neural networks: we first analyze the kind of class priors that are induced by our strategy, and then we move to the \cifar experiment where we also discuss the cases of reduced and corrupted training data.

\paragraph{Analysis on the prior class labels.}
As already mentioned, \fg is the most popular prior for Bayesian \glspl{CNN} \citep{Wenzel2020, zhang2020csgmcmc, heek2019bayesian}.
This prior over parameters combined with a structured function form, such as a convolutional neural network, induces a structured prior distribution over functions.
However, as shown by \cite{Wenzel2020}, this is a poor functional prior because the sample function strongly favors a single class over the entire dataset.

\begin{figure}[t]
	\scriptsize

	\begin{subfigure}{1.02 \textwidth}
		\centering
		\includegraphics{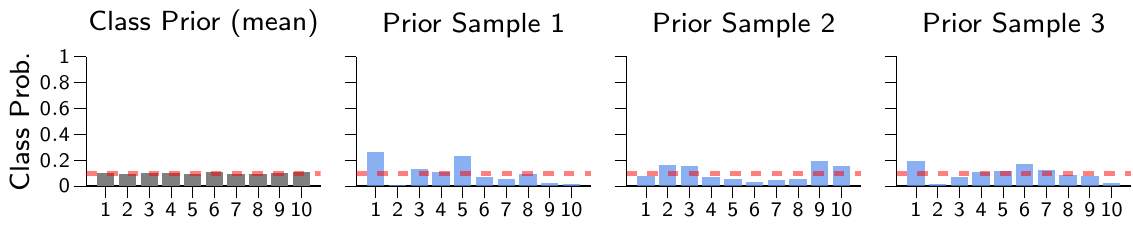}
		\vspace{-1.5ex}
		\caption{Target \gls{GP} prior. \label{fig:mnist_prior_gp}}
	\end{subfigure}

	\rulesep
	\vspace{0.5ex}

	\begin{subfigure}{1.02\textwidth}
		\centering
		\includegraphics{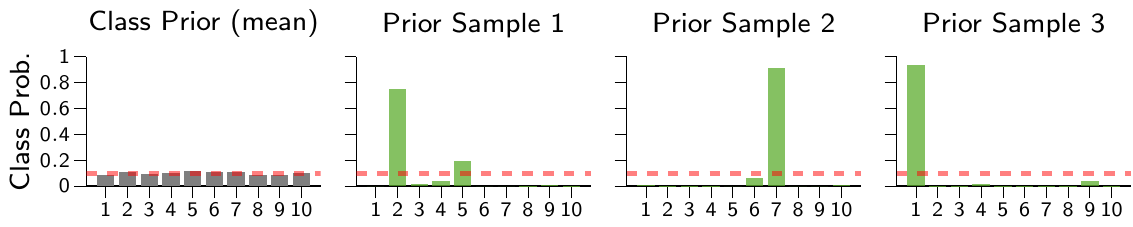}
		\vspace{-1.5ex}
		\caption{\gls{BNN} - \fg prior. \label{fig:mnist_prior_std_gauss}}
	\end{subfigure}\\

	\begin{subfigure}{1.02\textwidth}
		\centering
		\includegraphics{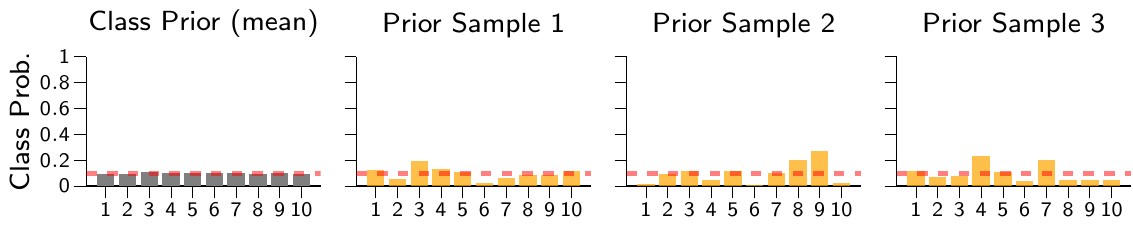}
		\vspace{-1.5ex}
		\caption{\gls{BNN} - \gpig prior. \label{fig:mnist_prior_optim_gauss}}
	\end{subfigure}

	\rulesep
	\vspace{0.5ex}

	\begin{subfigure}{1.02\textwidth}
		\centering
		\includegraphics{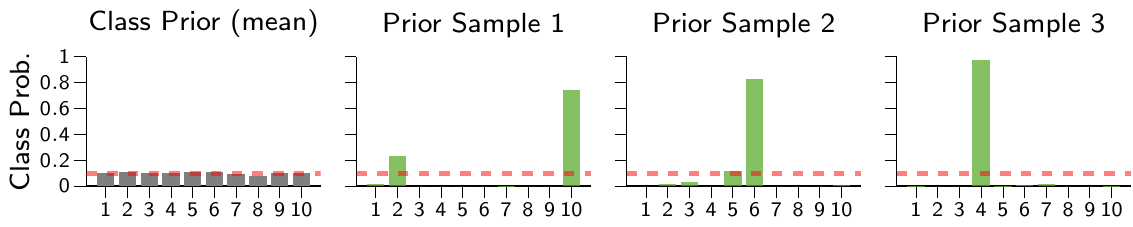}
		\vspace{-1.5ex}
		\caption{\gls{BNN} - \fh prior. \label{fig:mnist_prior_std_scale}}
	\end{subfigure}

	\begin{subfigure}{1.02\textwidth}
		\centering
		\includegraphics{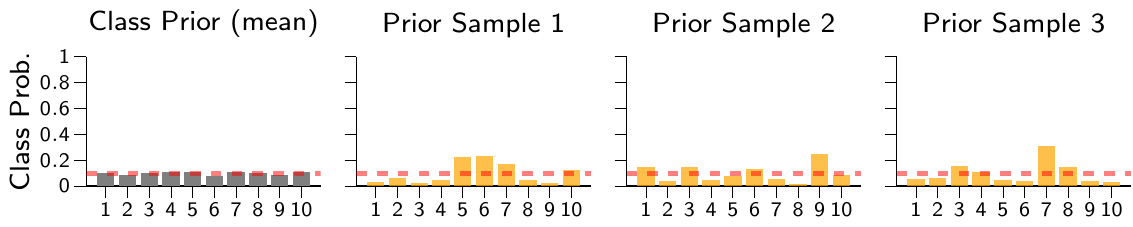}
		\vspace{-1.5ex}
		\caption{\gls{BNN} - \gpih prior. \label{fig:mnist_prior_optim_scale}}
	\end{subfigure}

	\caption{
		Average class probabilities over all training data of \mnist for three prior samples of parameters (three right columns),
		and prior distribution averaged over 200 samples of parameters (leftmost column).
		The \gpig and \gpih priors were obtained by mapping from a target \gls{GP} prior (top row) using our proposed method. \label{fig:mnist_predictive_priors}}
\end{figure}

We reproduce this finding for the \lenet model \citep{lecun1998gradient} on the \mnist dataset.
In particular, we draw three parameter samples from the \fg prior, and we observe the induced prior over classes for each parameter sample (see the three rightmost columns of \cref{fig:mnist_prior_std_gauss}). We also visualize the average prior distribution obtained from $200$ samples of parameters (see the leftmost column of \cref{fig:mnist_prior_std_gauss}).
Although the average prior distribution is fairly uniform, the distribution for each sample of parameters is highly concentrated on a single class.
As illustrated in \cref{fig:mnist_prior_std_scale}, the same problem happens for the \fh prior.

This pathology does not manifest in our approach, as a more sensible functional prior is imposed.
In particular, we choose a target \gls{GP} prior with an \gls{RBF} kernel having amplitude $\alpha=1$, such that the prior distribution for each \gls{GP} function sample is close to the uniform class distribution (\cref{fig:mnist_prior_gp}), and a lengthscale $l=256$ .
We then map this \gls{GP} prior to \gpig and \gpih priors by using our Wasserstein optimization scheme.
\cref{fig:mnist_prior_optim_gauss} and \cref{fig:mnist_prior_optim_scale} demonstrate that the resulting functional priors are more reasonable as evidenced by the uniformly-distributed prior distributions over all classes.

\setlength\figureheight{.19\textwidth}
\setlength\figurewidth{.26\textwidth}

\paragraph{Deep convolutional neural networks on CIFAR10}
We continue the experimental campaign on the \cifar benchmark \citep{krizhevsky2009learning} with a number of popular \gls{CNN} architectures: \lenet \citep{lecun1998gradient}, \vgg \citep{simonyan2014very} and \preresnet \citep{he2016identity}.
Regarding posterior inference with \gls{SGHMC}, after a burn-in phase of 10,000 iterations, we collect $200$ samples with 10,000 simulation steps in between.
For a fair comparison, we do not use techniques such as data augmentation or adversarial examples in any of the experiments.
Regarding the target hierarchical-\gls{GP} prior, we place a hyper-prior $\mathrm{LogNormal}(\log 8, 0.3)$ for variance, whereas the hyper-prior for length-scale is $\mathrm{LogNormal}(\log 512, 0.3)$.
We use a mini-batch size of $N_s=128$ and $N_{\mathcal{M}}=32$ measurement points sampled from the empirical distribution of the training data regarding prior optimization.

\cref{table:cnn_results} summarizes the results on the \cifar test set with respect to accuracy and \gls{NLL}.
These results demonstrate the effectiveness of the \gls{GP}-induced priors, as evidenced by the improvements in predictive performance when using \gpig and \gpih priors compared to using \fg and \fh priors, respectively.
Noticeably, the \gpih prior offers the best performance with 76.51\%, 87.03\%, and 88.20\% predictive accuracy on \lenet, \vgg, and \preresnet respectively.
We observe that for complex models (e.g., \preresnet and \vgg), \fg prior's results are improved by a large margin by tempering the posterior.
This is in line with the results showed by \cite{Wenzel2020}.
By contrast, in the case of \lenet, the predictive performance dramatically degraded when using temperature scaling.
In addition to the results in \cref{ssec:uci_regression}, this observation supports our conjecture that a ``tempered'' posterior is only useful for over-parameterized models.
Instead, by using \gls{GP}-induced priors, we consistently obtain the best results in most cases.

\begin{table}[t]
	\centering
	\footnotesize
	\caption{Results for different convolutional neural networks on the \cifar dataset (errors are $\pm 1$ standard error computed over 4 running times).}
	\setlength{\tabcolsep}{4pt}
\begin{tabular}{c  l  c c c}
    \toprule[1.25pt]

    \textbf{Architecture} & \textbf{Method}             & \textbf{Accuracy - \%} ($\uparrow$) & \textbf{NLL} ($\downarrow$)         & \\
    \midrule[1.25pt]
    \multirow{6}{*}{
        \lenet
    }
                          & Deep Ensemble               & 71.13 \tiny{$\pm$ 0.10}             & 0.8548 \tiny{$\pm$ 0.0010}            \\
    \cline{2-5} \rule{0pt}{2.5ex}
                          & \fg prior                   & 74.65 \tiny{$\pm$ 0.25}             & 0.7482 \tiny{$\pm$ 0.0025}            \\
                          & \fgts                       & 74.08 \tiny{$\pm$ 0.24}             & 0.7558 \tiny{$\pm$ 0.0024}            \\
                          & \gpig prior (\textbf{ours}) & 75.15 \tiny{$\pm$ 0.24}             & 0.7360 \tiny{$\pm$ 0.0024}            \\
    \cline{2-5} \rule{0pt}{2.5ex}
                          & \fh prior                   & 75.22 \tiny{$\pm$ 0.40}             & 0.7209 \tiny{$\pm$ 0.0040}            \\
                          & \gpih prior (\textbf{ours}) & \textbf{76.51 \tiny{$\pm$ 0.21}}    & \textbf{0.6952 \tiny{$\pm$ 0.0021}}   \\

    \midrule[1pt]
    \multirow{6}{*}{
        \preresnet
    }
                          & Deep Ensemble               & 87.77 \tiny{$\pm$ 0.03}             & 0.3927 \tiny{$\pm$ 0.0003}            \\
    \cline{2-5} \rule{0pt}{2.5ex}
                          & \fg prior                   & 85.34 \tiny{$\pm$ 0.13}             & 0.4975 \tiny{$\pm$ 0.0013}            \\
                          & \fgts                       & 87.70 \tiny{$\pm$ 0.11}             & 0.3956 \tiny{$\pm$ 0.0011}            \\
                          & \gpig prior (\textbf{ours}) & 86.86 \tiny{$\pm$ 0.27}             & 0.4286 \tiny{$\pm$ 0.0027}            \\
    \cline{2-5} \rule{0pt}{2.5ex}
                          & \fh prior                   & 87.26 \tiny{$\pm$ 0.09}             & 0.4086 \tiny{$\pm$ 0.0009}            \\
                          & \gpih prior (\textbf{ours}) & \textbf{88.20 \tiny{$\pm$ 0.07}}    & \textbf{0.3808 \tiny{$\pm$ 0.0007}}   \\

    \midrule[1pt]
    \multirow{6}{*}{
        \vgg
    }
                          & Deep Ensemble               & 81.96 \tiny{$\pm$ 0.33}             & 0.7759 \tiny{$\pm$ 0.0033}            \\
    \cline{2-5} \rule{0pt}{2.5ex}
                          & \fg prior                   & 81.47 \tiny{$\pm$ 0.33}             & 0.5808 \tiny{$\pm$ 0.0033}            \\
                          & \fgts                       & 82.25 \tiny{$\pm$ 0.15}             & 0.5398 \tiny{$\pm$ 0.0015}            \\
                          & \gpig prior (\textbf{ours}) & 83.34 \tiny{$\pm$ 0.53}             & 0.5176 \tiny{$\pm$ 0.0053}            \\
    \cline{2-5} \rule{0pt}{2.5ex}
                          & \fh prior                   & 86.03 \tiny{$\pm$ 0.20}             & 0.4345 \tiny{$\pm$ 0.0020}            \\
                          & \gpih prior (\textbf{ours}) & \textbf{87.03 \tiny{$\pm$ 0.07}}    & \textbf{0.4127 \tiny{$\pm$ 0.0007}}   \\
    \bottomrule[1.25pt]
\end{tabular}
	\vskip 0.15in
	\label{table:cnn_results}
\end{table}

\paragraph{Robustness to covariate shift.}
Covariate shift describes a situation where the test input data has a different distribution than the training data.
In this experiment, we evaluate the behavior of \gls{GP}-induced priors under such circumstances.
We also compare to Deep Ensemble, which is well-known for its robustness properties under covariate shift \citep{ovadia2019can}.

Using the protocol from \cite{ovadia2019can}, we train models on \cifar and then evaluate on the \cifarc dataset, which is generated by applying 16 different corruptions with 5 levels of intensity for each corruption \citep{hendrycks2018benchmarking}.
Our results are summarized in \cref{fig:preresnet_corrupted_data} (additional results are available in the appendix).
For \preresnet, there is a clear improvement in robustness to distribution shift by using the \gls{GP}-induced priors.
Remarkably, the \gpih prior performs best and outperforms Deep Ensemble at all corruption levels in terms of accuracy and \gls{NLL}.
Meanwhile, the \gls{NLL} results of \gls{SGHMC} are significantly better than those of Deep Ensemble.
We also notice that the \gpig prior offers considerable improvements in predictive performance compared to the \fg prior.

\begin{figure}
	\centering
	\setlength\figurewidth{.5\textwidth}
	\setlength\figureheight{.35\textwidth}
	\scriptsize
	\includegraphics{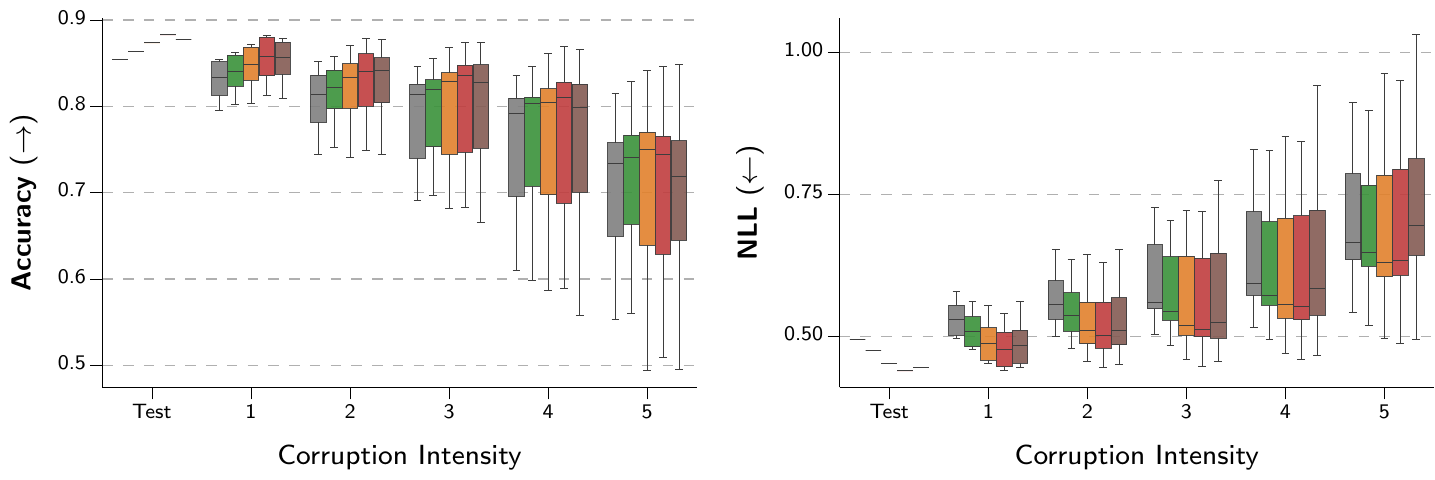}
	\vspace{1ex}
	\definecolor{neal}{rgb}{0.49609375,0.49609375,0.49609375}
\definecolor{optim_gaussian}{rgb}{0.171875,0.625,0.171875}
\definecolor{scale}{rgb}{0.99609375,0.49609375,0.0546875}
\definecolor{optim_scale}{rgb}{0.8359375,0.15234375,0.15625}
\definecolor{ensemble}{rgb}{0.546875,0.3359375,0.29296875}
\begin{tabular}{lllll}
{\protect\tikz[baseline=-1ex]\protect\draw[thick, color=neal, fill=neal, mark=square*, mark size=1.8pt, line width=1.8pt] plot[] (.0,-0.03);} \fg prior &
{\protect\tikz[baseline=-1ex]\protect\draw[thick, color=optim_gaussian, fill=optim_gaussian, mark=square*, mark size=1.8pt, line width=1.8pt] plot[] (.0,-0.03);} \gpig prior \textbf{(ours)} &
{\protect\tikz[baseline=-1ex]\protect\draw[thick, color=scale, fill=scale, mark=square*, mark size=1.8pt, line width=1.8pt] plot[] (.0,-0.03);} \fh prior &
{\protect\tikz[baseline=-1ex]\protect\draw[thick, color=optim_scale, fill=optim_scale, mark=square*, mark size=1.8pt, line width=1.8pt] plot[] (.0,-0.03);} \gpih prior \textbf{(ours)} &
{\protect\tikz[baseline=-1ex]\protect\draw[thick, color=ensemble, fill=ensemble, mark=square*, mark size=1.8pt, line width=1.8pt] plot[] (.0,-0.03);} Deep Ensemble
\end{tabular}
	\caption{Accuracy and \gls{NLL} on \cifarc at varying corruption severities.
		Here, we use the \preresnet architecture.
		For each method, we show the mean on the test set and the results on each level of corruption with a box plot.
		Boxes show the quartiles of performance over each corruption while the error bars indicate the minimum and maximum.
		\label{fig:preresnet_corrupted_data}}
\end{figure}
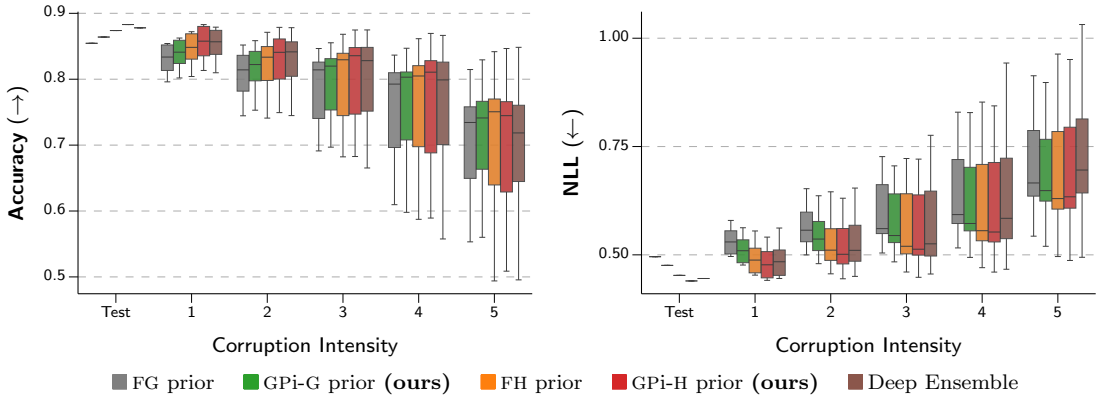

\setlength\figureheight{.38\textwidth}
\setlength\figurewidth{.3\textwidth}
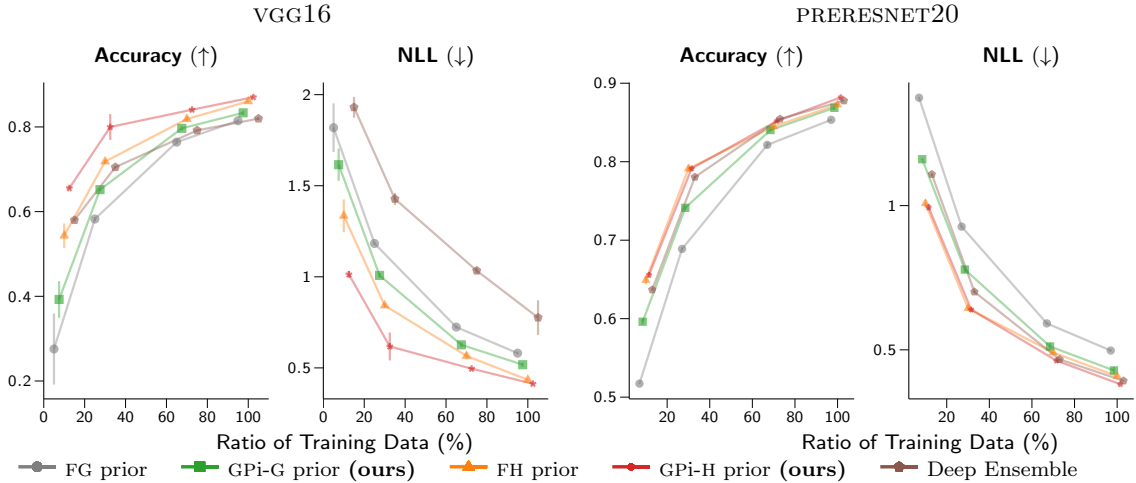
\begin{figure}[t]
	\scriptsize
	\captionsetup[subfigure]{oneside,margin={1cm,0cm},labelformat=empty,skip=3pt}
	\begin{subfigure}[t]{0.45\textwidth}
		\caption{\vgg \label{fig:vgg_small_data}}
		\includegraphics{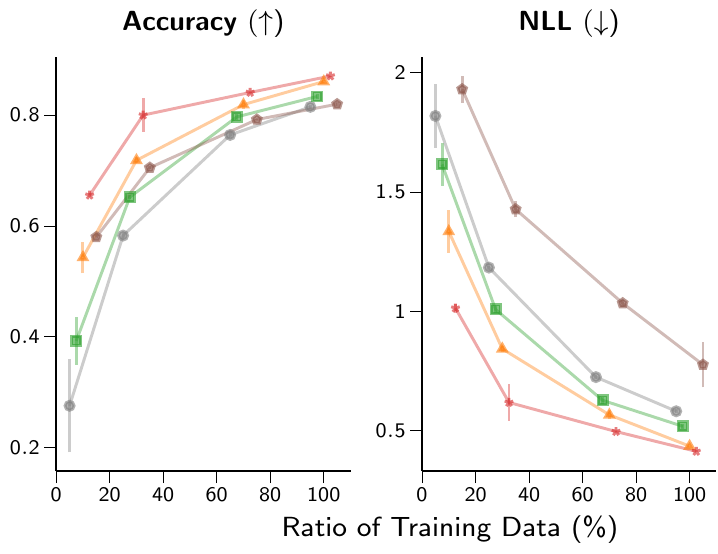}

		\end{subfigure}
	\hspace{6ex}
	\begin{subfigure}[t]{0.45\textwidth}
		\caption{\preresnet \label{fig:preresnet_small_data}}
		\includegraphics{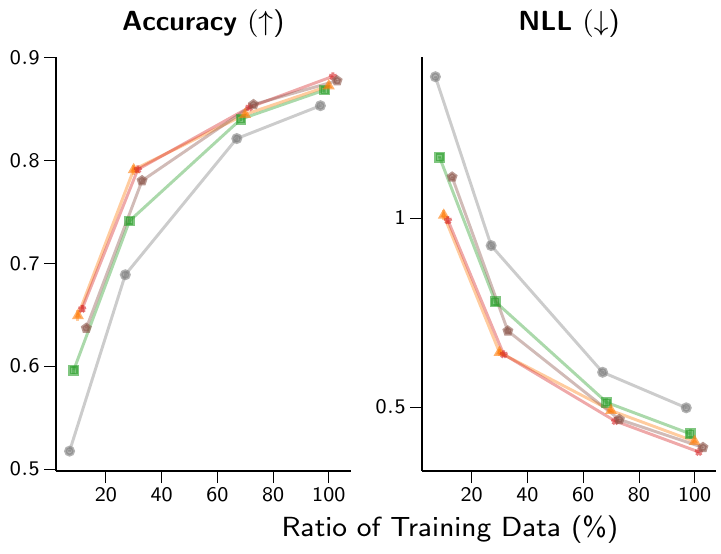}

	\end{subfigure}\vskip -0.03in
	\definecolor{neal}{rgb}{0.49609375,0.49609375,0.49609375}
\definecolor{optim_gaussian}{rgb}{0.171875,0.625,0.171875}
\definecolor{scale}{rgb}{0.99609375,0.49609375,0.0546875}
\definecolor{optim_scale}{rgb}{0.8359375,0.15234375,0.15625}
\definecolor{ensemble}{rgb}{0.546875,0.3359375,0.29296875}
\begin{tabular}{lllll}
{\protect\tikz[baseline=-1ex]\protect\draw[thick, color=neal, fill=neal, mark=*, mark size=1.3pt, line width=1.3pt] plot[] (-.0, 0)--(.25,0)--(-.25,0);} \fg prior &
{\protect\tikz[baseline=-1ex]\protect\draw[thick, color=optim_gaussian, fill=optim_gaussian, mark=square, mark size=1.3pt, line width=1.3pt] plot[] (-.0, 0)--(.25,0)--(-.25,0);} \gpig prior \textbf{(ours)} &
{\protect\tikz[baseline=-1ex]\protect\draw[thick, color=scale, fill=scale, mark=triangle, mark size=1.3pt, line width=1.3pt] plot[] (-.0, 0)--(.25,0)--(-.25,0);} \fh prior &
{\protect\tikz[baseline=-1ex]\protect\draw[thick, color=optim_scale, fill=optim_scale, mark=star, mark size=1.3pt, line width=1.3pt] plot[] (-.0, 0)--(.25,0)--(-.25,0);} \gpih prior \textbf{(ours)} &
{\protect\tikz[baseline=-1ex]\protect\draw[thick, color=ensemble, fill=ensemble, mark=pentagon, mark size=1.3pt, line width=1.3pt] plot[] (-.0, 0)--(.25,0)--(-.25,0);} Deep Ensemble
\end{tabular}
	\caption{Accuracy and \gls{NLL} on \cifar at varying the training set's size.
		The bars indicate one standard error. \label{fig:cifar_small_data}}
\end{figure}

\paragraph{Performance on small training data} \label{ssec:ood_test}
For small and high-dimensional datasets, the importance of choosing a sensible prior is more prominent because the prior's influence on the posterior is not overwhelmed by the likelihood.
To compare priors in this scenario, we use subsets of the \cifar dataset with different training set sizes, keeping the classes balanced.
\cref{fig:cifar_small_data} shows the accuracy and \gls{NLL} on the test set.
The \fg prior yields poor predictive performance in small training data cases.
Indeed, we observe that the \gpig prior performs much better than the \fg prior in all cases.
Besides, the \gpih prior offers superior predictive performance across all proportions of training/test data. These results again demonstrate the usefulness of the \gls{GP}-induced priors for the predictive performance of \glspl{BNN}.

\paragraph{Entropy analysis on out-of-distribition data.}

Next, we demonstrate with another experiment that the proposed \gls{GP}-based priors offer superior predictive uncertainties compared to competing approaches by considering the task of uncertainty estimation on out-of-distribution samples \citep{lakshminarayanan2017simple}.
Our choice of the target functional prior is reasonable for this type of task because, ideally, the predictive distribution should be uniform over the out-of-distribution classes--which results in maximum entropy--rather than being concentrated on a particular class.
Following the experimental protocol from \cite{Louizos2017}, we train \lenet on the standard \mnist training set, and estimate the entropy of the predictive distribution on both \mnist and \notmnist datasets\footnote{\notmnist dataset is available at  \url{http://yaroslavvb.blogspot.fr/2011/09/notmnist-dataset.html}.}.
The images in the \notmnist dataset have the same size as the \mnist, but represent other characters.
For posterior inference with \gls{SGHMC}, after a burn-in phase of 10,000 iterations, we draw $100$ samples with 10,000 iterations in between.
We also consider the ``tempered'' posterior with the \fg prior and Deep Ensemble as competitors.

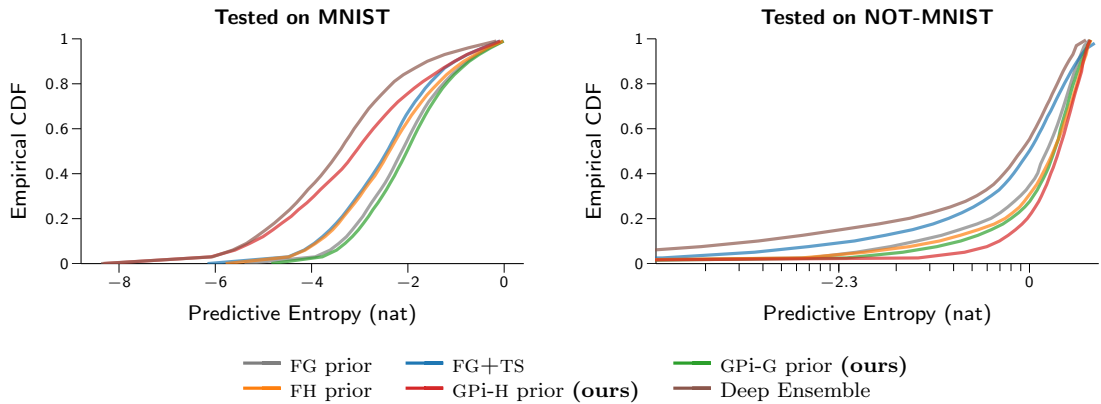
\begin{figure}[t]
	\centering
	\scriptsize
	\begin{subfigure}[t]{0.495\textwidth}
		\includegraphics{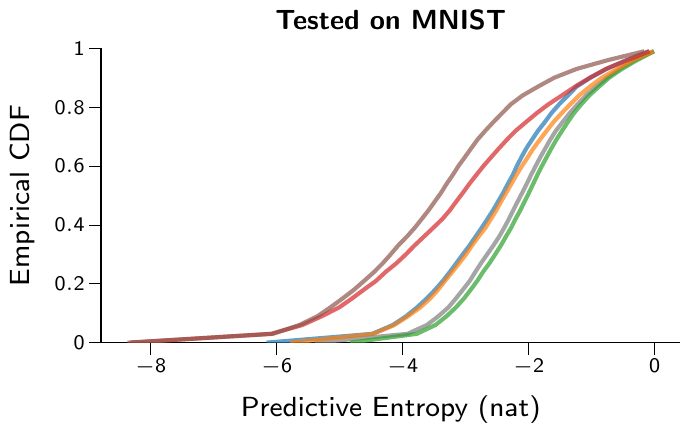}
	\end{subfigure}
	\begin{subfigure}[t]{0.495\textwidth}
		\includegraphics{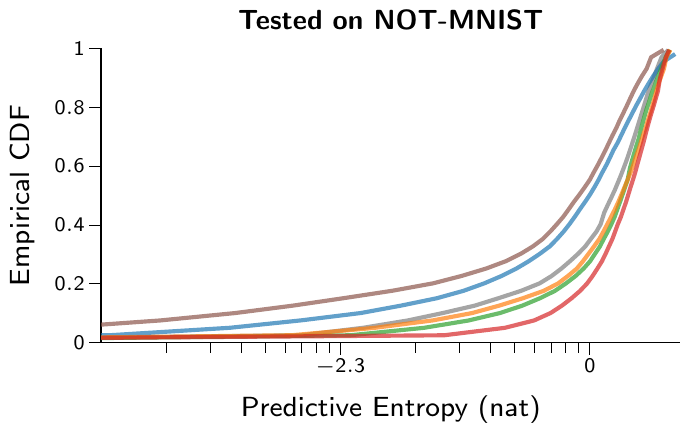}
	\end{subfigure}
	\vskip 0.1in
	\definecolor{fsp}{rgb}{0.49609375,0.49609375,0.49609375}
\definecolor{temperature}{rgb}{0.12109375,0.46484375,0.703125}
\definecolor{osp}{rgb}{0.171875,0.625,0.171875}
\definecolor{fhp}{rgb}{0.99609375,0.49609375,0.0546875}
\definecolor{ohp}{rgb}{0.8359375,0.15234375,0.15625}
\definecolor{ensemble}{rgb}{0.546875,0.3359375,0.29296875}
\begin{tabular}{lll}
	{\protect\tikz[baseline=-1ex]\protect\draw[thick, color=fsp, fill=fsp, mark=*, mark size=0, line width=1.2pt] plot[] (-.0, 0)--(.25,0)--(-.25,0);} \fg prior &	{\protect\tikz[baseline=-1ex]\protect\draw[thick, color=temperature, fill=temperature, mark=diamond, mark size=0, line width=1.2pt] plot[] (-.0, 0)--(.25,0)--(-.25,0);} \fgts &	{\protect\tikz[baseline=-1ex]\protect\draw[thick, color=osp, fill=osp, mark=square, mark size=0, line width=1.2pt] plot[] (-.0, 0)--(.25,0)--(-.25,0);} \gpig prior \textbf{(ours)} \\ {\protect\tikz[baseline=-1ex]\protect\draw[thick, color=fhp, fill=fhp, mark=triangle, mark size=0, line width=1.2pt] plot[] (-.0, 0)--(.25,0)--(-.25,0);} \fh prior &	{\protect\tikz[baseline=-1ex]\protect\draw[thick, color=ohp, fill=ohp, mark=star, mark size=0, line width=1.2pt] plot[] (-.0, 0)--(.25,0)--(-.25,0);} \gpih prior \textbf{(ours)} &	{\protect\tikz[baseline=-1ex]\protect\draw[thick, color=ensemble, fill=ensemble, mark=pentagon, mark size=0, line width=1.2pt] plot[] (-.0, 0)--(.25,0)--(-.25,0);} Deep Ensemble
\end{tabular}
	\caption{Cumulative distribution function plot of predictive entropies when the models trained on \mnist are tested on \mnist (left, the higher the better) and \notmnist (right, the lower the better).  \label{fig:mnist_ood_ecdf}}
\end{figure}

\cref{fig:mnist_ood_ecdf} shows the empirical CDF for the entropy of the predictive distributions on \mnist and \notmnist.
For the \notmnist dataset, the curves that are closer to the bottom right are preferable, as they indicate that the probability of predicting classes with a high confidence prediction is low.
In contrast,  the curves closer to the top left are better for the \mnist dataset.
As expected, we observe that the uncertainty estimates on out-of-distribution data for the \gls{GP}-induced priors are better than those obtained by the fixed priors.
In line with the results from \cite{Louizos2017}, Deep Ensemble tends to produce overconfident predictions on both in-distribution and out-of-distribution predictions. For tempered posteriors, we can interpret decreasing the temperature as artificially sharpening the posterior by overcounting the training data.
This is the reason why a tempered posterior tends to be overconfident.

\setlength\figureheight{.3\textwidth}
\setlength\figurewidth{.49\textwidth}

\subsection{Optimizing priors with data: cross-validation and empirical Bayes} \label{ssec:empirical_bayes}

\begin{figure}[t]
	\captionsetup[subfigure]{labelformat=empty}
	\centering
	\scriptsize
	\definecolor{color1}{rgb}{0.121,0.466,0.705}
	\definecolor{color0}{rgb}{0.66921568627451,0.518627450980392,0.385686274509804}
	\definecolor{color11}{rgb}{0.331764705882353,0.468235294117647,0.331764705882353}
	\definecolor{color2}{rgb}{1,0.498039215686275,0.0549019607843137}
	\definecolor{color3}{rgb}{0.172549019607843,0.627450980392157,0.172549019607843}
	\definecolor{color4}{rgb}{0.83921568627451,0.152941176470588,0.156862745098039}
	\setlength\figureheight{.30\textwidth}
	\setlength\figurewidth{.5\textwidth}
	\begin{subfigure}[t]{0.48\textwidth}
		\centering
		\includegraphics{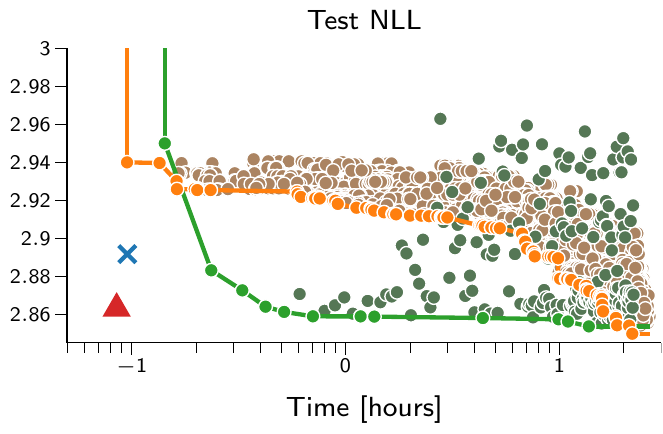}
	\end{subfigure}
	\begin{subfigure}[t]{0.48\textwidth}
		\centering
		\includegraphics{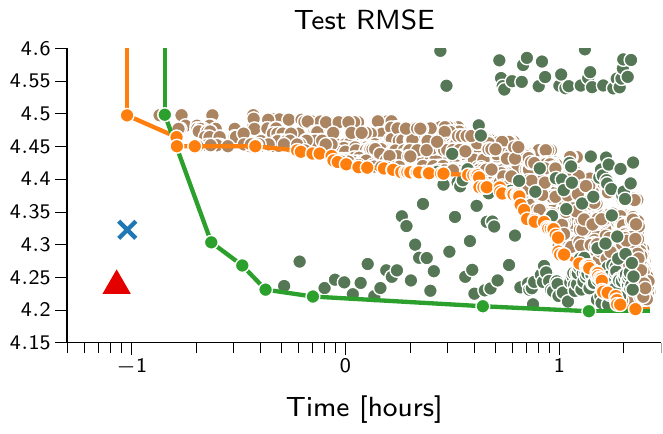}
	\end{subfigure}
	
	\caption{A timing comparison between imposing functional prior and cross-validation with
		either grid-search (\protect\tikz[baseline=-.65ex] \protect\draw[line width=5pt, draw=color2] plot (0, 0) --+(.5, 0);)
		or Bayesian optimization (\protect\tikz[baseline=-.65ex] \protect\draw[line width=5pt, draw=color3] plot (0, 0) --+(.5, 0);).
		In the plots, each \protect\tikz[baseline=-.65ex] \protect\draw[fill=gray,  draw=gray] plot[mark size=3, mark=*, only marks] (0, 0) --+(.2, 0) --+ (-.2, 0);  corresponds to a run of a single configuration, while
		\protect\tikz[baseline=-.65ex] \protect\draw[fill=gray,line width=2pt, draw=gray] plot[draw=white,  mark=*,] (0, 0) --+(.35, 0) --+ (-.35, 0); highlights the Pareto front of the cross validation procedure.
		The figure also reports the $\cN(0,1)$ prior as \protect\tikz[baseline=-.65ex] \protect\draw[fill=color1,  draw=color1] plot[mark size=5, mark=x, only marks] (0, 0) --+(.2, 0) --+ (-.2, 0);, while
		\protect\tikz[baseline=-.65ex] \protect\draw[fill=red!89.80!black,  draw=red!89.80!black] plot[mark size=5, mark=triangle*, only marks] (0, 0) --+(.2, 0) --+ (-.2, 0); is our proposal of using functional prior (\gpig).
	}
	\label{fig:CV}
	
\end{figure}

{
Although we advocate for functional priors over \glspl{BNN}, we acknowledge that a prior of this kind is essentially heuristic.
A potentially more useful prior might be discovered by traditional means such as \gls{CV} or by running an empirical Bayes procedure (a.k.a. type-II maximum likelihood), which maximizes the marginal likelihood $p(\cD;\mbpsi) = \int p(\cD\g\mbw)p(\mbw;\mbpsi)\d\mbw$ w.r.t. the prior parameters.
However, these methods present significant challenges:
(i) for \gls{CV}, 
the number of hyper-parameter combinations that needs to be explored becomes exponentially large as the complexity of the neural network grows, or as the exploration grid becomes more fine-grained.
(ii) for empirical Bayes, we need to compute the exact marginal likelihood, which is always intractable for \glspl{BNN}, thus requiring additional approximations like \gls{VI} or the Laplace approximation.
We next demonstrate these issues empirically.

\paragraph{Cross-Validation.}
We consider a simple case of a \gls{BNN} with one hidden layer only; by adopting the simple parameterization of \cref{ssec:parameterization}, we shall have four parameters to optimize in total (i.e. the weight and bias variances of the hidden and the output layer).
In \cref{fig:CV}, we demonstrate how our scheme behaves in comparison with a \gls{CV} strategy featuring a grid size of 9 (for a total of 6561 configurations).
To get results for the cross-validation procedure and to massively exploit all possible parallelization opportunities, we allocated a cloud platform with 16 server-grade machines, for a total of 512 computing cores and 64 maximum parallel jobs.
This required a bit more than one day, although the total CPU time approached 3 months.
While grid-based routines are widely adopted by practitioners for cross-validation, we acknowledge that there are more efficient alternatives.
To this extent, we also include Bayesian optimization \citep{marchuk1975,Snoek2012,fernando2014}, a classical method for black-box optimization which uses a Gaussian process as the surrogate function to be maximized (or minimized).
As expected, \gls{CV} indeed found marginally better configurations, but the amount of resources and time needed, even for such a small model, is orders of magnitude larger than what required by our scheme, making this procedure computationally infeasible for larger models, like \glspl{CNN}.
To put things into perspective, our Wasserstein-based functional prior could be run on a 4-core laptop in a reasonable time.}

\begin{figure}[t]
	\centering
	\scriptsize
	\includegraphics{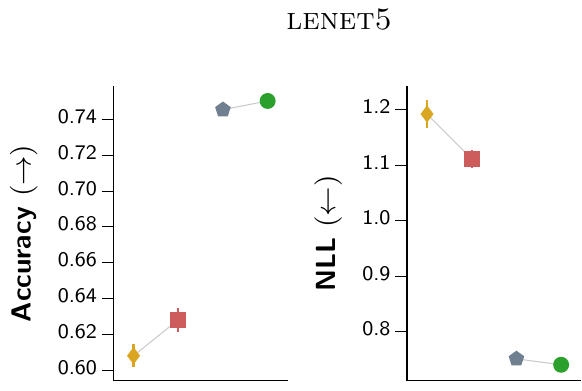}
	\hspace{3ex}
	\includegraphics{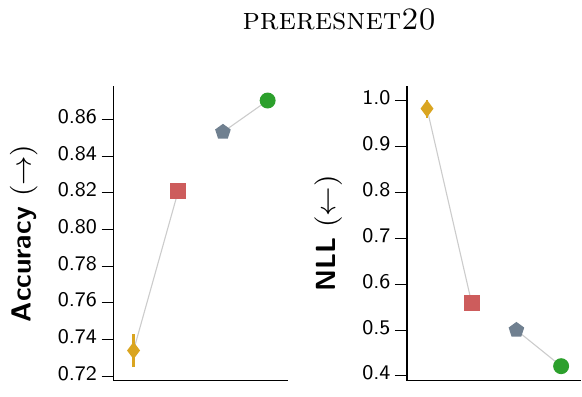}
	\vskip 0.07in
	\definecolor{std_laplace}{rgb}{0.85,0.65,0.13}
\definecolor{optim_gaussian}{rgb}{0.0703125,0.6328125,0.2265625}
\definecolor{wasserstein_laplace}{rgb}{0.51171875,0.0,0.48828125}
\definecolor{optim_fbnn}{rgb}{0.421875,0.31640625,0.78125}
\definecolor{neal}{rgb}{0.49609375,0.49609375,0.49609375}
\definecolor{optim_laplace}{rgb}{0.81640625,0.36328125,0.3671875}
\definecolor{sghmc_laplace}{rgb}{0.0,0.6953125,0.6640625} \begin{tabular}{ll} 
    {\protect\tikz[baseline=-1ex]\protect\draw[thick, color=std_laplace, fill=std_laplace, mark=diamond*, mark size=1.5pt, line width=0.9pt] plot[] (-.0, 0)--(.25,0)--(-.25,0);}  {\fg prior + \laplace-\ggn} &
    {\protect\tikz[baseline=-1ex]\protect\draw[thick, color=neal, fill=neal, mark=pentagon*, mark size=1.5pt, line width=0.9pt] plot[] (-.0, 0)--(.25,0)--(-.25,0);}  {\fg prior + \sghmc}
    \\
    {\protect\tikz[baseline=-1ex]\protect\draw[thick, color=optim_laplace, fill=optim_laplace, mark=square*, mark size=1.5pt, line width=0.9pt] plot[] (-.0, 0)--(.25,0)--(-.25,0);}  {\lagnn} &
    {\protect\tikz[baseline=-1ex]\protect\draw[thick, color=optim_gaussian, fill=optim_gaussian, mark=*, mark size=1.5pt, line width=0.9pt] plot[] (-.0, 0)--(.25,0)--(-.25,0);}  {\gpig prior + \sghmc (\textbf{ours})} 
    
    \end{tabular}
	\caption{Comparison with empirical Bayes and functional inference approaches on the \cifar dataset.
		A thin black line is used as an aid to see the performance improvement by using the optimized prior instead of the fixed prior (the standard Gaussian prior).
		The error bars indicate one standard deviation which is estimated by running 4 different random initializations.
		\label{fig:cifar_empirical_bayes}}
\end{figure}

{
\paragraph{Empirical Bayes.}
We now discuss state-of-the-art methods for empirical Bayes when using variational inference and Laplace approximation.
We demonstrate that our proposal outperforms these approaches through an extensive series of experiments on \uci regression and \cifar benchmarks.
More specifically, we evaluate our approach using \gls{SGHMC} with the \gpig prior and compare it against \fbnn, a method of functional variational inference \citep{Sun2019}
which imposes a \gls{GP} prior directly over the space of functions of \glspl{BNN}.
The hyper-parameters of the \gls{GP} prior for \fbnn are obtained by maximizing the marginal likelihood.
As in the original proposal of \fbnn, we only consider this baseline in experiments on regression datasets.
We consider a comparison with the Gaussian prior obtained by the empirical Bayes approach of \cite{ImmerBFRK21}.
This method uses the Laplace and \gls{GGN} methods to approximate the marginal likelihood, and referred to \lagnn.
Here, we use the same parameterization as for the \gpig prior where we optimize the variance of the Gaussian prior on the weights and biases of each layer individually.
The resulting prior obtained by this approach is denoted as \lamarglik.
The details of experimental settings are described in \cref{ssec:empirical_bayes_setup}.
In \cref{fig:uci_empirical_bayes}, we show the results of one-layer \gls{MLP} with tanh activation function on the \uci regression datasets.
Our approach using the \gls{SGHMC} sampler with the \gpig prior outperforms the baselines of functional inference on most datasets and across metrics.
Moreover, we find that our \gpig prior is consistently better than the \lamarglik prior when used together with \gls{SGHMC} for inference, denoted as ``\lamarglik + \sghmc''.
These observations are further highlighted in the experiments with Bayesian \glspl{CNN} on the \cifar benchmark.
As can be seen from \cref{fig:cifar_empirical_bayes}, thanks to using a good prior and a powerful sampling scheme for inference, our proposal consistently achieves the best results in all cases.
More comprehensive analyses with Bayesian \glspl{CNN} are available in \cref{ssec:empirical_bayes_appendix}.

\setlength\figureheight{.27\textwidth}
\setlength\figurewidth{.18\textwidth}
\begin{figure}[t]
	\centering
	\scriptsize
	\includegraphics{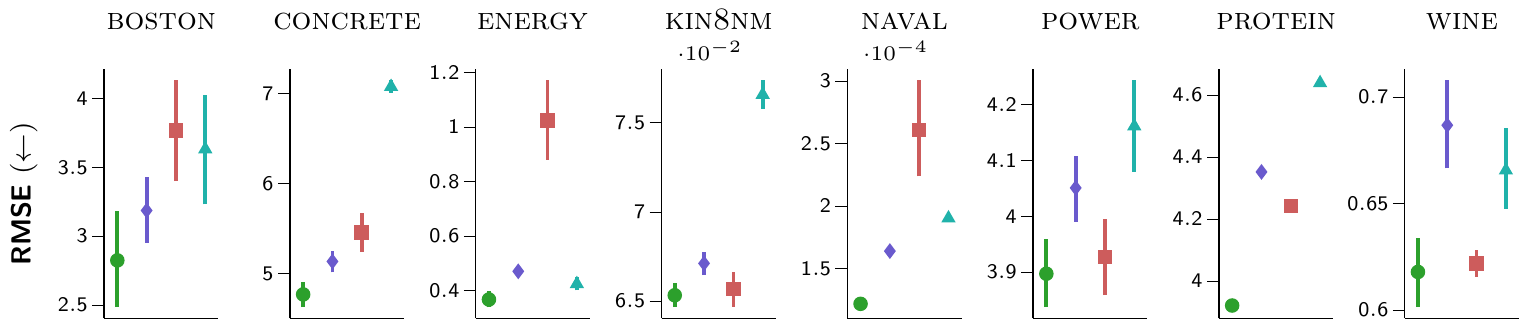}
	\includegraphics{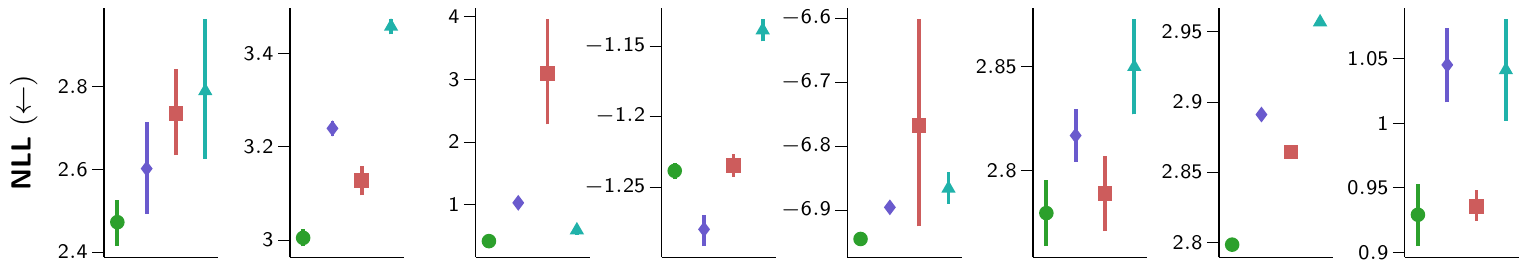}
	\vskip -0.01in
	\definecolor{optim_gaussian}{rgb}{0.0703125,0.6328125,0.2265625}
\definecolor{optim_fbnn}{rgb}{0.421875,0.31640625,0.78125}
\definecolor{optim_laplace}{rgb}{0.81640625,0.36328125,0.3671875}
\definecolor{sghmc_laplace}{rgb}{0.0,0.6953125,0.6640625} \begin{tabular}{cccc}
    {\protect\tikz[baseline=-1ex]\protect\draw[thick, color=optim_gaussian, fill=optim_gaussian, mark=*, mark size=1.2pt, line width=0.9pt] plot[] (-.0, 0)--(.25,0)--(-.25,0);}  {\gpig prior + \sghmc (\textbf{ours})} &  {\protect\tikz[baseline=-1ex]\protect\draw[thick, color=optim_fbnn, fill=optim_fbnn, mark=diamond*, mark size=1.2pt, line width=0.9pt] plot[] (-.0, 0)--(.25,0)--(-.25,0);}  {\fbnn} &  {\protect\tikz[baseline=-1ex]\protect\draw[thick, color=optim_laplace, fill=optim_laplace, mark=square*, mark size=1.2pt, line width=0.9pt] plot[] (-.0, 0)--(.25,0)--(-.25,0);}  {\lagnn} &  {\protect\tikz[baseline=-1ex]\protect\draw[thick, color=sghmc_laplace, fill=sghmc_laplace, mark=triangle*, mark size=1.2pt, line width=0.9pt] plot[] (-.0, 0)--(.25,0)--(-.25,0);}  {\lamarglik + \sghmc}
\end{tabular}
	\caption{Comparison with empirical Bayes and functional inference approaches on the \uci regression datasets.
		The dots and error bars represent the means and standard errors over the test splits, respectively. \label{fig:uci_empirical_bayes}}
\end{figure}

\par
From a more philosophical point of view, it is worth noting that cross-validating prior parameters, though perfectly legitimate, is not compatible with the classical Bayesian principles.
On the other hand, empirical Bayes is widely accepted as a framework to determine prior parameters in terms of a Bayesian context; nevertheless it still has to rely on part of the data.
In contrast to both of these alternatives, our procedure returns an appropriate prior \textit{without} having taken any data into consideration.

}

\setlength\figureheight{.30\textwidth}
\setlength\figurewidth{.22\textwidth}

\subsection{Active learning} \label{ssec:active_learning}

\begin{table}[t]

	\centering
	\resizebox{0.8\columnwidth}{!}{
		\begin{tabular}{l||cccc}
			\toprule
			\textbf{Data set} & \fg prior                  & \gpig prior \textbf{(ours)} & \fh prior         & \gpih prior \textbf{(ours)} \\
			\midrule
			\midrule
			\boston           & 3.199 $\pm$ 0.390          & 2.999 $\pm$ 0.382           & 3.030 $\pm$ 0.365 & \textbf{2.990} $\pm$ 0.384  \\
			\concrete         & 5.488 $\pm$ 0.218          & 5.036 $\pm$ 0.239           & 5.154 $\pm$ 0.251 & \textbf{4.919} $\pm$ 0.299  \\
			\energy           & \textbf{0.442} $\pm$ 0.041 & 0.461 $\pm$ 0.032           & 0.458 $\pm$ 0.050 & 0.446 $\pm$ 0.025           \\
			\kinm             & 0.069 $\pm$ 0.001          & 0.067 $\pm$ 0.001           & 0.068 $\pm$ 0.001 & \textbf{0.066} $\pm$ 0.001  \\
			\naval            & 0.000 $\pm$ 0.000          & 0.000 $\pm$ 0.000           & 0.000 $\pm$ 0.000 & 0.000 $\pm$ 0.000           \\
			\power            & 4.015 $\pm$ 0.059          & \textbf{3.834} $\pm$ 0.068  & 4.172 $\pm$ 0.051 & 3.851 $\pm$ 0.066           \\
			\protein          & 4.429 $\pm$ 0.016          & 4.036 $\pm$ 0.014           & 4.080 $\pm$ 0.018 & \textbf{3.993} $\pm$ 0.014  \\
			\wine             & 0.634 $\pm$ 0.013          & 0.617 $\pm$ 0.008           & 0.625 $\pm$ 0.010 & \textbf{0.612} $\pm$ 0.011  \\
			\bottomrule
		\end{tabular}
		}
	\vskip 0.15in
	\caption{Results for the active learning scenario. Average test RMSE evaluated at the last step of the iterative data gathering procedure. \label{table:active learning}}
	\end{table}

\setlength\figureheight{.24\textwidth}
\setlength\figurewidth{.28\textwidth}
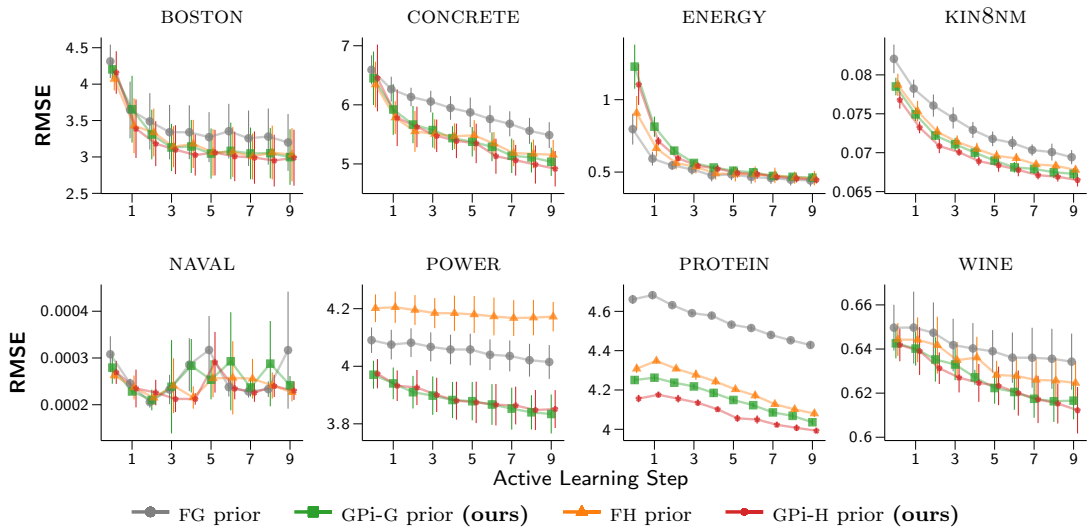
\begin{figure}[t]
	\centering
	\scriptsize
	\includegraphics{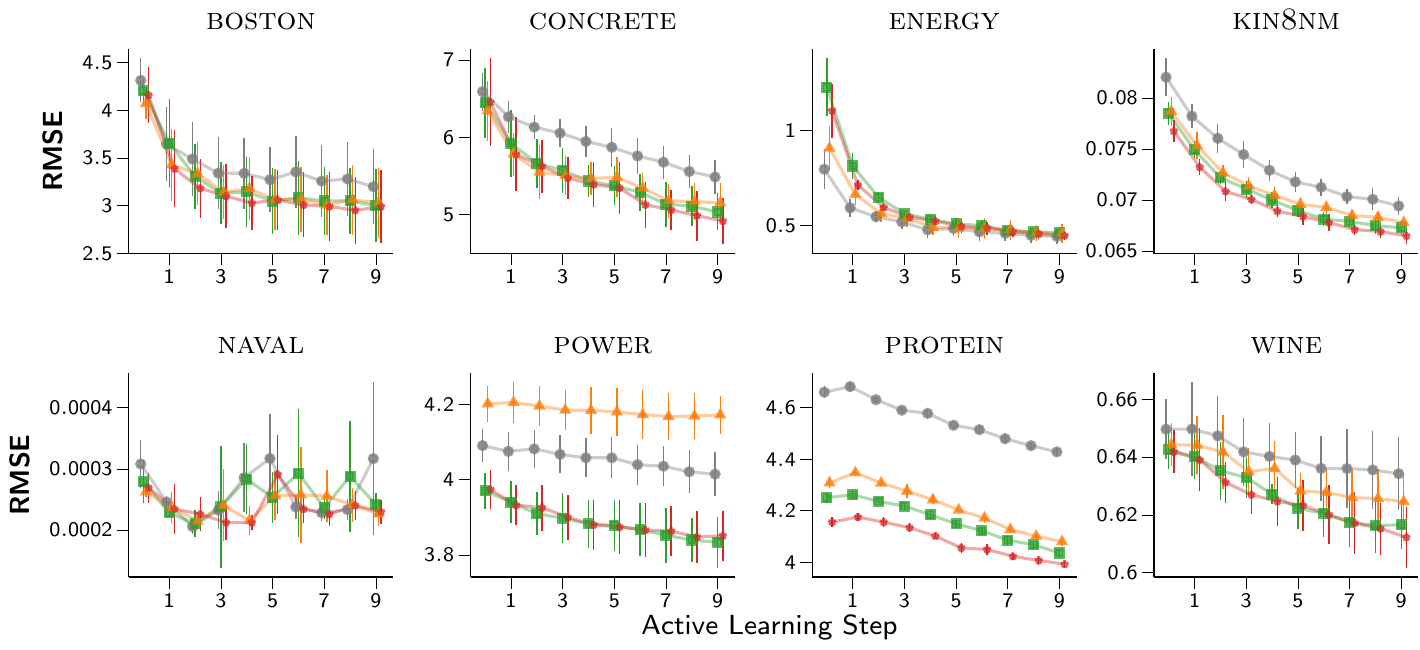}
	\definecolor{neal}{rgb}{0.49609375,0.49609375,0.49609375}
\definecolor{optim_gaussian}{rgb}{0.171875,0.625,0.171875}
\definecolor{scale}{rgb}{0.99609375,0.49609375,0.0546875}
\definecolor{optim_scale}{rgb}{0.8359375,0.15234375,0.15625}
\begin{tabular}{llll}
	{\protect\tikz[baseline=-1ex]\protect\draw[thick, color=neal, fill=neal, mark=*, mark size=1.3pt, line width=1.3pt] plot[] (-.0, 0)--(.25,0)--(-.25,0);} \fg prior & 	{\protect\tikz[baseline=-1ex]\protect\draw[thick, color=optim_gaussian, fill=optim_gaussian, mark=square, mark size=1.3pt, line width=1.3pt] plot[] (-.0, 0)--(.25,0)--(-.25,0);} \gpig prior \textbf{(ours)} & 	{\protect\tikz[baseline=-1ex]\protect\draw[thick, color=scale, fill=scale, mark=triangle, mark size=1.3pt, line width=1.3pt] plot[] (-.0, 0)--(.25,0)--(-.25,0);} \fh prior & 	{\protect\tikz[baseline=-1ex]\protect\draw[thick, color=optim_scale, fill=optim_scale, mark=star, mark size=1.3pt, line width=1.3pt] plot[] (-.0, 0)--(.25,0)--(-.25,0);} \gpih prior \textbf{(ours)} \end{tabular}
	\caption{The progressions of average test \gls{RMSE} and standard errors in the active learning experiment. \label{fig:active_learning}}
\end{figure}

We next perform a series of experiments within an active learning scenario \citep{settles2009active}.
In this type of task, it is crucial to produce accurate estimates of uncertainty to obtain good performance.
We use the same network architectures and datasets as used in the \uci regression benchmark.
We adopt the experimental setting of \cite{skafte2019reliable}, where each dataset is split into $20$\% train, $60$\% pool, and $20$\% test sets.
For each active learning step, we first train models and then estimate uncertainty for all data instances in the pool set.
To actively collect data from the pool set, we follow the information-based approach described by \cite{mackay1992information}.
More specifically, we choose the $n$ data points with the highest posterior entropy and add them to the training set.
Under the assumption of i.i.d. Gaussian noise, this is equivalent to choosing the unlabeled examples with the largest predictive variance \citep{houlsby2012collaborative}.
We define $n = 5\%$ of the initial size of the pool set. We use 10 active-learning steps and repeat each experiment 5 times per dataset on random training-test splits to compute standard errors.

\cref{fig:active_learning} shows the progressions of average test \gls{RMSE} during the data collection process.
We observe that, on most datasets (\concrete, \kinm, \power, \protein, and \wine), the \gpig and \gpih priors achieve faster learning than \fg and \fh priors, respectively.
For the other datasets, \fh prior is on par with \gpih, while \fg consistently results in the worse performance, except in one case (\energy).
We also report the average test \gls{RMSE} at the last step in \cref{table:active learning}.
These results show that the \gpih prior performs best, while the \gpig prior outperforms the \fg prior in most cases.

\subsection{Maximum-a-posteriori (MAP) estimation with GP-induced prior} \label{ssec:map}
\begin{figure}[t]
	\centering
	\scriptsize
	\includegraphics{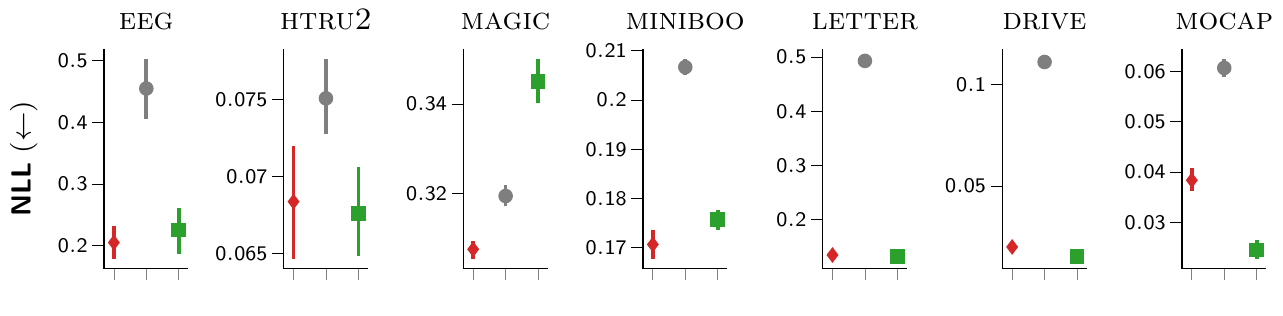}
	\includegraphics{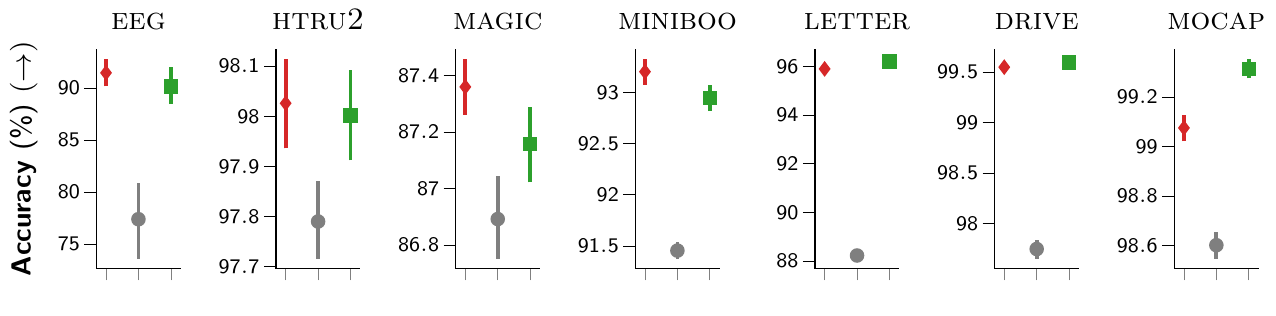}
	\definecolor{early_stopping}{rgb}{0.8359375,0.15234375,0.15625}
\definecolor{map_std}{rgb}{0.49609375,0.49609375,0.49609375}
\definecolor{map_optim}{rgb}{0.171875,0.625,0.171875} \begin{tabular}{ccc}  {\protect\tikz[baseline=-1ex]\protect\draw[thick, color=early_stopping, fill=early_stopping, mark=diamond, mark size=1.2pt, line width=0.9pt] plot[] (-.0, 0)--(.25,0)--(-.25,0);}  Early Stopping &  {\protect\tikz[baseline=-1ex]\protect\draw[thick, color=map_std, fill=map_std, mark=*, mark size=1.2pt, line width=0.9pt] plot[] (-.0, 0)--(.25,0)--(-.25,0);}  \acrshort{MAP} with \fg prior &  {\protect\tikz[baseline=-1ex]\protect\draw[thick, color=map_optim, fill=map_optim, mark=square, mark size=1.2pt, line width=0.9pt] plot[] (-.0, 0)--(.25,0)--(-.25,0);}  \acrshort{MAP} with \gpig prior (\textbf{ours})  \end{tabular}
	\caption{Comparison between early stopping and \gls{MAP} optimization with the \fg and \gpig priors on the \uci classification datasets.  \label{fig:uci_class_earlystop_map}}
\end{figure}

\begin{figure}[t]
	\centering
	\scriptsize
	\includegraphics{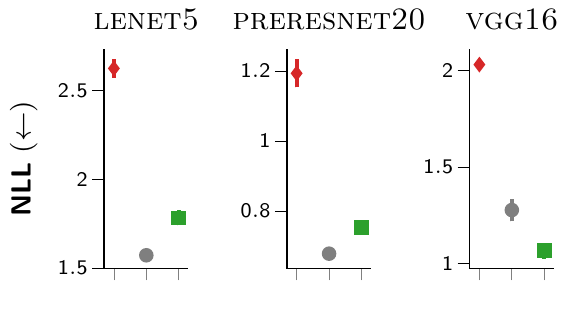}
	\hskip 0.5in
	\includegraphics{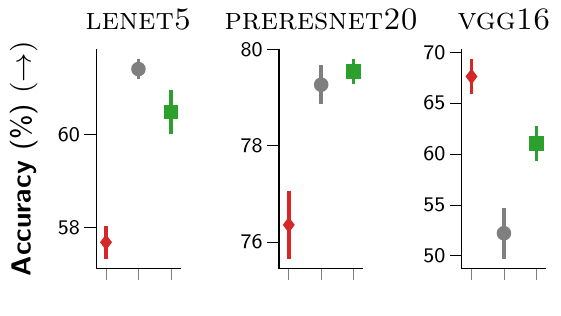}
	\definecolor{early_stopping}{rgb}{0.8359375,0.15234375,0.15625}
\definecolor{map_std}{rgb}{0.49609375,0.49609375,0.49609375}
\definecolor{map_optim}{rgb}{0.171875,0.625,0.171875} \begin{tabular}{ccc}  {\protect\tikz[baseline=-1ex]\protect\draw[thick, color=early_stopping, fill=early_stopping, mark=diamond, mark size=1.2pt, line width=0.9pt] plot[] (-.0, 0)--(.25,0)--(-.25,0);}  Early Stopping &  {\protect\tikz[baseline=-1ex]\protect\draw[thick, color=map_std, fill=map_std, mark=*, mark size=1.2pt, line width=0.9pt] plot[] (-.0, 0)--(.25,0)--(-.25,0);}  \acrshort{MAP} with \fg prior &  {\protect\tikz[baseline=-1ex]\protect\draw[thick, color=map_optim, fill=map_optim, mark=square, mark size=1.2pt, line width=0.9pt] plot[] (-.0, 0)--(.25,0)--(-.25,0);}  \acrshort{MAP} with \gpig prior (\textbf{ours})  \end{tabular}
	\caption{Comparison between early stopping and \gls{MAP} optimization with the \fg and \gpig priors for three different \gls{CNN} architectures on the \cifar dataset.  \label{fig:cnn_earlystop_map}}
\end{figure}

In the last experiment, we demonstrate that the \gpig prior is useful not only for Bayesian inference but also for \gls{MAP} estimation.
We investigate the impact of the \gpig priors obtained in the previous experiments and the \fg prior on the performance of \gls{MAP} estimation.
We additionally compare to \textit{early stopping}, which is a popular regularization method for neural networks.
Compared to early stopping, \gls{MAP} is a more principled regularization method even though early stopping should exhibit similar behavior to  \gls{MAP} regularization in some cases, such as those involving a quadratic error function \citep{rosasco2007early}.
Regarding the experimental setup, we train all networks for $150$ epochs using the Adam optimizer with a fixed learning rate $0.01$.
For early stopping, we stop training as soon as there is no improvement for $10$ consecutive epochs on validation \gls{NLL} for  classification tasks.
For the \uci classification datasets, \gls{MAP} estimation for the \gpig prior is comparable with early stopping and significantly outperforms the one for the \fg prior (\cref{fig:uci_class_earlystop_map}).
For the \glspl{CNN}, as shown in \cref{fig:cnn_earlystop_map}, we observe that the \gls{MAP} estimations outperform early stopping in most cases.
Besides, it is not clear which prior is better.
We think this can be attributed to the fact that optimization for very deep nets is non-trivial.
As suggested in the literature \citep{Wenzel2020, ashukhapitfalls}, one has to use complicated training strategies such as a learning rate scheduler to obtain good performance for deterministic \glspl{CNN} on high-dimensional data like \cifar.

\setlength\figureheight{.25\textwidth}
\setlength\figurewidth{.16\textwidth}

\setlength\figureheight{.25\textwidth}
\setlength\figurewidth{.16\textwidth}

\section{Conclusions}  \label{sec:conclusions}
In most machine learning tasks, function estimation is a fundamental and ubiquitous problem.
Being able to perform Bayesian inference of neural networks represents a much sought-after objective to equip extremely flexible models with the capability of expressing uncertainty in a sound way \citep{MacKay03,neal1996bayesian}.
Recent advances in \gls{MCMC} sampling enabling for efficient parameter space exploration, combined with mini-batching \citep{Cheni2014}, have turned this long-standing challenge into a concrete possibility.
However despite these advances, there have been only few success stories involving the use of Bayesian inference techniques for neural networks \citep{Osawa2019,zhang2020csgmcmc}.
We attribute this to the difficulties in specifying sensible priors for thousands/millions of parameters, while being able to understand and control the effect of these choices in the behavior of their output functions \citep{Duvenaud14}.

The difficulty in reasoning about functional priors for neural networks, made us consider the possibility to enforce these by minimizing their distance to tractable functional priors, effectively optimizing the priors over model parameters so as to reflect these functional specifications.
We chose to consider Gaussian processes, as they are a natural and popular choice to construct functional priors, whereby the characteristics of prior functions are determined by the form and parameters of Gaussian process kernel/covariance functions.
While previous works attempted this by using the \gls{KL} divergence between the functional priors \citep{Flam2017,Flam2018}, the objective proves difficult to work with due to the need to estimate an entropy term based on samples, which is notoriously difficult.
In this work, we proposed a novel objective based on the Wasserstein distance, and we showed that this objective offers a tractable and stable way to optimize the priors over model parameters.
The attractive property of this objective is that it does not require a closed form for the target functional prior, as long as it is possible to obtain samples from it.
We studied different parameterizations of the priors with increasing flexibility, and we showed that more flexibility makes it indeed possible to improve the match to Gaussian process priors, especially when the activation functions are not suitable to model the target Gaussian processes.
It is worth noting that, as far as we know, normalizing flows have never been proposed to model priors for neural networks, and this represents an interesting line of investigation that deserves some attention for future work.
We are also planning to investigate our proposal on unsupervised/latent variable models, and study ways to reduce the complexity of the optimization of the Wasserstein distance.

After describing our strategy to optimize the Wasserstein distance, we moved on to show the empirical benefits of choosing sensible priors on a large variety of neural network models, including convolutional neural networks, and modeling tasks such as regression and classification under standard conditions, covariate shift, and active learning.
We demonstrated consistent performance improvements over alternatives ways of choosing priors, and we also showed better performance compared to state-of-the-art approximate methods in Bayesian deep learning.
In all, this work confirms the hypothesis that choosing sensible priors for deep models matters, and it offers a practical way to do so.

\acks{MF gratefully acknowledges support from the AXA Research Fund and the Agence Nationale de la Recherche (grant ANR-18-CE46-0002 and ANR-19-P3IA-0002).
	The Authors wish to thank the anonymous reviewers and the action editor for the insightful discussions, comments and questions which helped to improve and clarify this manuscript.
}

\newpage
\appendix

\section{Implementation and experimental details} \label{sec:implementation_details}
In this section, we present details on implementation and hyperparameters used in our experimental campaign.
Our implementation is mainly in PyTorch \citep{paszke2019pytorch}.
We follow the standard
protocol of training, validation and testing.
The hyperparameters are selected according to the \gls{NLL} performance on a validation set, which is created by randomly choosing $20\%$ of the data
points from the training set.
We standardize all the input features and the outputs using the statistics of the training set.
Regarding prior optimization, unless otherwise specified, for the inner loop of \cref{alg:main}, we use the Adagrad optimizer \citep{duchi2011adaptive} with a learning rate of $0.02$, a Lipschitz regularization coefficient $\lambda=10$, and a number of Lipschitz iterations $n_{\text{Lipschitz}} = 200$.
Whereas, for the outer loop of \cref{alg:main} we use the RMSprop optimizer \citep{Tieleman2012} with a learning rate of $0.05$ for the experiments on the \uci and \banana datasets, and a learning rate of $0.01$ for the rest.
See \cref{ssec:wasserstein_convergence} for the progressions of prior optimization.

\subsection{Deep Ensemble}
Deep Ensemble \citep{lakshminarayanan2017simple} averages the predictions across networks trained independently starting from different initializations.
In our experiments, we use an ensemble of $5$ neural networks.
Every member of the ensemble is trained with the $L_2$-regularized objective
\begin{equation}
	\cL(\mbw) := -\frac{1}{N} \sum_{i=1}^{N} \log p(y_i \g \mbx_i, \mbw) + \frac{\lambda}{2} \| \mbw \|_2^2,
\end{equation}
where $N$ is the size of training data, $\lambda$ is the weight decay coefficient, $\log p(y_i \g \mbx_i, \mbw)$ is the log likelihood evaluated at the data point $(\mbx_i, y_i)$.
Following \cite{lakshminarayanan2017simple}, for regression task, in order to capture predictive uncertainty, we use a network that outputs the predicted mean $\mu_{\mbw}(\mbx)$ and variance $\sigma_{\mbw}^2(\mbx)$.
Assume that the observed value follows a heteroscedastic Gaussian distribution, the log likelihood is then
\begin{equation}
	\log p(y_i \g \mbx_i, \mbw) = -\frac{1}{2} \log \sigma_{\mbw}^2(\mbx_i) - \frac{(y_i - \mu_{\mbw}(\mbx_i))^2}{2 \sigma_{\mbw}^2(\mbx_i)} + \mathrm{const}.
\end{equation}
For the classification task, the log likelihood is simply the softmax cross-entropy loss.

We use the Adam optimizer \citep{jlb2015adam} to train all the networks.
For \glspl{MLP}, we use a fixed learning rate $0.01$ and total epochs of 50.
Whereas \glspl{CNN} are trained for $200$ epochs. The learning rate starts from $10^{-2}$ and decays to ($10^{-3}$, $10^{-4}$, $10^{-5}$) at epochs (50, 100, 150).
The $L_2$ regularization strength is tuned over a grid $\lambda \in \left\{ 10^{k} \g \text{$k$ from -8 to -1} \right\}$.

\subsection{Likelihoods for BNNs} \label{ssec:likelihood_for_bnns}
Similarly to the prior, the likelihood for \glspl{BNN} is a modeling choice. It is a function of the model predictions $\hat{\mby}$ and the correct targets $\mby$.
For multi-class $C$-way classification, the \gls{NN} have $C$ output units over which a softmax function is applied, hence the network outputs class probabilities.
The likelihood is commonly chosen as a multinomial distribution, $p(\cD \g \mbw) = \prod_{n=1}^{N} \prod_{c=1}^{C}\hat{y}_{n,c}^{y_{n,c}}$, for $C$ classes, where $\hat{y} \in [0, 1]$ denotes predicted probability, and $y_{n,c}$ is the true targets.

For regression, one usually models output noise as a zero-mean Gaussian: $\epsilon \sim \cN(0, \sigma^2_{\epsilon})$, where $\sigma^2_{\epsilon}$ is the variance of the noise.
The likelihood is then the Gaussian $p(\cD \g \mbw) = \cN(\mby \g \hat{\mby}, \sigma^2_{\epsilon})$.
Notice that the noise variance $\sigma^2_{\epsilon}$ is treated as a hyperparameter.
We do choose this hyperparameter over the grid $\sigma^2_{\epsilon} \in \left\{ 5^{k}, 10^{k} \g \text{$k$ from -3 to -1} \right\} $.
The optimal values are selected according to the \gls{NLL} result of the predictive posterior.
\cref{table:uci_reg_rmse_full} and \cref{table:uci_reg_nll_full} present $\sigma^2_{\epsilon}$ used in the \uci regression experiments.

\subsection{Sampling from the posterior using scale-adapted SGHMC}
As mentioned in \cref{ssec:bayesian_neural_networks}, we make use of the \gls{SGHMC} \citep{Cheni2014} to generate posterior samples for \glspl{BNN}.
One caveat of \gls{SGHMC} and \gls{MCMC} algorithms, in general, is the difficulty of choosing hyperparameters.
To mitigate this problem, in our experiments, we use a scale-apdated version of \gls{SGHMC} \citep{Springenberg2016}, where the hyperparameters are adjusted automatically during a burn-in phase.
After this period, all hyper-parameters stay fixed.

\paragraph{Estimating $\mbM$.} We set the mass matrix $\mbM^{-1} = \mathrm{diag}\left(\hat{V}_{\mbw}^{-1/2} \right)$, where $\hat{V}_{\mbw}$ is an estimate of the uncentered variance of the gradient, $\hat{V}_{\mbw} \approx \mathbb{E}[( \nabla \tilde{U}(\mbw))^2] $, which can be estimated by using exponential moving average as follows
\begin{equation}
	\Delta \hat{V}_{\mbw} = -\tau^{-1} \hat{V}_{\mbw} + \tau^{-1} \nabla (\tilde{U}(\mbw))^2,
\end{equation}
where $\tau$ is a parameter vector that specifies the moving average windows. This parameter can be automatically chosen by using an adaptive estimate \citep{Springenberg2016} as follows
\begin{equation}
	\Delta \tau = -g_{\mbw}^2 \hat{V}^{-1}_{\mbw} \tau + 1, \quad \text{and}, \quad \Delta g_{\mbw} = -\tau^{-1} g_{\mbw} + \tau^{-1} \nabla \tilde{U}(\mbw),
\end{equation}
where $g_{\mbw}$ is a smoothed estimate of the gradient $\nabla U(\mbw)$.

\paragraph{Estimating $\tilde{\mbB}$.} For the estimate for the noise of the gradient evaluation, $\tilde{\mbB}$, it should be ideally the estimate of the empirical Fisher information matrix of $U(\mbw)$, which is prohibitively expensive to compute.
We therefore use a diagonal approximation, $\tilde{\mbB} = \frac{1}{2} \varepsilon \hat{V}_{\mbw}$, which is already available from the step of estimating $\mbM$.

\paragraph{Choosing $\mbC$.} For the friction matrix, in practice, one can simply choose $\mbC = C \mbI$, i.e. the same independent noise for each element of $\mbw$.

\paragraph{The discretized Hamiltonian dynamics.} By substituting $\mbv := \varepsilon \hat{V}_{\mbw}^{-1/2} \mbr$, \cref{eq:discr_hamil_weight} and \cref{eq:discr_hamil_momentum} become
\begin{align}
	\Delta \mbw & = \mbv,                                                                                                                                                                           \\
	\Delta \mbv & = -\varepsilon^{2} \hat{V}_{\mbw}^{-1/2} \nabla \tilde{U}(\mbw) - \varepsilon C \hat{V}_{\mbw}^{-1/2}  \mbv + \cN(0, 2 \varepsilon^3 C \hat{V}_{\mbw}^{-1} - \varepsilon^4 \mbI).
\end{align}
Following \citep{Springenberg2016}, we choose $C$ such that $\varepsilon C \hat{V}_{\mbw}^{-1/2} = \alpha \mbI$. This is equivalent to using a constant momentum coefficient of $\alpha$. The final discretized dynamics are then
\begin{align}
	\Delta \mbw & = \mbv,                                                                                                                                                     \\
	\Delta \mbv & = -\varepsilon^{2} \hat{V}_{\mbw}^{-1/2} \nabla \tilde{U}(\mbw) - \alpha  \mbv + \cN(0, 2 \varepsilon^2 \alpha \hat{V}_{\mbw}^{-1/2} - \varepsilon^4 \mbI).
\end{align}

\paragraph{Experimental configurations.} In all experiments, unless otherwise specified, we use a momentum coefficient $\alpha = 0.01$, and a step size $\varepsilon = 0.01$.
For the \uci regression experiments, we sample four independent chains; for each chain, the number of collected samples after thinning is $30$ except for the large dataset (\protein), where a number of $60$ samples is used.
The thinning intervals are $2000$ and $5000$ iterations for the small and large datasets, respectively.
The burn-in period lasts $2000$ iterations for the \boston, \concrete, \energy, \wine datasets, and $5000$ iterations for the rest.
For the \uci classification experiments, we also use four chains, in which the number of burn-in iterations are $2000$ for small datasets (\eeg, \htru, \magic, and \mocap) and $5000$ for large datasets (\miniboo, \letter, and \drive).
We draw $30$ samples for each chain with a thinning interval of $2000$ iterations for the small datasets and $5000$ iterations for the large datasets.
In the experiments with \glspl{CNN} on \cifar, after a burn-in phase of $10,000$ iterations, we collect $200$ samples with a thinning interval of $10,000$ iterations.

\subsection{Tempered posterior}
We follow the approach of \cite{Wenzel2020} for tempering the posterior as follows
\begin{equation}
	p(\mbw \g \cD) \propto \exp(-U(\mbw) / T),
\end{equation}
where $U(\mbw) = -\log p(\cD \g \mbw) - \log p(\mbw)$ is the potential energy, and $T$ is the temperature value. As suggested by \cite{Wenzel2020}, we only study the ``cold'' posterior, where a temperature $T < 1$ is used.
In this case, we artificially sharpen the posterior by overcounting the training data by a factor of $1 / T$ and rescaling the prior as $p(\mbw)^{\frac{1}{T}}$.
As a result, the posterior distribution is more concentrated around solutions with high likelihood.
In our experiments, we do grid-search over temperature values $T \in \left\{0.5, 0.1, 10^{-2}, 10^{-3}, 10^{-4} \right\}$.

\subsection{Details on the sampling scheme for BNN hierarchical priors}
As mentioned in \cref{ssec:parameterization}, for the \gls{BNN} hierarchical priors, we firstly place a Gaussian prior on the network parameters. For simplicity, let's consider only the weights in $l$-th layer. We have
\begin{align}
	w_l^{(1)}, ..., w_l^{(N_l)} \overset{\text{i.i.d.}}{\sim} \cN(0, \sigma^2_{l_{w}}),
\end{align}
where $w_l^{(i)}$ is the $i$-th weight, $N_l$ is the number of weights in layer $l$. We further place an Inverse-Gamma prior on the variance:
\begin{align}
	\sigma^2_{l_{w}} \sim \Gamma^{-1}(\alpha_{l_{w}}, \beta_{l_{w}}).
\end{align}
We aim to generate samples from the posterior $p\left (\sigma^2_{l_{w}}, \{ w_l^{(i)} \}_{i=1}^{N_l} \g \cD \right)$. As done by \cite{Cheni2014}, the sampling procedure is carried out by alternating the following steps:
\begin{enumerate}[label=(\roman*)]
	\item \label{item:gibb_step_1} Sample weights from $p\left(\{ w_l^{(i)} \}_{i=1}^{N_l} \g \sigma^2_{l_{w}}, \cD \right)$ using the \gls{SGHMC} sampler. We sample the weights for $K$ steps before resampling the variance.
	\item Sample the variance from $p\left(\sigma^2_{l_{w}} \g \{ w_l^{(i)} \}_{i=1}^{N_l} \right)$ using a Gibbs step.
\end{enumerate}
Assume we observed the weights $ \{ w_l^{(i)} \}_{i=1}$ after the step \ref{item:gibb_step_1}, the posterior for the variance can be obtained in a closed form as follows
\begin{align}
	p\left(\sigma^2_{l_{w}} \g \{ w_l^{(i)} \}_{i=1}^{N_l} \right) & \propto \left( \prod_{i=1}^{N_l} p\left(w_l^{(i)} \g \sigma^2_{l_{w}} \right) \right) p\left( \sigma^2_{l_{w}} \g \alpha_{l_w}, \beta_{l_w} \right) \nonumber                                                                                                                            \\
	                                                               & \propto \left( \prod_{i=1}^{N_l} \left(\sigma^2_{l_{w}}\right)^{-1/2} \exp\left\{ -\frac{1}{2\sigma^2_{l_{w}}}\left(w_l^{(i)}\right)^2 \right\} \right) \left( \sigma^2_{l_{w}} \right)^{- \alpha_{l_{w}} - 1} \exp \left\{ -\frac{1}{\sigma^2_{l_{w}}} \beta_{l_{w}} \right\} \nonumber \\
	                                                               & = \left(\sigma^2_{l_{w}} \right)^{-\left(\alpha_{l_w} + N_l /2\right) - 1} \exp\left\{ -\frac{1}{\sigma^2_{l_{w}}}\left( \beta_{l_w} + \frac{1}{2}\sum_{i=1}^{N_l} \left( w_{l}^{(i)} \right)^2 \right) \right\} \nonumber                                                               \\
	                                                               & \propto \Gamma^{-1}\left( \alpha_{l_{w}} + \frac{N_l}{2}, \beta_{l_{w}} + \frac{1}{2}\sum_{i=1}^{N_l} \left( w_{l}^{(i)} \right)^2 \right).
\end{align}
As a default, in our experiments, we set the resampling interval $K = 100$ except for the experiment on the 1D synthetic dataset (\cref{ssec:1d_regression}), in which we use $K = 20$.

\subsection{MAP estimation with Gaussian prior}
For completeness, we describe the \gls{MAP} estimation for the case of Gaussian prior used in \cref{ssec:map}.
This derives interpretation of the regularization effect from the prior for deterministic networks.
We aim at finding a point estimate that maximizes the posterior:
\begin{align}
	\mbw_{\mathrm{MAP}} & = \argmax_{\mbw} p(\mbw \g \cD) \nonumber                \\
	                    & = \argmax_{\mbw} p(\cD \g \mbw) p(\mbw) \nonumber        \\
	                    & = \argmax_{\mbw} \{\log p(\cD \g \mbw) + \log p(\mbw)\}.
\end{align}
If the prior is a Gaussian distribution, $p(\mbw) = \cN(\mbmu, \mbSigma)$, we have
\begin{align}
	\mbw_{\mathrm{MAP}} & = \argmax_{\mbw} \left\{
	\log p(\cD \g \mbw) - \frac{1}{2} (\mbw - \mbmu)^{\top} \mbSigma^{-1} (\mbw - \mbmu) \right\}.
\end{align}
In our experiments, the prior covariance is set as isotropic, $\mbSigma = \sigma^2_{\mathrm{prior}} \mbI$, and prior mean is zero, $\mbmu = \mathbf{0}$. Thus, we have
\begin{align}
	\mbw_{\mathrm{MAP}} & = \argmax_{\mbw} \left\{ \log p(\cD \g \mbw) - \frac{1}{2 \sigma^2_{\mathrm{prior}}} \| \mbw \|_{2}^{2} \right\}
\end{align}
Here, we use the same likelihoods $p(\cD | \mbw)$ as in \cref{ssec:likelihood_for_bnns}.

\paragraph{Regression task.} For the Gaussian likelihood $p(\cD \g \mbw) = \cN(\mby \g \hat{\mby}, \sigma^2_{\epsilon})$, the \gls{MAP} estimation is then
\begin{align}
	\mbw_{\mathrm{MAP}} & = \argmax_{\mbw} \left\{ -\frac{1}{2 \sigma^2_{\epsilon}} \| \hat{\mby} - \mby \|_{2}^{2} - \frac{1}{2 \sigma^2_{\mathrm{prior}}} \| \mbw \|_{2}^{2} \right\}.
\end{align}
This is equivalent to minimizing the $L_2$-regularized squared-error objective:
\begin{align}
	\mbw_{\mathrm{MAP}} & = \argmin_{\mbw} \left\{ \sum_{n=1}^{N} (\hat{y}_{n} - y_{n})^2 + \frac{\sigma^2_{\epsilon}}{\sigma^2_{\mathrm{prior}}} \| \mbw \|_{2}^{2} \right\}.
\end{align}
Here, we can interpret that the term $\frac{\sigma^2_{\epsilon}}{\sigma^2_{\mathrm{prior}}}$ controls the regularization strength.

\paragraph{Classification task.} For the multinomial likelihood $p(\cD \g \mbw) = \prod_{n=1}^{N} \prod_{c=1}^{C}\hat{y}_{n,c}^{y_{n,c}}$, estimating \gls{MAP} is equivalent to minimizing the $L_2$-regularized cross-entropy objective:
\begin{align}
	\mbw_{\mathrm{MAP}} & = \argmin_{\mbw}\left\{ - \sum_{n=1}^{N} \sum_{c=1}^{C} y_{n,c} \log(\hat{y}_{n, c}) + \frac{1}{2 \sigma^2_{\mathrm{prior}}} \| \mbw \|_{2}^{2} \right\},
\end{align}
where $\frac{1}{\sigma^2_{\mathrm{prior}}}$ is the regularization coefficient.

\begin{table}[H]
	\centering
	\footnotesize
	\begin{minipage}[b]{0.45\textwidth}
		\centering
		\begin{tabular}{cc}
			\toprule
			\textbf{Layer} & \textbf{Dimensions}              \\ \hline \hline
			Conv2D         & $3 \times 6  \times 5 \times 5 $ \\
			Conv2D         & $6 \times 16  \times 5 \times 5$ \\ \hline
			Linear-ReLU    & $400 \times 120$                 \\
			Linear-ReLU    & $120 \times 84$                  \\
			Linear-Softmax & $84 \times 10$                   \\ \bottomrule
		\end{tabular}
		\vskip 0.05in
		\captionof{table}{\lenet \label{tab:architecture_lenet}}

		\vspace{0.5cm}

		\begin{tabular}{cc}
			\toprule
			\textbf{Layer} & \textbf{Dimensions}                                  \\ \hline \hline
			Conv2D         & $3 \times 16 \times 3 \times 3$                      \\ \hline
			Residual Block & $\left[ \begin{matrix} 3 \times 3, 16 \\ 3 \times 3, 16 \end{matrix} \right] \times 3$ \\ \hline
			Residual Block & $\left[ \begin{matrix} 3 \times 3, 32 \\ 3 \times 3, 32 \end{matrix} \right] \times 3$ \\ \hline
			Residual Block & $\left[ \begin{matrix} 3 \times 3, 64 \\ 3 \times 3, 64 \end{matrix} \right] \times 3$ \\ \hline
			AvgPool        & $8 \times 8$                                         \\
			Linear-Softmax & $64 \times 10$                                       \\ \bottomrule
		\end{tabular}
		\vskip 0.05in
		\captionof{table}{\preresnet \label{tab:architecture_preresnet}}

	\end{minipage}
	\begin{minipage}[b]{0.45\textwidth}
		\centering
		\begin{tabular}[t]{cc}
			\toprule
			\textbf{Layer} & \textbf{Dimensions}                  \\ \hline \hline
			Conv2D         & $3 \times 32 \times 3 \times 3 $     \\
			Conv2D         & $32 \times 32  \times 3 \times 3 $   \\
			MaxPool        & $2 \times 2 $                        \\ \hline
			Conv2D         & $32 \times 64 \times 3 \times 3 $    \\
			Conv2D         & $64 \times 64  \times 3 \times 3 $   \\
			MaxPool        & $2 \times 2 $                        \\ \hline
			Conv2D         & $64 \times 128 \times 3 \times 3 $   \\
			Conv2D         & $128 \times 128  \times 3 \times 3 $ \\
			Conv2D         & $128 \times 128  \times 3 \times 3 $ \\
			MaxPool        & $2 \times 2 $                        \\ \hline
			Conv2D         & $128 \times 256 \times 3 \times 3 $  \\
			Conv2D         & $256 \times 256  \times 3 \times 3 $ \\
			Conv2D         & $256 \times 256  \times 3 \times 3 $ \\
			MaxPool        & $2 \times 2 $                        \\ \hline
			Conv2D         & $256 \times 256 \times 3 \times 3 $  \\
			Conv2D         & $256 \times 256  \times 3 \times 3 $ \\
			Conv2D         & $256 \times 256  \times 3 \times 3 $ \\
			MaxPool        & $2 \times 2 $                        \\ \hline
			Linear-ReLU    & $256 \times 256$                     \\
			Linear-ReLU    & $256 \times 256$                     \\
			Linear-Softmax & $256 \times 10$                      \\\bottomrule
		\end{tabular}
		\vskip 0.05in
		\captionof{table}{\vgg \label{tab:architecture_vgg}}

	\end{minipage}
\end{table}

\subsection{Network architectures}
As previously mentioned in \cref{ssec:bayesian_neural_networks}, we employ the NTK parameterization \citep{jacot2018neural, lee2020finite} for \glspl{MLP} and \glspl{CNN}.
We initialize the weights $w_{l} \sim \cN(0, 1)$ and $b_{l} = 0$ for both fully-connected and convolutional layers.
\cref{tab:architecture_lenet,tab:architecture_preresnet,tab:architecture_vgg} show details on the \glspl{CNN} architectures used in our experimental campaign.
These networks are adapted to the \cifar dataset.
The parameters of batch normalization layers of \preresnet are treated as constants.
In particular, we set the scale and shift parameters to $1$ and $0$, respectively.

\subsection{Measuring similarity between GPs and BNNs using maximum mean discrepancy} \label{ssec:mmd_appendix}
{
	In \cref{ssec:wasserstein_vs_kl}, we adopted the approach of \cite{Matthews2018} to measure the similarity between \glspl{GP} and \glspl{BNN} using a kernel two-sample test based on \gls{MMD} \cite{GrettonBRSS12}.
	The \gls{MMD} between two distributions $p_\sub{gp}$ and $p_\sub{nn}$ is defined as follows
	\begin{equation}
		\text{MMD}(p_\sub{gp}, p_\sub{nn}) = \sup_{\|h\|_{\mathcal{H}} \leq 1} \Big[ \mathbb{E}_{p_\sub{gp}}[h] - \mathbb{E}_{p_\sub{nn}}[h] \Big],
	\end{equation}
	where $\mathcal{H}$ denotes a \gls{RKHS} induced by a characteristic kernel $K$.
	Similarly to the Wasserstein distance, \gls{MMD} is an integral probability metric \citep{muller1997integral}.
	The main difference is the choice of class functions $\mathcal{H}$ as we consider the class of 1-Lipschitz functions for the Wasserstein distance.
	In fact, under some mild conditions, these metrics are equivalent.

	By considering two stochastic processes $p_\sub{gp}$ and $p_\sub{nn}$ at a finite number of measurement points $\mbX_{\mathcal{M}}$, we can obtain the closed form of \gls{MMD} as follows
	\begin{align}
		\text{MMD}^2 (p_\sub{gp}, p_\sub{nn}) & = \mathbb{E}_{\mbf_{\mathcal{M}}, {\mbf'}_{\mathcal{M}} \sim p_\sub{gp}}[K(\mbf_{\mathcal{M}}, {\mbf'}_{\mathcal{M}})] + \mathbb{E}_{\mbf_{\mathcal{M}}, {\mbf'}_{\mathcal{M}} \sim p_\sub{nn}}[K(\mbf_{\mathcal{M}}, {\mbf'}_{\mathcal{M}})] \\
		                                      & - 2 \mathbb{E}_{\mbf_{\mathcal{M}} \sim p_\sub{gp}, {\mbf'}_{\mathcal{M}} \sim p_\sub{nn} }[K(\mbf_{\mathcal{M}}, {\mbf'}_{\mathcal{M}})] \nonumber,
	\end{align}
	which can be estimated by using samples from $p_\sub{nn}$ and $p_\sub{gp}$ evaluated at $\mbX_{\mathcal{M}}$ \citep{GrettonBRSS12}.
	For the \gls{MMD} estimate, we use  an \gls{RBF} kernel with a characteristic lengthscale of $l = \sqrt{2D}$, where $D$ is the number of dimensions of the input features, and $5000$ samples from $p_\sub{nn}$ and $p_\sub{gp}$.
	The measurement set is comprised of $500$ test points.
}

{
	\subsection{Details on the experiments with functional BNNs and empirical Bayes} \label{ssec:empirical_bayes_setup}

	In the experiments with \fbnn, we keep the same settings as used in \cite{Sun2019}\footnote{\url{https://github.com/ssydasheng/FBNN}}.
	In particular, we use a \gls{GP} with \gls{RBF} kernels for small  \uci datasets with less than $2000$ data points, while a \gls{GP} with \gls{NKN} kernels is employed for large \uci datasets.

	In the experiments with the empirical Bayes approach \citep{ImmerBFRK21}, following the Authors' repository\footnote{\url{https://github.com/AlexImmer/marglik}}, we use the \textit{Laplace} library \citep{Daxberger2021LaplaceR} for the implementation.
	We use the Kronecker-factored Laplace for Hessian approximation.
	We follow the same experimental protocol of \citep{ImmerBFRK21} including the optimizer, the early stopping scheme and the frequency of updating the prior.

}

\section{Additional results} \label{sec:additional_results}

\subsection{Additional results on MAP estimation with GP-induced priors}
\begin{figure}[t]
	\centering
	\scriptsize
	\includegraphics{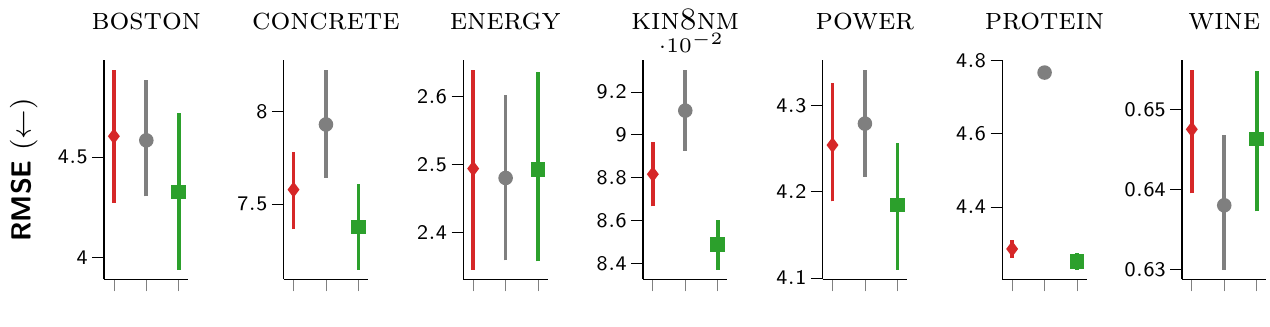}
	\definecolor{early_stopping}{rgb}{0.8359375,0.15234375,0.15625}
\definecolor{map_std}{rgb}{0.49609375,0.49609375,0.49609375}
\definecolor{map_optim}{rgb}{0.171875,0.625,0.171875} \begin{tabular}{ccc}  {\protect\tikz[baseline=-1ex]\protect\draw[thick, color=early_stopping, fill=early_stopping, mark=diamond, mark size=1.2pt, line width=0.9pt] plot[] (-.0, 0)--(.25,0)--(-.25,0);}  Early Stopping &  {\protect\tikz[baseline=-1ex]\protect\draw[thick, color=map_std, fill=map_std, mark=*, mark size=1.2pt, line width=0.9pt] plot[] (-.0, 0)--(.25,0)--(-.25,0);}  \acrshort{MAP} with \fg prior &  {\protect\tikz[baseline=-1ex]\protect\draw[thick, color=map_optim, fill=map_optim, mark=square, mark size=1.2pt, line width=0.9pt] plot[] (-.0, 0)--(.25,0)--(-.25,0);}  \acrshort{MAP} with \gpig prior (\textbf{ours})  \end{tabular}
	\caption{Comparison between early stopping and \gls{MAP} estimations with respect to the \fg and \gpig priors on the \uci regression datasets.  \label{fig:uci_reg_earlystop_map}}
\end{figure}

\cref{fig:uci_reg_earlystop_map} illustrates the comparison between early stopping, and \gls{MAP} estimation with the \fg and \gpig priors on the \uci regression datasets.
We use the same setup as in \cref{ssec:map}.
We observe that the predictive performance obtained by \gls{MAP} with the \gpig prior outperforms those of early stopping and  \gls{MAP} with the \fg prior in most cases.

\subsection{Tabular results on the UCI benchmarks}
Detailed results on the \uci regression and classification datasets are reported in \cref{table:uci_reg_rmse_full,table:uci_reg_nll_full,table:uci_class_accuracy_full,table:uci_class_nll_full}.

\vspace{-2ex}
\subsection{Convergence of Wasserstein optimization} \label{ssec:wasserstein_convergence}
\begin{figure}[th!]
	\centering
	\scriptsize
	\begin{subfigure}[t]{1.0\textwidth}
		\includegraphics{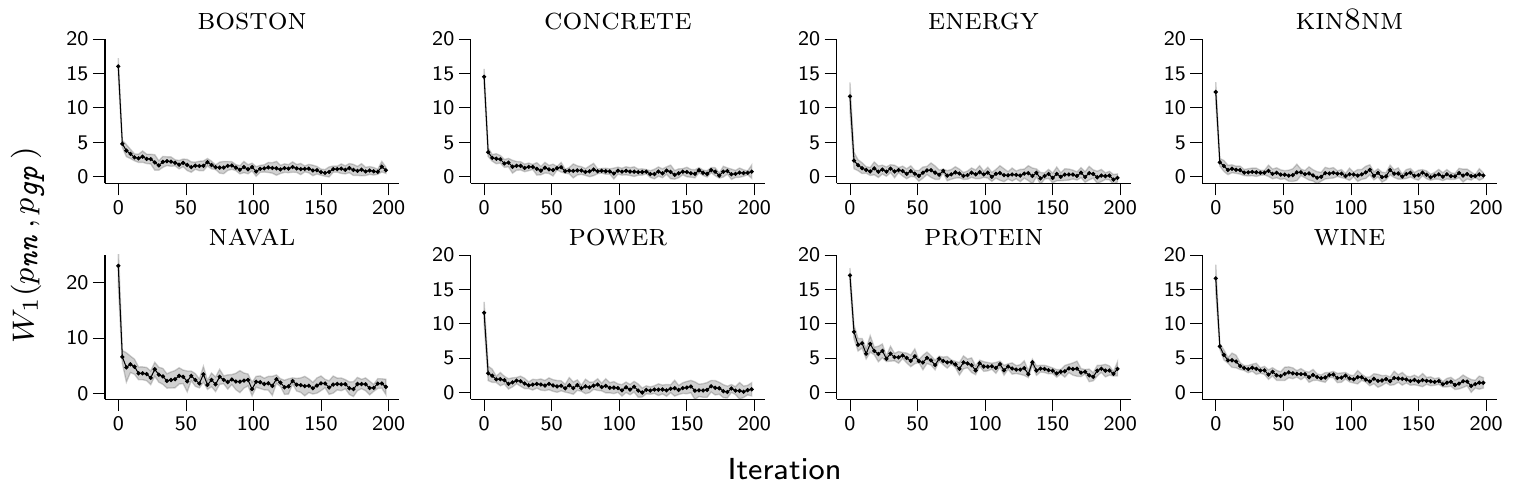}
		\vskip -0.1in
		\caption{\gpig prior}
	\end{subfigure}
	\vskip 0.1in

	\begin{subfigure}[t]{1.0\textwidth}
		\includegraphics{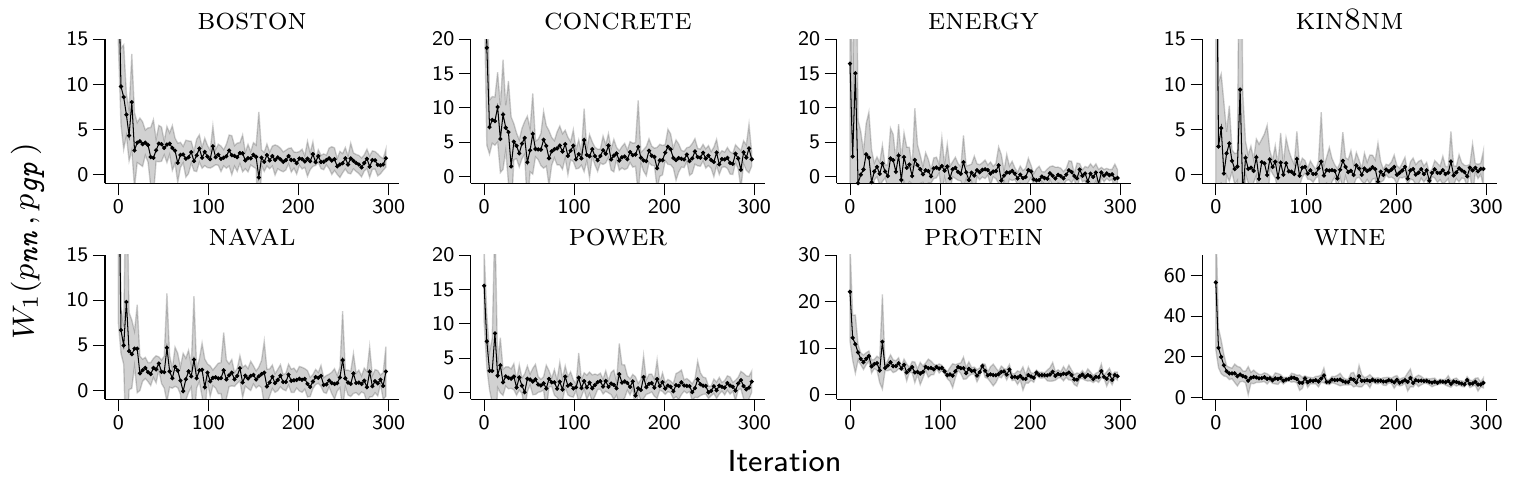}
		\vskip -0.1in
		\caption{\gpih prior}
	\end{subfigure}
	\vskip 0.1in

	\begin{subfigure}[t]{1.0\textwidth}
		\includegraphics{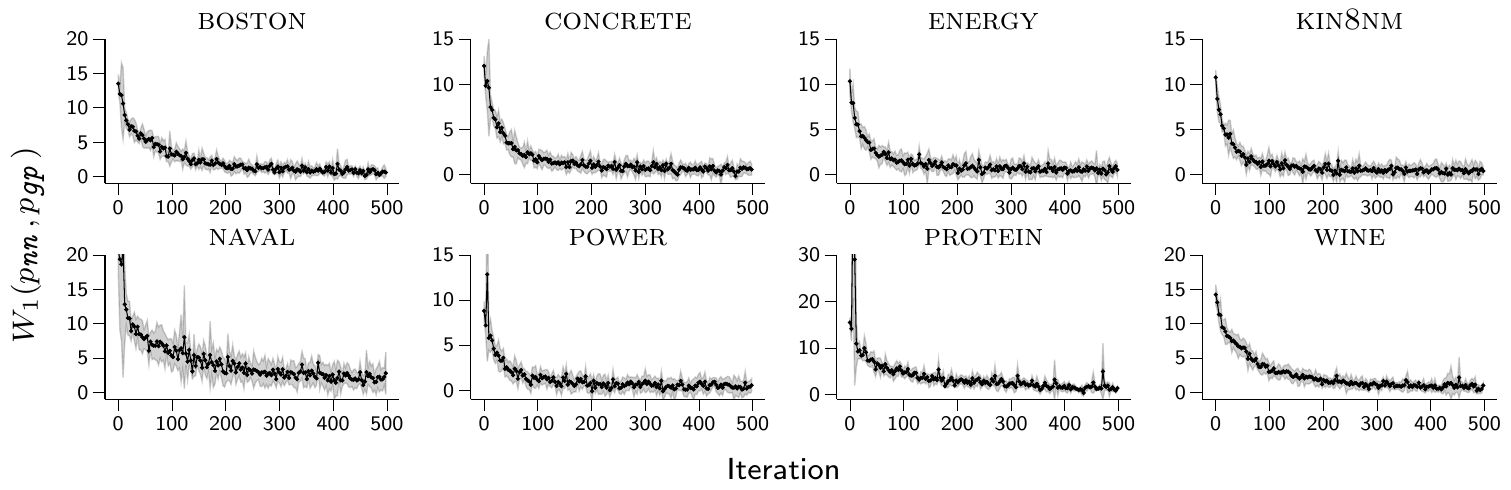}
		\vskip -0.1in
		\caption{\gpinf prior}
	\end{subfigure}

	\caption{Convergence of Wasserstein optimization for two-layer \glspl{MLP} on the \uci regression datasets.  \label{fig:convergence_uci_reg}}
\end{figure}

\begin{figure}[th!]
	\centering
	\scriptsize
	\begin{subfigure}[t]{1.0\textwidth}
		\includegraphics{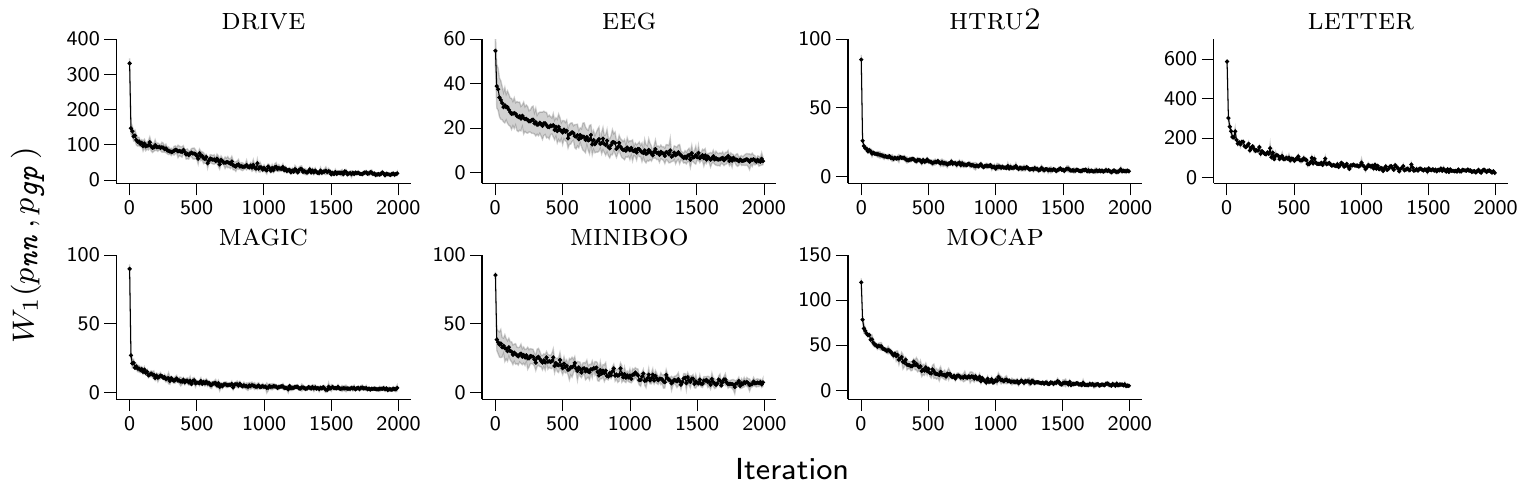}
		\vskip -0.1in
		\caption{\gpig prior}
	\end{subfigure}
	\vskip 0.1in

	\begin{subfigure}[t]{1.0\textwidth}
		\includegraphics{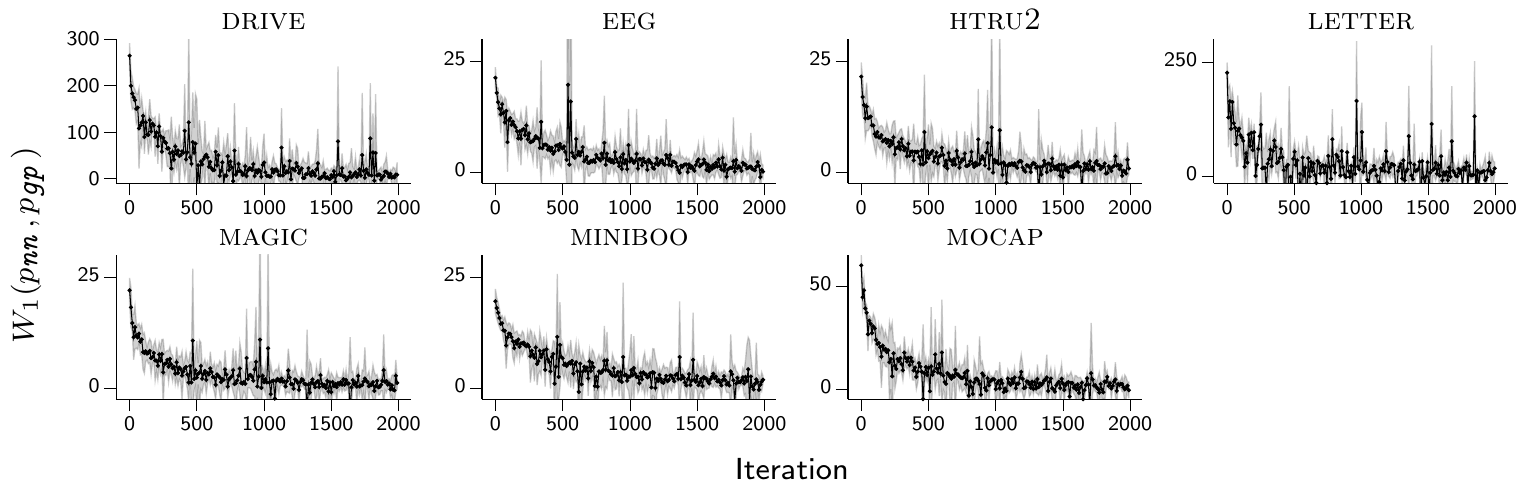}
		\vskip -0.1in
		\caption{\gpih prior}
	\end{subfigure}

	\caption{Convergence of Wasserstein optimization for two-layer \glspl{MLP} on the \uci classification datasets.  \label{fig:convergence_uci_class}}
\end{figure}

\begin{figure}[H]
	\scriptsize
	\begin{subfigure}[t]{1.0\textwidth}
		\centering
		\includegraphics{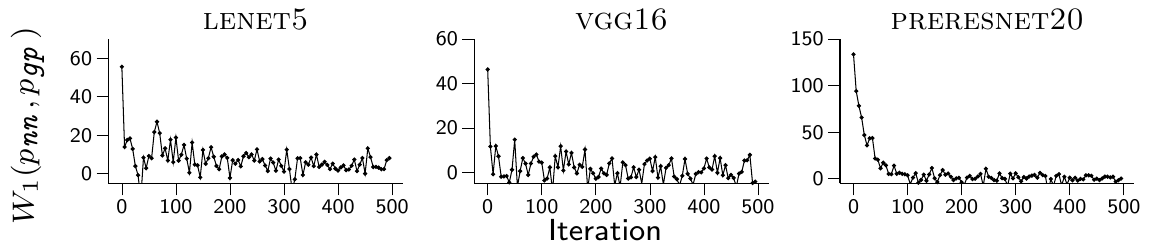}
		\caption{\gpig prior}
	\end{subfigure}
	\vskip 0.1in

	\begin{subfigure}[t]{1.0\textwidth}
		\centering
		\includegraphics{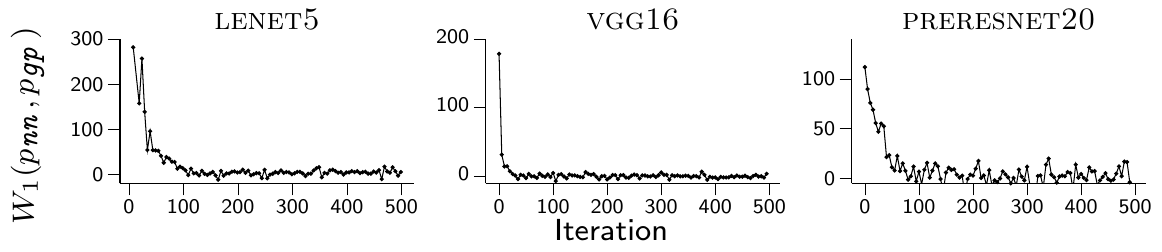}
		\caption{\gpih prior}
	\end{subfigure}

	\caption{Convergence of Wasserstein optimization for \glspl{CNN} on the \cifar dataset.  \label{fig:convergence_cnn}}
\end{figure}

\cref{fig:convergence_uci_reg,fig:convergence_uci_class,fig:convergence_cnn} depict the progressions of Wasserstein optimization in the \uci regression, \uci classification, and \cifar experiments, respectively.

\begin{figure}[t]
	\centering
	\scriptsize
	\includegraphics{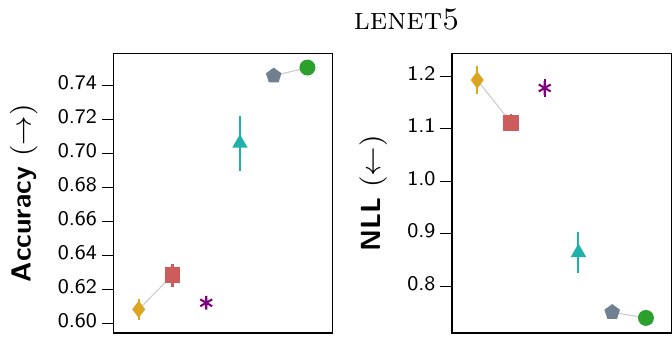}
	\hspace{3ex}
	\includegraphics{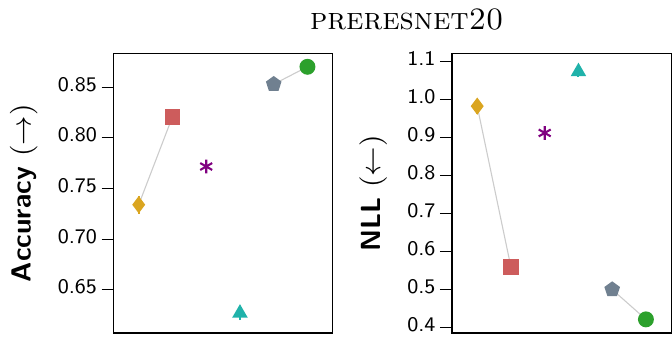}
	\vskip 0.07in
	\definecolor{std_laplace}{rgb}{0.85,0.65,0.13}
\definecolor{optim_gaussian}{rgb}{0.0703125,0.6328125,0.2265625}
\definecolor{wasserstein_laplace}{rgb}{0.51171875,0.0,0.48828125}
\definecolor{optim_fbnn}{rgb}{0.421875,0.31640625,0.78125}
\definecolor{neal}{rgb}{0.49609375,0.49609375,0.49609375}
\definecolor{optim_laplace}{rgb}{0.81640625,0.36328125,0.3671875}
\definecolor{sghmc_laplace}{rgb}{0.0,0.6953125,0.6640625} \begin{tabular}{lll} 
    {\protect\tikz[baseline=-1ex]\protect\draw[thick, color=std_laplace, fill=std_laplace, mark=diamond*, mark size=1.5pt, line width=0.9pt] plot[] (-.0, 0)--(.25,0)--(-.25,0);}  {\fg prior + \laplace-\ggn} &
    {\protect\tikz[baseline=-1ex]\protect\draw[thick, color=optim_laplace, fill=optim_laplace, mark=square*, mark size=1.5pt, line width=0.9pt] plot[] (-.0, 0)--(.25,0)--(-.25,0);}  {\lagnn} &
    {\protect\tikz[baseline=-1ex]\protect\draw[thick, color=wasserstein_laplace, fill=wasserstein_laplace, mark=asterisk, mark size=1.5pt, line width=0.9pt] plot[] (-.0, 0)--(.25,0)--(-.25,0);}  {\gpig prior + \lagnn} 
    \\
    {\protect\tikz[baseline=-1ex]\protect\draw[thick, color=sghmc_laplace, fill=sghmc_laplace, mark=triangle*, mark size=1.5pt, line width=0.9pt] plot[] (-.0, 0)--(.25,0)--(-.25,0);}  {\lamarglik + \sghmc} &
    {\protect\tikz[baseline=-1ex]\protect\draw[thick, color=neal, fill=neal, mark=pentagon*, mark size=1.5pt, line width=0.9pt] plot[] (-.0, 0)--(.25,0)--(-.25,0);}  {\fg prior + \sghmc} &
    {\protect\tikz[baseline=-1ex]\protect\draw[thick, color=optim_gaussian, fill=optim_gaussian, mark=*, mark size=1.5pt, line width=0.9pt] plot[] (-.0, 0)--(.25,0)--(-.25,0);}  {\gpig prior + \sghmc (\textbf{ours})} 
\end{tabular}
	\vspace{-1ex}
	\caption{Comparison with empirical Bayes and functional inference methods on \cifar dataset.
		\label{fig:cifar_empirical_bayes_appendix}}
\end{figure}
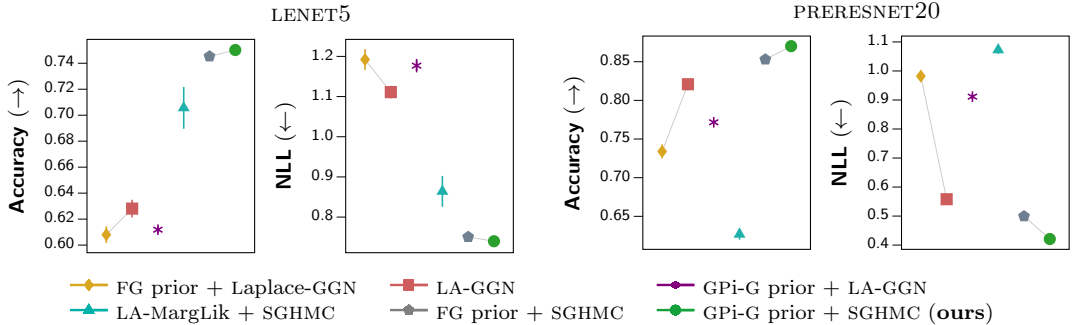

	{
		\vspace{-2ex}
		\subsection{Additional comparisons with the empirical Bayes approach} \label{ssec:empirical_bayes_appendix}
		We complement the results presented in \cref{ssec:empirical_bayes} with different scenarios of optimizing the prior and carrying out the inference.
		In particular, we evaluate our \gpig prior when employed with the scalable \gls{LA} approach \citep{ImmerKB21} for inference, refered as ``\gpig prior + \lagnn''.
		As shown in \cref{fig:cifar_empirical_bayes_appendix}, the \gpig prior still outperforms the fixed prior (\fg prior).
		In addition, we consider the case where the prior optimized on the approximated marginal likelihood \citep{ImmerBFRK21} is used together with \gls{SGHMC}.
		We denote this approach ``\lamarglik + \sghmc''.
		As it can be seen from the results, this prior is not helpful and even worse than the fixed prior when employed with the \gls{SGHMC}.
		This is reasonable because the \lamarglik prior is closely tied with the \lagnn inference method; the marginal likelihood is optimized jointly with the approximate posterior, and the same optimized hyper-parameters might not work just as well for a different posterior approximation.
	}

\setlength\figureheight{.24\textwidth}
\setlength\figurewidth{.16\textwidth}

\setlength\figureheight{.29\textwidth}
\setlength\figurewidth{.25\textwidth}

\clearpage

\begin{table}[p]
	\centering
	\footnotesize
	\resizebox{1.0\columnwidth}{!}{
		\begin{tabular}{l||rr|r|c|cccccc}
			\toprule
			Data set                   & $N$                    & $D$                 & $\sigma^2_{\epsilon}$  & \textit{Depth} & \fg prior                  & \fgts                      & \gpig prior                & \fh prior                  & \gpih prior                & Deep Ensemble              \\
			\midrule
			\midrule
			\multirow{4}{*}{\boston}   & \multirow{4}{*}{506}   & \multirow{4}{*}{13} & \multirow{4}{*}{0.1}   & 1              & 3.124 $\pm$ 1.065          & 3.065 $\pm$ 0.964          & \textbf{2.823} $\pm$ 0.960 & 2.949 $\pm$ 1.041          & 2.850 $\pm$ 1.007          & 3.764 $\pm$ 1.122          \\
			                           &                        &                     &                        & 2              & 3.093 $\pm$ 1.001          & 3.020 $\pm$ 0.938          & 2.835 $\pm$ 0.922          & 2.945 $\pm$ 0.996          & \textbf{2.826} $\pm$ 0.909 & 3.688 $\pm$ 1.147          \\
			                           &                        &                     &                        & 4              & 3.120 $\pm$ 0.961          & 2.975 $\pm$ 0.906          & \textbf{2.869} $\pm$ 0.881 & 2.941 $\pm$ 0.944          & 2.931 $\pm$ 0.875          & 3.540 $\pm$ 1.166          \\
			                           &                        &                     &                        & 8              & 3.228 $\pm$ 0.924          & \textbf{2.973} $\pm$ 0.849 & 2.976 $\pm$ 0.957          & 3.078 $\pm$ 1.004          & 3.110 $\pm$ 0.950          & 3.542 $\pm$ 1.068          \\
			\midrule

			\multirow{4}{*}{\concrete} & \multirow{4}{*}{1030}  & \multirow{4}{*}{8}  & \multirow{4}{*}{0.1}   & 1              & 5.442 $\pm$ 0.263          & 5.419 $\pm$ 0.250          & \textbf{4.765} $\pm$ 0.386 & 4.930 $\pm$ 0.390          & 4.781 $\pm$ 0.443          & 5.632 $\pm$ 0.563          \\
			                           &                        &                     &                        & 2              & 5.488 $\pm$ 0.253          & 5.388 $\pm$ 0.296          & \textbf{4.801} $\pm$ 0.416 & 5.179 $\pm$ 0.280          & 4.822 $\pm$ 0.396          & 5.226 $\pm$ 0.631          \\
			                           &                        &                     &                        & 4              & 5.651 $\pm$ 0.262          & 5.326 $\pm$ 0.337          & 5.024 $\pm$ 0.321          & 5.557 $\pm$ 0.245          & \textbf{4.946} $\pm$ 0.384 & 5.011 $\pm$ 0.560          \\
			                           &                        &                     &                        & 8              & 5.839 $\pm$ 0.311          & 5.289 $\pm$ 0.365          & 5.515 $\pm$ 0.339          & 5.757 $\pm$ 0.274          & 5.184 $\pm$ 0.315          & \textbf{5.124} $\pm$ 0.517 \\
			\midrule

			\multirow{4}{*}{\energy}   & \multirow{4}{*}{768}   & \multirow{4}{*}{8}  & \multirow{4}{*}{0.001} & 1              & 0.395 $\pm$ 0.071          & 0.392 $\pm$ 0.071          & \textbf{0.366} $\pm$ 0.080 & 0.393 $\pm$ 0.074          & 0.370 $\pm$ 0.076          & 2.252 $\pm$ 0.241          \\
			                           &                        &                     &                        & 2              & 0.389 $\pm$ 0.062          & 0.381 $\pm$ 0.068          & \textbf{0.343} $\pm$ 0.071 & 0.439 $\pm$ 0.063          & 0.358 $\pm$ 0.071          & 1.382 $\pm$ 0.348          \\
			                           &                        &                     &                        & 4              & 0.422 $\pm$ 0.051          & 0.402 $\pm$ 0.061          & 0.396 $\pm$ 0.063          & 0.428 $\pm$ 0.061          & \textbf{0.394} $\pm$ 0.063 & 1.049 $\pm$ 0.340          \\
			                           &                        &                     &                        & 8              & 0.457 $\pm$ 0.052          & \textbf{0.418} $\pm$ 0.063 & 0.475 $\pm$ 0.056          & 0.467 $\pm$ 0.055          & 0.437 $\pm$ 0.058          & 1.041 $\pm$ 0.323          \\
			\midrule

			\multirow{4}{*}{\kinm}     & \multirow{4}{*}{8192}  & \multirow{4}{*}{8}  & \multirow{4}{*}{0.1}   & 1              & 0.066 $\pm$ 0.002          & 0.066 $\pm$ 0.002          & \textbf{0.065} $\pm$ 0.002 & \textbf{0.065} $\pm$ 0.002 & \textbf{0.065} $\pm$ 0.002 & 0.071 $\pm$ 0.004          \\
			                           &                        &                     &                        & 2              & 0.066 $\pm$ 0.002          & 0.065 $\pm$ 0.002          & \textbf{0.064} $\pm$ 0.002 & 0.065 $\pm$ 0.002          & \textbf{0.064} $\pm$ 0.002 & 0.068 $\pm$ 0.004          \\
			                           &                        &                     &                        & 4              & 0.067 $\pm$ 0.002          & 0.065 $\pm$ 0.002          & 0.065 $\pm$ 0.002          & 0.069 $\pm$ 0.002          & \textbf{0.064 }$\pm$ 0.002 & 0.070 $\pm$ 0.003          \\
			                           &                        &                     &                        & 8              & 0.069 $\pm$ 0.002          & \textbf{0.065} $\pm$ 0.002 & 0.070 $\pm$ 0.002          & 0.072 $\pm$ 0.002          & \textbf{0.065} $\pm$ 0.002 & 0.071 $\pm$ 0.003          \\
			\midrule

			\multirow{4}{*}{\naval}    & \multirow{4}{*}{11934} & \multirow{4}{*}{16} & \multirow{4}{*}{0.001} & 1              & \textbf{0.000} $\pm$ 0.000 & \textbf{0.000} $\pm$ 0.000 & \textbf{0.000} $\pm$ 0.000 & \textbf{0.000} $\pm$ 0.000 & \textbf{0.000} $\pm$ 0.000 & 0.004 $\pm$ 0.000          \\
			                           &                        &                     &                        & 2              & \textbf{0.000} $\pm$ 0.000 & \textbf{0.000} $\pm$ 0.000 & \textbf{0.000} $\pm$ 0.000 & \textbf{0.000} $\pm$ 0.000 & \textbf{0.000} $\pm$ 0.000 & 0.003 $\pm$ 0.000          \\
			                           &                        &                     &                        & 4              & \textbf{0.000} $\pm$ 0.000 & \textbf{0.000} $\pm$ 0.000 & \textbf{0.000} $\pm$ 0.000 & \textbf{0.000} $\pm$ 0.000 & \textbf{0.000} $\pm$ 0.000 & 0.003 $\pm$ 0.000          \\
			                           &                        &                     &                        & 8              & \textbf{0.001} $\pm$ 0.000 & \textbf{0.001} $\pm$ 0.000 & \textbf{0.001} $\pm$ 0.000 & \textbf{0.001} $\pm$ 0.000 & \textbf{0.001} $\pm$ 0.000 & 0.004 $\pm$ 0.000          \\
			\midrule

			\multirow{4}{*}{\power}    & \multirow{4}{*}{9568}  & \multirow{4}{*}{4}  & \multirow{4}{*}{0.05}  & 1              & 4.003 $\pm$ 0.162          & 4.000 $\pm$ 0.164          & \textbf{3.897} $\pm$ 0.177 & 4.022 $\pm$ 0.159          & 3.936 $\pm$ 0.170          & 4.008 $\pm$ 0.182          \\
			                           &                        &                     &                        & 2              & 4.008 $\pm$ 0.168          & 3.999 $\pm$ 0.170          & \textbf{3.723} $\pm$ 0.183 & 4.054 $\pm$ 0.155          & 3.823 $\pm$ 0.179          & 3.857 $\pm$ 0.191          \\
			                           &                        &                     &                        & 4              & 4.064 $\pm$ 0.163          & 4.014 $\pm$ 0.165          & 3.835 $\pm$ 0.173          & 4.163 $\pm$ 0.147          & \textbf{3.814} $\pm$ 0.177 & 3.826 $\pm$ 0.186          \\
			                           &                        &                     &                        & 8              & 4.105 $\pm$ 0.160          & 4.042 $\pm$ 0.165          & 4.062 $\pm$ 0.188          & 4.205 $\pm$ 0.149          & 3.895 $\pm$ 0.167          & \textbf{3.854} $\pm$ 0.179 \\
			\midrule

			\multirow{4}{*}{\protein}  & \multirow{4}{*}{45730} & \multirow{4}{*}{9}  & \multirow{4}{*}{0.5}   & 1              & 4.374 $\pm$ 0.019          & 4.376 $\pm$ 0.015          & \textbf{3.922} $\pm$ 0.011 & 3.973 $\pm$ 0.019          & 3.926 $\pm$ 0.019          & 4.376 $\pm$ 0.019          \\
			                           &                        &                     &                        & 2              & 4.379 $\pm$ 0.019          & 4.330 $\pm$ 0.024          & 3.658 $\pm$ 0.021          & 3.713 $\pm$ 0.021          & \textbf{3.644} $\pm$ 0.025 & 4.443 $\pm$ 0.020          \\
			                           &                        &                     &                        & 4              & 4.509 $\pm$ 0.015          & 4.321 $\pm$ 0.019          & 4.082 $\pm$ 0.055          & 3.976 $\pm$ 0.035          & \textbf{3.774} $\pm$ 0.021 & 3.854 $\pm$ 0.038          \\
			                           &                        &                     &                        & 8              & 4.530 $\pm$ 0.020          & 4.362 $\pm$ 0.014          & 4.593 $\pm$ 0.108          & 4.148 $\pm$ 0.031          & \textbf{3.980} $\pm$ 0.022 & 3.997 $\pm$ 0.027          \\
			\midrule

			\multirow{4}{*}{\wine}     & \multirow{4}{*}{1599}  & \multirow{4}{*}{11} & \multirow{4}{*}{0.5}   & 1              & 0.637 $\pm$ 0.042          & 0.636 $\pm$ 0.044          & 0.618 $\pm$ 0.045          & 0.633 $\pm$ 0.044          & 0.622 $\pm$ 0.045          & \textbf{0.612} $\pm$ 0.020 \\
			                           &                        &                     &                        & 2              & 0.641 $\pm$ 0.044          & 0.641 $\pm$ 0.044          & \textbf{0.609} $\pm$ 0.046 & 0.637 $\pm$ 0.044          & 0.613 $\pm$ 0.046          & 0.615 $\pm$ 0.025          \\
			                           &                        &                     &                        & 4              & 0.650 $\pm$ 0.045          & 0.649 $\pm$ 0.046          & 0.608 $\pm$ 0.046          & 0.637 $\pm$ 0.044          & \textbf{0.602} $\pm$ 0.048 & \textbf{0.602} $\pm$ 0.031 \\
			                           &                        &                     &                        & 8              & 0.662 $\pm$ 0.049          & 0.660 $\pm$ 0.049          & 0.632 $\pm$ 0.046          & 0.646 $\pm$ 0.046          & 0.621 $\pm$ 0.048          & \textbf{0.609} $\pm$ 0.026 \\

			\bottomrule
		\end{tabular}
	}
	\vskip 0.15in
	\caption{Average test \gls{RMSE} on \uci regression datasets (errors are $\pm 1$ standard error). Bold results indicate the best performance. Here, $N$ is the size of dataset, $D$ is the number of input dimensions, $\sigma^2_{\epsilon}$ is the noise variance, and \textit{Depth} is the number of hidden layers of the \gls{MLP}. \label{table:uci_reg_rmse_full}}

\end{table}

\begin{table}[t]
	\centering
	\resizebox{1.0\columnwidth}{!}{
		\begin{tabular}{l||rrrr|cccccc}
			\toprule
			Data set & \textit{Classes} & $N_{\text{train}}$ & $N_{test}$ & $D$ & \fg prior        & \fgts            & \gpig prior               & \fh prior        & \gpih prior               & Deep Ensemble    \\
			\midrule
			\midrule
			\eeg     & 2                & 10980              & 4000       & 14  & 82.26 $\pm$ 7.17 & 81.63 $\pm$ 8.09 & 94.13 $\pm$ 1.96          & 93.31 $\pm$ 3.67 & \textbf{94.69} $\pm$ 2.17 & 89.94 $\pm$ 4.98 \\
			\htru    & 2                & 12898              & 5000       & 8   & 97.94 $\pm$ 0.23 & 97.93 $\pm$ 0.24 & \textbf{98.03} $\pm$ 0.24 & 98.01 $\pm$ 0.20 & 98.02 $\pm$ 0.26          & 98.01 $\pm$ 0.24 \\
			\magic   & 2                & 14020              & 5000       & 10  & 86.95 $\pm$ 0.39 & 87.15 $\pm$ 0.34 & 88.37 $\pm$ 0.29          & 87.65 $\pm$ 0.25 & \textbf{88.49} $\pm$ 0.26 & 87.87 $\pm$ 0.27 \\
			\miniboo & 2                & 120064             & 10000      & 50  & 90.81 $\pm$ 0.22 & 90.99 $\pm$ 0.21 & 92.74 $\pm$ 0.39          & 93.26 $\pm$ 0.28 & \textbf{93.37} $\pm$ 0.27 & 91.42 $\pm$ 0.21 \\
			\letter  & 26               & 15000              & 5000       & 16  & 90.45 $\pm$ 0.41 & 90.75 $\pm$ 0.37 & 96.90 $\pm$ 0.29          & 97.41 $\pm$ 0.26 & \textbf{97.67} $\pm$ 0.20 & 96.46 $\pm$ 0.27 \\
			\drive   & 11               & 48509              & 10000      & 48  & 98.55 $\pm$ 0.10 & 98.71 $\pm$ 0.09 & 99.69 $\pm$ 0.04          & 99.71 $\pm$ 0.04 & \textbf{99.74} $\pm$ 0.05 & 99.31 $\pm$ 0.06 \\
			\mocap   & 5                & 68095              & 10000      & 37  & 98.80 $\pm$ 0.10 & 98.98 $\pm$ 0.09 & 99.24 $\pm$ 0.10          & 99.41 $\pm$ 0.08 & \textbf{99.49} $\pm$ 0.07 & 99.12 $\pm$ 0.09 \\
			\bottomrule
		\end{tabular}
	}
	\vskip 0.15in
	\caption{Average test accuracy (\%) on \uci classification datasets (errors are $\pm 1$ standard error). Bold results indicate the best performance. Here, \textit{Classes} is the number of classes, $N_{\text{train}}$, $N_{test}$ is the sizes of training set and test set, respectively; $D$ is the number of input dimensions. \label{table:uci_class_accuracy_full}}
\end{table}

\clearpage
\begin{table}[t]
	\centering
	\footnotesize
	\resizebox{1.0\columnwidth}{!}{
		\begin{tabular}{l||rr|r|c|cccccc}
			\toprule
			Data set                   & $N$                    & $D$                 & $\sigma^2_{\epsilon}$  & \textit{Depth} & \fg prior          & \fgts              & \gpig  prior                & \fh prior                   & \gpih prior                 & Deep Ensemble               \\
			\midrule
			\midrule
			\multirow{4}{*}{\boston}   & \multirow{4}{*}{506}   & \multirow{4}{*}{13} & \multirow{4}{*}{0.1}   & 1              & 2.558 $\pm$ 0.294  & 2.582 $\pm$ 0.365  & 2.472 $\pm$ 0.153           & 2.498 $\pm$ 0.212           & \textbf{2.469} $\pm$ 0.160  & 3.177 $\pm$ 1.188           \\
			                           &                        &                     &                        & 2              & 2.541 $\pm$ 0.251  & 2.563 $\pm$ 0.343  & 2.475 $\pm$ 0.115           & 2.489 $\pm$ 0.196           & \textbf{2.458} $\pm$ 0.110  & 3.249 $\pm$ 1.111           \\
			                           &                        &                     &                        & 4              & 2.548 $\pm$ 0.207  & 2.542 $\pm$ 0.304  & 2.475 $\pm$ 0.095           & \textbf{2.473} $\pm$ 0.140  & 2.486 $\pm$ 0.080           & 3.448 $\pm$ 1.483           \\
			                           &                        &                     &                        & 8              & 2.581 $\pm$ 0.170  & 2.541 $\pm$ 0.259  & \textbf{2.474} $\pm$ 0.094  & 2.496 $\pm$ 0.128           & 2.529 $\pm$ 0.083           & 3.004 $\pm$ 0.915           \\
			\midrule

			\multirow{4}{*}{\concrete} & \multirow{4}{*}{1030}  & \multirow{4}{*}{8}  & \multirow{4}{*}{0.1}   & 1              & 3.104 $\pm$ 0.039  & 3.106 $\pm$ 0.048  & \textbf{3.004} $\pm$ 0.050  & 3.027 $\pm$ 0.051           & 3.007 $\pm$ 0.057           & 3.113 $\pm$ 0.214           \\
			                           &                        &                     &                        & 2              & 3.114 $\pm$ 0.037  & 3.099 $\pm$ 0.054  & 3.028 $\pm$ 0.050           & 3.066 $\pm$ 0.036           & \textbf{3.024} $\pm$ 0.044  & 3.065 $\pm$ 0.259           \\
			                           &                        &                     &                        & 4              & 3.145 $\pm$ 0.040  & 3.091 $\pm$ 0.055  & 3.060 $\pm$ 0.037           & 3.127 $\pm$ 0.039           & \textbf{3.056} $\pm$ 0.046  & 3.034 $\pm$ 0.251           \\
			                           &                        &                     &                        & 8              & 3.184 $\pm$ 0.047  & 3.092 $\pm$ 0.058  & 3.128 $\pm$ 0.046           & 3.169 $\pm$ 0.042           & 3.109 $\pm$ 0.037           & \textbf{3.054} $\pm$ 0.189  \\
			\midrule

			\multirow{4}{*}{\energy}   & \multirow{4}{*}{768}   & \multirow{4}{*}{8}  & \multirow{4}{*}{0.001} & 1              & 0.496 $\pm$ 0.216  & 0.496 $\pm$ 0.222  & \textbf{0.417} $\pm$ 0.227  & 0.489 $\pm$ 0.210           & 0.425 $\pm$ 0.210           & 2.076 $\pm$ 0.500           \\
			                           &                        &                     &                        & 2              & 0.471 $\pm$ 0.174  & 0.454 $\pm$ 0.196  & \textbf{0.347} $\pm$ 0.150  & 0.648 $\pm$ 0.116           & 0.392 $\pm$ 0.180           & 2.062 $\pm$ 1.014           \\
			                           &                        &                     &                        & 4              & 0.558 $\pm$ 0.145  & 0.506 $\pm$ 0.180  & 0.478 $\pm$ 0.168           & 0.681 $\pm$ 0.080           & \textbf{0.476} $\pm$ 0.166  & 1.935 $\pm$ 0.981           \\
			                           &                        &                     &                        & 8              & 0.636 $\pm$ 0.123  & 0.585 $\pm$ 0.134  & 0.657 $\pm$ 0.154           & 0.867 $\pm$ 0.056           & \textbf{0.562} $\pm$ 0.152  & 1.713 $\pm$ 0.736           \\
			\midrule

			\multirow{4}{*}{\kinm}     & \multirow{4}{*}{8192}  & \multirow{4}{*}{8}  & \multirow{4}{*}{0.1}   & 1              & -1.233 $\pm$ 0.018 & -1.238 $\pm$ 0.017 & -1.238 $\pm$ 0.015          & -1.243 $\pm$ 0.016          & -1.241 $\pm$ 0.015          & \textbf{-1.317} $\pm$ 0.061 \\
			                           &                        &                     &                        & 2              & -1.227 $\pm$ 0.018 & -1.243 $\pm$ 0.017 & -1.233 $\pm$ 0.012          & -1.230 $\pm$ 0.016          & -1.241 $\pm$ 0.014          & \textbf{-1.317} $\pm$ 0.076 \\
			                           &                        &                     &                        & 4              & -1.201 $\pm$ 0.013 & -1.235 $\pm$ 0.015 & -1.219 $\pm$ 0.011          & -1.180 $\pm$ 0.015          & -1.223 $\pm$ 0.013          & \textbf{-1.256} $\pm$ 0.074 \\
			                           &                        &                     &                        & 8              & -1.169 $\pm$ 0.015 & -1.222 $\pm$ 0.014 & -1.159 $\pm$ 0.020          & -1.138 $\pm$ 0.015          & -1.211 $\pm$ 0.013          & \textbf{-1.264} $\pm$ 0.070 \\
			\midrule

			\multirow{4}{*}{\naval}    & \multirow{4}{*}{11934} & \multirow{4}{*}{16} & \multirow{4}{*}{0.001} & 1              & -6.943 $\pm$ 0.028 & -6.935 $\pm$ 0.028 & -6.944 $\pm$ 0.031          & \textbf{-6.946} $\pm$ 0.028 & -6.923 $\pm$ 0.062          & -5.172 $\pm$ 0.227          \\
			                           &                        &                     &                        & 2              & -6.410 $\pm$ 0.087 & -6.373 $\pm$ 0.099 & \textbf{-6.430} $\pm$ 0.156 & -6.429 $\pm$ 0.097          & -6.397 $\pm$ 0.098          & -5.248 $\pm$ 0.274          \\
			                           &                        &                     &                        & 4              & -6.289 $\pm$ 0.079 & -6.291 $\pm$ 0.064 & \textbf{-6.359} $\pm$ 0.063 & -6.323 $\pm$ 0.043          & -6.347 $\pm$ 0.051          & -5.122 $\pm$ 0.259          \\
			                           &                        &                     &                        & 8              & -5.869 $\pm$ 0.046 & -5.893 $\pm$ 0.042 & -5.886 $\pm$ 0.040          & -5.926 $\pm$ 0.051          & \textbf{-5.895} $\pm$ 0.051 & -4.934 $\pm$ 0.428          \\
			\midrule

			\multirow{4}{*}{\power}    & \multirow{4}{*}{9568}  & \multirow{4}{*}{4}  & \multirow{4}{*}{0.05}  & 1              & 2.807 $\pm$ 0.042  & 2.807 $\pm$ 0.043  & \textbf{2.780} $\pm$ 0.044  & 2.812 $\pm$ 0.042           & 2.790 $\pm$ 0.043           & 2.799 $\pm$ 0.045           \\
			                           &                        &                     &                        & 2              & 2.808 $\pm$ 0.043  & 2.806 $\pm$ 0.044  & \textbf{2.738} $\pm$ 0.042  & 2.819 $\pm$ 0.040           & 2.761 $\pm$ 0.043           & 2.754 $\pm$ 0.053           \\
			                           &                        &                     &                        & 4              & 2.821 $\pm$ 0.039  & 2.809 $\pm$ 0.042  & 2.766 $\pm$ 0.040           & 2.844 $\pm$ 0.035           & 2.762 $\pm$ 0.041           & \textbf{2.738} $\pm$ 0.059  \\
			                           &                        &                     &                        & 8              & 2.833 $\pm$ 0.036  & 2.817 $\pm$ 0.040  & 2.821 $\pm$ 0.043           & 2.857 $\pm$ 0.032           & 2.783 $\pm$ 0.038           & \textbf{2.753} $\pm$ 0.037  \\
			\midrule

			\multirow{4}{*}{\protein}  & \multirow{4}{*}{45730} & \multirow{4}{*}{9}  & \multirow{4}{*}{0.5}   & 1              & 2.894 $\pm$ 0.004  & 2.894 $\pm$ 0.003  & 2.798 $\pm$ 0.002           & 2.809 $\pm$ 0.003           & 2.799 $\pm$ 0.004           & \textbf{2.753} $\pm$ 0.009  \\
			                           &                        &                     &                        & 2              & 2.892 $\pm$ 0.004  & 2.881 $\pm$ 0.005  & 2.752 $\pm$ 0.004           & 2.760 $\pm$ 0.004           & \textbf{2.748} $\pm$ 0.004  & 2.796 $\pm$ 0.016           \\
			                           &                        &                     &                        & 4              & 2.916 $\pm$ 0.003  & 2.875 $\pm$ 0.004  & 2.825 $\pm$ 0.011           & 2.801 $\pm$ 0.007           & 2.764 $\pm$ 0.004           & \textbf{2.606} $\pm$ 0.039  \\
			                           &                        &                     &                        & 8              & 2.919 $\pm$ 0.004  & 2.883 $\pm$ 0.003  & 2.933 $\pm$ 0.025           & 2.838 $\pm$ 0.006           & 2.802 $\pm$ 0.004           & \textbf{2.658} $\pm$ 0.013  \\
			\midrule

			\multirow{4}{*}{\wine}     & \multirow{4}{*}{1599}  & \multirow{4}{*}{11} & \multirow{4}{*}{0.5}   & 1              & 0.973 $\pm$ 0.080  & 0.983 $\pm$ 0.090  & \textbf{0.929} $\pm$ 0.067  & 0.962 $\pm$ 0.079           & 0.936 $\pm$ 0.069           & 1.008 $\pm$ 0.162           \\
			                           &                        &                     &                        & 2              & 0.983 $\pm$ 0.082  & 0.990 $\pm$ 0.087  & \textbf{0.915} $\pm$ 0.063  & 0.974 $\pm$ 0.082           & 0.922 $\pm$ 0.067           & 1.081 $\pm$ 0.193           \\
			                           &                        &                     &                        & 4              & 0.999 $\pm$ 0.085  & 1.004 $\pm$ 0.090  & 0.915 $\pm$ 0.064           & 0.973 $\pm$ 0.081           & \textbf{0.908} $\pm$ 0.064  & 1.774 $\pm$ 0.468           \\
			                           &                        &                     &                        & 8              & 1.016 $\pm$ 0.093  & 1.023 $\pm$ 0.095  & 0.953 $\pm$ 0.075           & 0.988 $\pm$ 0.084           & 0.938 $\pm$ 0.072           & \textbf{0.927} $\pm$ 0.100  \\

			\bottomrule
		\end{tabular}
	}
	\vskip 0.15in
	\caption{Average test \gls{NLL} in nats on \uci regression datasets (errors are $\pm 1$ standard error). Bold results indicate the best performance. Here, $N$ is the size of dataset, $D$ is the number of input dimensions, $\sigma^2_{\epsilon}$ is the noise variance, and \textit{Depth} is the number of hidden layers of the \gls{MLP}. \label{table:uci_reg_nll_full}
	}
\end{table}

\begin{table}[t]
	\centering
	\resizebox{1.0\columnwidth}{!}{
		\begin{tabular}{l||rrrr|cccccc}
			\toprule
			Data set & \textit{Classes} & $N_{\text{train}}$ & $N_{test}$ & $D$ & \fg prior         & \fgts             & \gpig prior       & \fh prior         & \gpih prior                & Deep Ensemble     \\
			\midrule
			\midrule
			\eeg     & 2                & 10980              & 4000       & 14  & 0.404 $\pm$ 0.120 & 0.406 $\pm$ 0.129 & 0.179 $\pm$ 0.046 & 0.179 $\pm$ 0.075 & \textbf{0.150} $\pm$ 0.053 & 0.240 $\pm$ 0.097 \\
			\htru    & 2                & 12898              & 5000       & 8   & 0.071 $\pm$ 0.007 & 0.072 $\pm$ 0.007 & 0.066 $\pm$ 0.008 & 0.068 $\pm$ 0.007 & \textbf{0.066} $\pm$ 0.008 & 0.067 $\pm$ 0.008 \\
			\magic   & 2                & 14020              & 5000       & 10  & 0.316 $\pm$ 0.006 & 0.312 $\pm$ 0.005 & 0.286 $\pm$ 0.005 & 0.298 $\pm$ 0.004 & \textbf{0.284} $\pm$ 0.005 & 0.294 $\pm$ 0.005 \\
			\miniboo & 2                & 120064             & 10000      & 50  & 0.218 $\pm$ 0.004 & 0.215 $\pm$ 0.004 & 0.179 $\pm$ 0.007 & 0.168 $\pm$ 0.004 & \textbf{0.165} $\pm$ 0.004 & 0.207 $\pm$ 0.004 \\
			\letter  & 26               & 15000              & 5000       & 16  & 0.445 $\pm$ 0.008 & 0.409 $\pm$ 0.008 & 0.166 $\pm$ 0.006 & 0.128 $\pm$ 0.005 & \textbf{0.115} $\pm$ 0.005 & 0.147 $\pm$ 0.006 \\
			\drive   & 11               & 48509              & 10000      & 48  & 0.098 $\pm$ 0.002 & 0.088 $\pm$ 0.002 & 0.028 $\pm$ 0.001 & 0.023 $\pm$ 0.001 & \textbf{0.022} $\pm$ 0.001 & 0.049 $\pm$ 0.002 \\
			\mocap   & 5                & 68095              & 10000      & 37  & 0.060 $\pm$ 0.002 & 0.050 $\pm$ 0.002 & 0.032 $\pm$ 0.002 & 0.027 $\pm$ 0.001 & \textbf{0.021} $\pm$ 0.001 & 0.040 $\pm$ 0.002 \\
			\bottomrule
		\end{tabular}
	}
	\vskip 0.15in
	\caption{Average test \gls{NLL} in nats on \uci classification datasets (errors are $\pm 1$ standard error). Bold results indicate the best performance. Here, \textit{Classes} is the number of classes; the  $N_{\text{train}}$, $N_{test}$ is the sizes of training set and test set, respectively; $D$ is the number of input dimensions. \label{table:uci_class_nll_full}}
\end{table}

\subsection{Additional results with full-batch Hamiltonian Monte Carlo}
\cref{table:uci_hmc} shows a comparison between full-batch \gls{HMC} and \gls{SGHMC} using the \fg and our \gpig priors on small \uci regression datasets.
We use the \gls{NUTS} extension \citep{HoffmanG14} of \gls{HMC} with the NumPyro's implementation \citep{phan2019composable}.
\gls{NUTS} adaptively sets the trajectory length of HMC, which along with the adaptation of the mass matrix and the step size.
We have simulated $4$ chains with a burn-in phase of $200$ iterations and $200$ collected samples for each chain.
We see that \gls{SGHMC} performs remarkably similar to a carefully tuned \gls{HMC} algorithm, despite the discretization error.

\begin{table}[t]
	\centering
	\resizebox{0.7\columnwidth}{!}{
		\begin{tabular}{l||l|ll|ll}
			\toprule
			\multicolumn{1}{c||}{} & \multicolumn{1}{c|}{} & \multicolumn{2}{c|}{\fg prior} & \multicolumn{2}{c}{\gpig prior}                                                              \\
			Data set               & $\sigma^2_{\epsilon}$ & \multicolumn{1}{c}{\gls{HMC}}  & \multicolumn{1}{c|}{\sghmc}     & \multicolumn{1}{c}{\gls{HMC}} & \multicolumn{1}{c}{\sghmc} \\ \midrule \midrule
			\boston                & 0.1                   & 3.065 $\pm$ 1.006              & 3.093 $\pm$ 1.001               & 2.821 $\pm$ 0.907             & 2.835 $\pm$ 0.922          \\
			\concrete              & 0.1                   & 5.369 $\pm$ 0.294              & 5.488 $\pm$ 0.253               & 4.715 $\pm$ 0.431             & 4.801 $\pm$ 0.416          \\
			\energy                & 0.001                 & 0.386 $\pm$ 0.064              & 0.389 $\pm$ 0.062               & 0.339 $\pm$ 0.075             & 0.343 $\pm$ 0.071          \\
			\power                 & 0.05                  & 3.931 $\pm$ 0.165              & 4.008 $\pm$ 0.168               & 3.438 $\pm$ 0.201             & 3.723 $\pm$ 0.183          \\
			\wine                  & 0.5                   & 0.637 $\pm$ 0.043              & 0.641 $\pm$ 0.044               & 0.606 $\pm$ 0.046             & 0.609 $\pm$ 0.046          \\ \bottomrule
		\end{tabular}
	}
	\vskip 0.15in
	\caption{Average test \gls{RMSE} results of full-batch \gls{HMC} and \sghmc on \uci regression datasets (errors are $\pm 1$ standard error).
	We use a \gls{MLP} with two hidden layers of $100$ neurons.
	$\sigma^2_{\epsilon}$ is the noise variance. \label{table:uci_hmc}}
\end{table}

\subsection{Additional discussion on the optimization of Wasserstein distance}
\label{ssec:wd_opt}

In the \cref{alg:main}, we have opted to separate the two optimization procedures for the Lipschitz function $\phi_{\mbtheta}$ and the Wasserstein distance.
We acknowledge that the two could have been optimized jointly in a single loop, as \cref{eq:objective_function} defines a minimax problem.
However, our choice allows $\phi_{\mbtheta}$ to be stabilized before a single Wasserstein minimization step takes place.
In fact, this is a common trick to make convergence more stable (see e.g., the original \cite{goodfellow2014generative} paper, which suggests to allow more training of the discriminator for each step of the generator).
\cref{fig:convergence_uci_mmd} illustrates the convergence behavior of these two algorithmic choices measured by the squared \gls{MMD} between the target \gls{GP} prior and the optimized \gls{BNN} prior on the \uci regression datasets (see \cref{ssec:mmd_appendix} for the experimental protocol).
Our optimization strategy demonstrates a much more stable convergence compared to the joint optimization approach.

\setlength\figureheight{.225\textwidth}
\setlength\figurewidth{.33\textwidth}
\definecolor{color1}{rgb}{0.8359375,0.15234375,0.15625}
\definecolor{color2}{rgb}{0.49609375,0.49609375,0.49609375}
\begin{figure}[th!]
	\centering
	\scriptsize
	\includegraphics{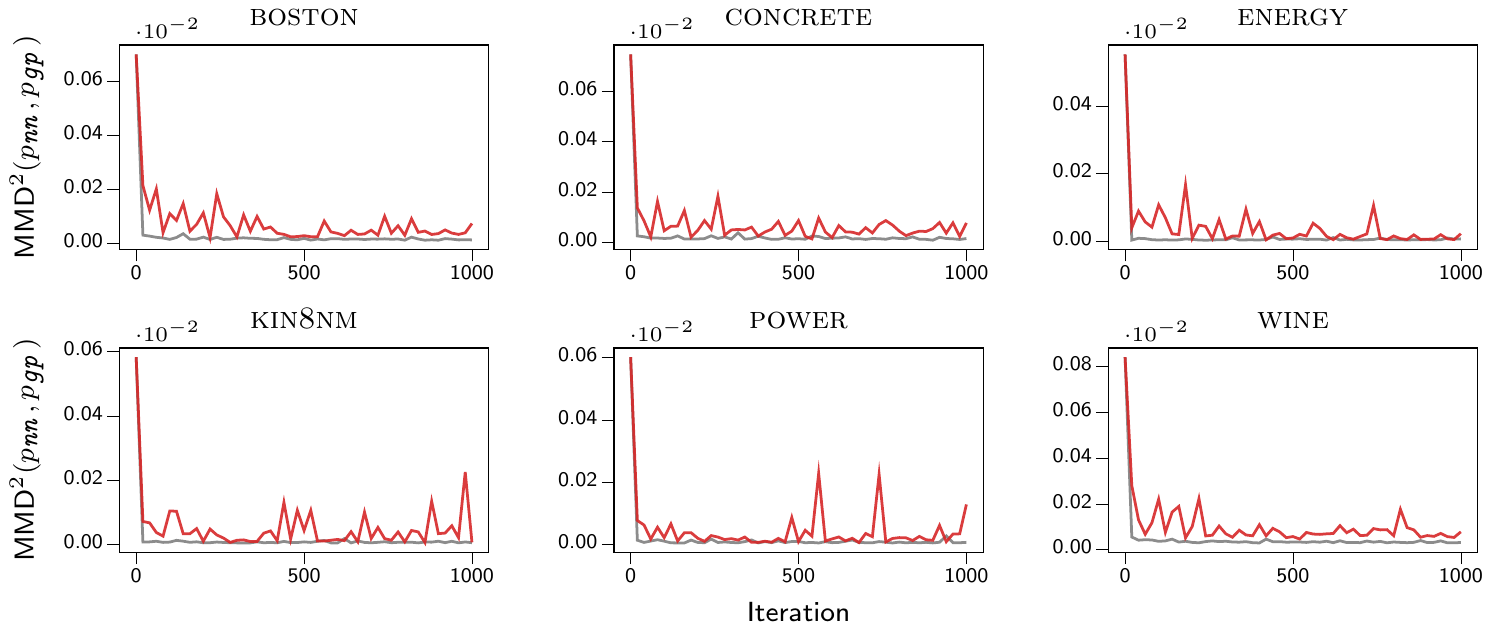}
	\vspace{-2ex}
	\caption{Comparison between strategies to optimize the Lipschitz function and the Wasserstein distance: (\protect\tikz[baseline=-.65ex] \protect\draw[line width=3pt, draw=color2] plot (0, 0) --+(.5, 0);) our strategy of separating these two operations; and (\protect\tikz[baseline=-.65ex] \protect\draw[line width=3pt, draw=color1] plot (0, 0) --+(.5, 0);) the strategy of joint optimization.
		Here, the convergence is measured by the squared \gls{MMD} between the target \gls{GP} prior and the optimized \gls{BNN} prior.\label{fig:convergence_uci_mmd}}
\end{figure}

\clearpage

\setlength\figureheight{.20\textwidth}
\setlength\figurewidth{.30\textwidth}

\clearpage

\setlength\figureheight{.20\textwidth}
\setlength\figurewidth{.30\textwidth}

\setlength\figureheight{.20\textwidth}
\setlength\figurewidth{.30\textwidth}

\section{A primer on Wasserstein Distance} \label{sec:primer_of_wasserstein_dist}
Given two {\it Borel's probability measures} $\pi(\mbx)$ and $\nu(\mby)$ defined on the {\it Polish space} $\cX$ and $\cY$ (i.e. any complete separable metric space such as a subset of $\bbR^d$), the {\bf $p$-Wasserstein distance} is defined as follows
\begin{align}
    W_p(\pi, \nu) = \left(\inf_{\gamma \in \Gamma(\pi, \nu)} \int_{\cX\times\cY}D(\mbx,\mby)^p \gamma(\mbx,\mby) \d{\mbx}\d{\mby}  \right)^{1/p}\,,
\end{align}
where $D(\mbx,\mby)$ is a proper distance metric between two points $\mbx$ and $\mby$ in the space $\cX\times\cY$ and $\Gamma(\pi, \nu)$ is the set of functionals of all possible joint densities whose marginals are indeed $\pi$ and $\nu$.

When the space of $\mbx$ and $\mby$ coincides (i.e. $\mbx,\mby\in\cX\subseteq\bbR^d$), the most used formulation is the 1-Wasserstein distance with Euclidian norm as distance,
\begin{align}
    \label{eq:wasserstein-1-append}
    W(\pi, \nu) = \inf_{\gamma \in \Gamma(\pi, \nu)} \int_{\cX\times\cX}\|\mbx - \mby\| \gamma(\mbx,\mby) \d{\mbx}\d{\mby}\,,
\end{align}
This is also known in the literature as the Earth-Mover distance.
Intuitively, here $\gamma$ measures how much mass must be transported from $\mbx$ to $\mby$ in order to transform the distributions $\pi$ into the distribution $\nu$.
Solving the Wasserstein distance means computing the minimum mass that needs to be moved. The question ``How?'' is answered by looking at the optimal transport plan (not the focus of these notes).

The remaining part of these notes will be dedicated to the proof of the dual formulation for \cref{eq:wasserstein-1-append}.
It is well known in the literature of optimization that linear programming problem with convex constrains admits a dual formulation.
Kantorovich introduced the dual formulation of the Wasserstein distance in 1942.

\begin{theorem}
    On the same setup as before, the Wasserstein distance defined as
    \begin{align}
        \label{eq:theo-wasserstein}
        W(\pi, \nu) = \inf_{\gamma \in \Gamma(\pi, \nu)} \int_{\cX\times\cX}\|\mbx - \mby\| \gamma(\mbx,\mby) \d{\mbx}\d{\mby}\,,
    \end{align}
    admits the following dual form
    \begin{align}
        \label{eq:theo-wasserstein-dual-appendix}
        W(\pi, \nu) = \sup_{\|f\|_L\leq1} \int_\cX f(\mbx)\pi(\mbx)d\mbx - \int_\cX f(\mby)\nu(\mby)d\mby
    \end{align}
    where $f$ is a 1-Lipschitz continuous function defined on $\cX\rightarrow\bbR$.
\end{theorem}

\subsection*{Step 1: Kantorovich duality}
First of all we start with the \textbf{Kantorovich duality}, which defines a dual form for the generic 1-Wasserstein.
\begin{theorem}
    Given a nonnegative measurable function $D:\cX\times\cX\rightarrow\bbR$, the 1-Wasserstein is computed as follows,
    \begin{align}
        W(\pi, \nu) = \inf_{\gamma \in \Gamma(\pi, \nu)} \int D(\mbx,\mby) \gamma(\mbx,\mby) \d{\mbx}\d{\mby}  \,,
    \end{align}
    The Kantorovich duality proves that this is equal to the following constrained optimization problem,
    \begin{align}
        W(\pi, \nu) = \sup_{\substack{f,g \\f(\mbx)+g(\mby)\leq D(\mbx, \mby)}} \int f(\mbx)\pi(\mbx)\d{\mbx} + \int g(\mby)\nu(\mby)\d{\mby}\,.
    \end{align}
\end{theorem}

We define $\iota_\Gamma(\gamma)$ the following quantity
\begin{align}
    \iota_\Gamma(\gamma) & = \sup_{f,g} \left[ \int f(\mbx)\pi(\mbx)\d{\mbx} + \int g(\mby)\nu(\mby)\d{\mby} - \iint \left[f(\mbx)+g(\mby)\right]\gamma(\mbx, \mby)  \d{\mbx}\d{\mby} \right] \nonumber
    \end{align}
and we observe that
\begin{align}
    \iota_\Gamma(\gamma) = \left\{
    \begin{aligned}
        0 \quad       & \text{if}\;\gamma\in\Gamma(\pi, \nu)\,, \\
        +\infty \quad & \text{otherwise}\,.
    \end{aligned}
    \right. \nonumber
\end{align}

This is true because given the definition of $\Gamma$, if $\gamma\in\Gamma(\pi, \nu)$ then $\pi(\mbx)=\int \gamma(\mbx,\mby)\d{\mby}$ and $\nu(\mby)=\int \gamma(\mbx,\mby)\d{\mbx}$.
By substituiting these quantities, it follows that
\begin{align}
    \int f(\mbx)\pi(\mbx)\d{\mbx} + \int g(\mby)\nu(\mby)\d{\mby} & = \int f(\mbx)\int \gamma(\mbx,\mby)\d{\mby}\d{\mbx} + \int g(\mby)\int \gamma(\mbx,\mby)\d{\mbx}\d{\mby}  \nonumber \\
                                                                  & = \iint \left[f(\mbx)+g(\mby)\right]\gamma(\mbx, \mby) \d{\mbx}\d{\mby} \,. \nonumber                                \end{align}
In other cases, $f$ and $g$ can be chosen such that the supremum becomes $+\infty$.
Given this property and the constrain on $\gamma$, we can add $\iota_\Gamma(\gamma)$ to the formulation of the Wasserstein distance in \cref{eq:theo-wasserstein},
\begin{align}
     & W(\pi, \nu)  = \inf_{\gamma \in \Gamma(\pi, \nu)} \left[\int D(\mbx, \mby) \gamma(\mbx,\mby) \d{\mbx}\d{\mby}\right] + \iota_\Gamma(\gamma) =                                                                                                                  \nonumber \\
     & = \inf_{\gamma} \left[\int D(\mbx, \mby) \gamma(\mbx,\mby) \d{\mbx}\d{\mby} + \sup_{f,g} \left[ \int f(\mbx)\pi(\mbx)\d{\mbx} + \int g(\mby)\nu(\mby)\d{\mby} - \right.\right.\\
     &\qquad\qquad\left.\left. \iint \left[f(\mbx)+g(\mby)\right]\gamma(\mbx, \mby)  \d{\mbx}\d{\mby} \right]\right]\,, \nonumber
\end{align}
Now, the original integral of the Wasserstein distance does not depend on $f$ and $g$; therefore the supremum can be moved in front,
\begin{align}
    W(\pi, \nu) &= \inf_{\gamma} \sup_{f,g} \Upsilon(\gamma, (f, g)) \\
    \Upsilon(\gamma, (f, g)) &\defeq 
     \int D(\mbx, \mby) \gamma(\mbx,\mby) \d{\mbx}\d{\mby} +  \int f(\mbx)\pi(\mbx)\d{\mbx} + \int g(\mby)\nu(\mby)\d{\mby} - \nonumber\\
     &\qquad \iint \left[f(\mbx)+g(\mby)\right]\gamma(\mbx, \mby)  \d{\mbx}\d{\mby} \nonumber
\end{align}
Under certain conditions stated by the \textit{minimax theorem}, i.e. $\Upsilon(\gamma, (f, g))$ is convex-concave function ($\Upsilon$ is concave for fixed $(f,g)$ while convex for fixed $\gamma$ ), we can swap the infinum and the supremum and rewrite the definition as follows,
\begin{align}
    W(\pi, \nu) = \sup_{f,g}{\inf_{\gamma} \int \left[ D(\mbx, \mby) - f(\mbx)-g(\mby)\right] \gamma(\mbx,\mby) \d{\mbx}\d{\mby}} +  \int f(\mbx)\pi(\mbx)\d{\mbx} + \int g(\mby)\nu(\mby)\d{\mby}\nonumber
\end{align}
Proofs that the hypothesis used for the minimax theorem hold for this case are presented in Theorem 1.9 of ``Topics in Optimal Transport'' \citep{Villani2003}.
Focusing on the infimum part, we can write
\begin{align}
    \inf_{\gamma} \int \left[ D(\mbx, \mby) - f(\mbx)-g(\mby)\right] \gamma(\mbx,\mby) \d{\mbx}\d{\mby} = \left\{
    \begin{aligned}
        0  \quad      & \text{if}\;{f(\mbx)+g(\mby)\leq  D(\mbx, \mby)} \,, \\
        -\infty \quad & \text{otherwise}\,.
    \end{aligned}
    \right.\nonumber
\end{align}
If the function $\zeta(\mbx, \mby) =  D(\mbx, \mby) - (f(\mbx)+g(\mby))$ takes a negative value at some point $(\mbx_0, \mby_0)$, then by choosing $\gamma=\lambda\delta(\mbx_0, \mby_0)$ with $\lambda\rightarrow+\infty$ (i.e. a Dirac delta in $(\mbx_0, \mby_0)$), we see that the infimum is infinite.
On the other hand, is $\zeta(\mbx, \mby)$ is nonnegative, then the infimum is obtained for $\gamma=0$.
Finally, this constrains can be added to the previous conditions making thus recovering the formulation in \cref{eq:theo-wasserstein-dual-appendix}.

\subsection*{Step 2: D-Transforms}
The next challenge is to find $f$ and $g$ such that we can easily recover the constrain optimization above.
We approach this problem by supposing to have chosen some $f(\mbx)$. This means that the objective is to find a good $g(\mby)$ that for all $\mbx,\mby$ satisfy the condition
\begin{align}
    f(\mbx) + g(\mby) \leq D(\mbx, \mby)\,.\nonumber
\end{align}
The trivial solution is $g(\mby) \leq D(\mbx, \mby) - f(\mbx)$.
This must be true for all $\mbx$, also in the worst case (when we take the infimum),
\begin{align}
    g(\mby) \leq \inf_\mbx \left[D(\mbx, \mby) - f(\mbx)\right].\nonumber
\end{align}
At this point, we observe that for a given $f$, if we want the supremum in Eq. 5 we cannot get a better $g$ then taking the equality,
\begin{align}
    \bar f(\mby) := \inf_\mbx \left[D(\mbx, \mby) - f(\mbx)\right].\nonumber
\end{align}
We therefore have the following formulation of the Wasserstein distance,
\begin{align}
    W(\pi, \nu) = \sup_{f}\left[\int f(\mbx)\pi(\mbx)\d{\mbx} + \int \bar f(\mby)\nu(\mby)\d{\mby} \right]\nonumber
\end{align}
If now we suppose to choose $g$, by following the same reasoning the best $f$ that we can get is defined
\begin{align}
    \bar{\bar f}(\mbx) = \bar g(\mbx) := \inf_\mby \left[D(\mbx, \mby) - g(\mby)\right].\nonumber
\end{align}
If we replace $g(\mby)$ with Eq. 17 we have yet another recursive definition of the Wasserstein distance,
\begin{align}
    W(\pi, \nu) = \sup_{f}\left[\int \bar{\bar f}(\mbx)\pi(\mbx)\d{\mbx} + \int \bar f(\mby)\nu(\mby)\d{\mby} \right]\nonumber
\end{align}
If we constrain $f$ to be $D$-concave, then $\bar{\bar f} = f$.

\subsection*{Step 2.1: Euclidean distance}
It's worth mentioning that this formulation is valid for any nonnegative measurable function $D$.
For the Euclidian distance this simplify even further.
\begin{theorem}
    When $D(\mbx, \mby) = \|\mbx - \mby\|$ and $f$ is 1-Lipschitz, $f$ is $D$-concave if and only if $\bar f = -f$
\end{theorem}
We prove the necessity condition of such result.
First of all, we observe that if $f$ is 1-Lipschitz then $\bar f$ is 1-Lipschitz too. This is true because for any given $\mbx$
\begin{align}
    \bar f_\mbx(\mby) = \| \mbx - \mby \| - f(\mbx)\nonumber
\end{align}
is 1-Lipschitz and therefore the infimum of $\bar f(\mby) = \inf_{\mbx} \| \mbx - \mby \| - f(\mbx)$ is 1-Lipschitz.
Since $\bar f$ is 1-Lipschitz, for all $\mbx$ and $\mby$ we have
\begin{align}
                & \left|\bar f(\mby) - \bar f(\mbx)\right| \leq \|\mby - \mbx\|                   \nonumber \\
    \implies  - & \bar f(\mbx) \leq \|\mbx-\mby\| - \bar f(\mby)\nonumber
\end{align}
Since this is true for all $\mby$,
\begin{align}
     & - \bar f(\mbx) \leq \inf_\mby\|\mbx-\mby\| - \bar f(\mby)\,                                                     \nonumber  \\
     & - \bar f(\mbx) \leq \underbrace{\inf_\mby\|\mbx-\mby\| - \bar f(\mby)}_{\bar{\bar f}\equiv f} \leq - \bar f(\mbx)\nonumber
\end{align}
where the right inequality follows by choosing $\mby = \mbx$ in the infimum.
We know that $\bar{\bar f}\equiv f$. This means that $-\bar f(\mbx)$ must be equal to $f(\mbx)$ for the last equation to hold.

\subsection*{Step 3. Putting everything together}
We started our discussion by proving the Kantovich duality, which states that
\begin{align}
        \inf_{\gamma \in \Gamma(\pi, \nu)} \int D(\mbx,\mby) \gamma(\mbx,\mby) \d{\mbx}\d{\mby} = \sup_{\substack{f,g \\f(\mbx)+g(\mby)\leq D(\mbx, \mby)}} \int f(\mbx)\pi(\mbx)\d{\mbx} + \int g(\mby)\nu(\mby)\d{\mby}\,,\nonumber
    \end{align}

\noindent We then proved that
\begin{align*}
        \sup_{\substack{f,g \\f(\mbx)+g(\mby)\leq D(\mbx, \mby)}} \left[\int f(\mbx)\pi(\mbx)\d{\mbx} + \int g(\mby)\nu(\mby)\d{\mby}\right] = \nonumber\\
        = \sup_{\substack{f   \\\bar f= \inf_\mbx D - f}}\left[\int f(\mbx)\pi(\mbx)\d{\mbx} + \int \bar f(\mby)\nu(\mby)\d{\mby} \right] \,,\nonumber
    \end{align*}

\noindent Finally, given $D(\mbx, \mby)$ to be the Euclidean distance, we discussed the shape of $\bar f$ when we restrict $f$ to be 1-Lipschitz, showing that $\bar f = -f$.
Putting everything together, we obtain the dual 1-Wasserstein distance in \cref{eq:theo-wasserstein-dual-appendix},
\begin{align}
        W(\pi, \nu) = \sup_{\|f\|_L\leq1} \int f(\mbx)\pi(\mbx)d\mbx - \int f(\mby)\nu(\mby)d\mby\nonumber
    \end{align}


\vskip 0.2in
\begin{thebibliography}{94}
\providecommand{\natexlab}[1]{#1}
\providecommand{\url}[1]{\texttt{#1}}
\expandafter\ifx\csname urlstyle\endcsname\relax
  \providecommand{\doi}[1]{doi: #1}\else
  \providecommand{\doi}{doi: \begingroup \urlstyle{rm}\Url}\fi

\bibitem[Amit and Meir(2018)]{Amit2018}
R.~Amit and R.~Meir.
\newblock {Meta-Learning by Adjusting Priors Based on Extended PAC-Bayes
  Theory}.
\newblock In \emph{Proceedings of the 35th International Conference on Machine
  Learning}, volume~80 of \emph{Proceedings of Machine Learning Research},
  pages 205--214. {PMLR}, 2018.

\bibitem[Arjovsky et~al.(2017)Arjovsky, Chintala, and
  Bottou]{arjovsky2017wasserstein}
M.~Arjovsky, S.~Chintala, and L.~Bottou.
\newblock {Wasserstein Generative Adversarial Networks}.
\newblock In \emph{Proceedings of the 34th International Conference on Machine
  Learning}, volume~70 of \emph{Proceedings of Machine Learning Research},
  pages 214--223. {PMLR}, 2017.

\bibitem[Ashukha et~al.(2020)Ashukha, Lyzhov, Molchanov, and
  Vetrov]{ashukhapitfalls}
A.~Ashukha, A.~Lyzhov, D.~Molchanov, and D.~Vetrov.
\newblock {Pitfalls of In-Domain Uncertainty Estimation and Ensembling in Deep
  Learning}.
\newblock In \emph{International Conference on Learning Representations}, 2020.

\bibitem[Atanov et~al.(2019)Atanov, Ashukha, Struminsky, Vetrov, and
  Welling]{Atanov19}
A.~Atanov, A.~Ashukha, K.~Struminsky, D.~Vetrov, and M.~Welling.
\newblock {The Deep Weight Prior}.
\newblock In \emph{International Conference on Learning Representations}, 2019.

\bibitem[Bishop(2006)]{Bishop06}
C.~M. Bishop.
\newblock \emph{{Pattern recognition and machine learning}}.
\newblock Springer, 1st ed. 2006. corr. 2nd printing 2011 edition, Aug. 2006.

\bibitem[Blundell et~al.(2015)Blundell, Cornebise, Kavukcuoglu, and
  Wierstra]{blundell2015weight}
C.~Blundell, J.~Cornebise, K.~Kavukcuoglu, and D.~Wierstra.
\newblock {Weight Uncertainty in Neural Network}.
\newblock In \emph{International Conference on Machine Learning}, pages
  1613--1622. PMLR, 2015.

\bibitem[Briol et~al.(2019)Briol, Oates, Girolami, Osborne, and
  Sejdinovic]{Briol2019}
F.-X. Briol, C.~J. Oates, M.~Girolami, M.~A. Osborne, and D.~Sejdinovic.
\newblock {Probabilistic Integration: A Role in Statistical Computation?}
\newblock \emph{Statistical Science}, 34\penalty0 (1):\penalty0 1--22, 02 2019.

\bibitem[Chen et~al.(2014)Chen, Fox, and Guestrin]{Cheni2014}
T.~Chen, E.~Fox, and C.~Guestrin.
\newblock {Stochastic Gradient Hamiltonian Monte Carlo}.
\newblock In \emph{Proceedings of the 31st International Conference on Machine
  Learning}, Proceedings of Machine Learning Research, pages 1683--1691. PMLR,
  2014.

\bibitem[Chen et~al.(2017)Chen, Kingma, Salimans, Duan, Dhariwal, Schulman,
  Sutskever, and Abbeel]{Chen2017VariationalLA}
X.~Chen, D.~P. Kingma, T.~Salimans, Y.~Duan, P.~Dhariwal, J.~Schulman,
  I.~Sutskever, and P.~Abbeel.
\newblock {Variational Lossy Autoencoder}.
\newblock In \emph{International Conference on Learning Representations}, 2017.

\bibitem[Cockayne et~al.(2019)Cockayne, Oates, Ipsen, and
  Girolami]{Cockayne2019}
J.~Cockayne, C.~J. Oates, I.~C. Ipsen, and M.~Girolami.
\newblock {A Bayesian Conjugate Gradient Method (with Discussion)}.
\newblock \emph{Bayesian Analysis}, 14\penalty0 (3):\penalty0 937--1012, 09
  2019.

\bibitem[Daxberger et~al.(2021)Daxberger, Kristiadi, Immer, Eschenhagen, Bauer,
  and Hennig]{Daxberger2021LaplaceR}
E.~A. Daxberger, A.~Kristiadi, A.~Immer, R.~Eschenhagen, M.~Bauer, and
  P.~Hennig.
\newblock {Laplace Redux -- Effortless Bayesian Deep Learning}.
\newblock In \emph{Advances in Neural Information Processing Systems},
  volume~34, pages 20089--20103, 2021.

\bibitem[Delattre and Fournier(2017)]{Delattre2017}
S.~Delattre and N.~Fournier.
\newblock {On the Kozachenko–Leonenko entropy estimator}.
\newblock \emph{Journal of Statistical Planning and Inference}, 185:\penalty0
  69--93, 2017.

\bibitem[Dua and Graff(2017)]{asuncion2007uci}
D.~Dua and C.~Graff.
\newblock {UCI} machine learning repository.
\newblock University of California, Irvine, School of Information and Computer
  Sciences, 2017.
\newblock URL \url{http://archive.ics.uci.edu/ml}.

\bibitem[Duane et~al.(1987)Duane, Kennedy, Pendleton, and Roweth]{Duane1987}
S.~Duane, A.~Kennedy, B.~J. Pendleton, and D.~Roweth.
\newblock {Hybrid Monte Carlo}.
\newblock \emph{Physics Letters B}, 195\penalty0 (2):\penalty0 216 -- 222,
  1987.

\bibitem[Duchi et~al.(2011)Duchi, Hazan, and Singer]{duchi2011adaptive}
J.~Duchi, E.~Hazan, and Y.~Singer.
\newblock {Adaptive Subgradient Methods for Online Learning and Stochastic
  Optimization}.
\newblock \emph{Journal of Machine Learning Research}, 12\penalty0
  (61):\penalty0 2121--2159, 2011.

\bibitem[Duvenaud et~al.(2014)Duvenaud, Rippel, Adams, and
  Ghahramani]{Duvenaud14}
D.~Duvenaud, O.~Rippel, R.~Adams, and Z.~Ghahramani.
\newblock {Avoiding Pathologies in Very Deep Networks}.
\newblock In \emph{Proceedings of the 17th International Conference on
  Artificial Intelligence and Statistics}, volume~33 of \emph{Proceedings of
  Machine Learning Research}, pages 202--210. PMLR, 2014.

\bibitem[Flam-Shepherd et~al.(2017)Flam-Shepherd, Requeima, and
  Duvenaud]{Flam2017}
D.~Flam-Shepherd, J.~Requeima, and D.~Duvenaud.
\newblock {Mapping Gaussian Process Priors to Bayesian Neural Networks}.
\newblock In \emph{NeurIPS workshop on Bayesian Deep Learning}, 2017.

\bibitem[Flam-Shepherd et~al.(2018)Flam-Shepherd, Requeima, and
  Duvenaud]{Flam2018}
D.~Flam-Shepherd, J.~Requeima, and D.~Duvenaud.
\newblock {Characterizing and Warping the Function space of Bayesian Neural
  Networks}.
\newblock In \emph{NeurIPS workshop on Bayesian Deep Learning}, 2018.

\bibitem[Gal and Ghahramani(2016)]{Gal2016}
Y.~Gal and Z.~Ghahramani.
\newblock {Dropout as a Bayesian Approximation: Representing Model Uncertainty
  in Deep Learning}.
\newblock In \emph{Proceedings of the 33nd International Conference on Machine
  Learning}, volume~48 of \emph{Proceedings of Machine Learning Research},
  pages 1050--1059. JMLR, June 19-24 2016.

\bibitem[Gelman and Rubin(1992)]{Gelman92}
A.~Gelman and D.~B. Rubin.
\newblock {Inference from Iterative Simulation using Multiple Sequences}.
\newblock \emph{Statistical Science}, 7\penalty0 (4):\penalty0 457--472, 1992.

\bibitem[Goodfellow et~al.(2014)Goodfellow, Pouget-Abadie, Mirza, Xu,
  Warde-Farley, Ozair, Courville, and Bengio]{goodfellow2014generative}
I.~Goodfellow, J.~Pouget-Abadie, M.~Mirza, B.~Xu, D.~Warde-Farley, S.~Ozair,
  A.~Courville, and Y.~Bengio.
\newblock Generative adversarial nets.
\newblock In \emph{Advances in Neural Information Processing Systems},
  volume~27, pages 2672--2680. Curran Associates, Inc., 2014.

\bibitem[Graves(2011)]{Graves2011}
A.~Graves.
\newblock {Practical Variational Inference for Neural Networks}.
\newblock In \emph{Advances in Neural Information Processing Systems},
  volume~24, pages 2348--2356. Curran Associates, Inc., 2011.

\bibitem[Gretton et~al.(2012)Gretton, Borgwardt, Rasch, Sch{\"{o}}lkopf, and
  Smola]{GrettonBRSS12}
A.~Gretton, K.~M. Borgwardt, M.~J. Rasch, B.~Sch{\"{o}}lkopf, and A.~J. Smola.
\newblock {A Kernel Two-Sample Test}.
\newblock \emph{Journal of Machine Learning Research}, 13:\penalty0 723--773,
  2012.

\bibitem[Grover et~al.(2018)Grover, Dhar, and Ermon]{Grover2018}
A.~Grover, M.~Dhar, and S.~Ermon.
\newblock {Flow-GAN: Combining Maximum Likelihood and Adversarial Learning in
  Generative Models}.
\newblock In \emph{Proceedings of the 32nd Conference on Artificial
  Intelligence}, pages 3069--3076. {AAAI} Press, 2018.

\bibitem[Gulrajani et~al.(2017)Gulrajani, Ahmed, Arjovsky, Dumoulin, and
  Courville]{gulrajani2017improved}
I.~Gulrajani, F.~Ahmed, M.~Arjovsky, V.~Dumoulin, and A.~C. Courville.
\newblock {Improved Training of Wasserstein GANs}.
\newblock In \emph{Advances in Neural Information Processing Systems},
  volume~30, pages 5767--5777. Curran Associates, Inc., 2017.

\bibitem[Ha et~al.(2017)Ha, Dai, and Le]{Ha2017}
D.~Ha, A.~M. Dai, and Q.~V. Le.
\newblock Hypernetworks.
\newblock In \emph{International Conference on Learning Representations}, 2017.

\bibitem[Hafner et~al.(2019)Hafner, Tran, Lillicrap, Irpan, and
  Davidson]{Hafner2019}
D.~Hafner, D.~Tran, T.~P. Lillicrap, A.~Irpan, and J.~Davidson.
\newblock {Noise Contrastive Priors for Functional Uncertainty}.
\newblock In \emph{Proceedings of the 35h Conference on Uncertainty in
  Artificial Intelligence}, page 332. {AUAI} Press, 2019.

\bibitem[He et~al.(2016)He, Zhang, Ren, and Sun]{he2016identity}
K.~He, X.~Zhang, S.~Ren, and J.~Sun.
\newblock {Identity Mappings in Deep Residual Networks}.
\newblock In \emph{Proceeding of the 14th European Conference on Computer
  Vision}, volume 9908 (Part {IV}) of \emph{Lecture Notes in Computer Science},
  pages 630--645. Springer, 2016.

\bibitem[Heek and Kalchbrenner(2019)]{heek2019bayesian}
J.~Heek and N.~Kalchbrenner.
\newblock {Bayesian Inference for Large Scale Image Classification}.
\newblock arXiv:1908.03491, 2019.

\bibitem[Hendrycks and Dietterich(2019)]{hendrycks2018benchmarking}
D.~Hendrycks and T.~Dietterich.
\newblock {Benchmarking Neural Network Robustness to Common Corruptions and
  Perturbations}.
\newblock In \emph{International Conference on Learning Representations}, 2019.

\bibitem[Hoffman and Gelman(2014)]{HoffmanG14}
M.~D. Hoffman and A.~Gelman.
\newblock {The No-U-turn Sampler: Adaptively Setting Path Lengths in
  Hamiltonian Monte Carlo}.
\newblock \emph{Journal of Machine Learning Research}, 15\penalty0
  (1):\penalty0 1593--1623, 2014.

\bibitem[Houlsby et~al.(2012)Houlsby, Huszar, Ghahramani, and
  Hern\'{a}ndez-lobato]{houlsby2012collaborative}
N.~Houlsby, F.~Huszar, Z.~Ghahramani, and J.~Hern\'{a}ndez-lobato.
\newblock {Collaborative Gaussian Processes for Preference Learning}.
\newblock In \emph{Advances in Neural Information Processing Systems},
  volume~25, pages 2096--2104. Curran Associates, Inc., 2012.

\bibitem[Immer et~al.(2021{\natexlab{a}})Immer, Bauer, Fortuin, R{\"{a}}tsch,
  and Khan]{ImmerBFRK21}
A.~Immer, M.~Bauer, V.~Fortuin, G.~R{\"{a}}tsch, and M.~E. Khan.
\newblock {Scalable Marginal Likelihood Estimation for Model Selection in Deep
  Learning}.
\newblock In \emph{Proceedings of the 38th International Conference on Machine
  Learning}, volume 139 of \emph{Proceedings of Machine Learning Research},
  pages 4563--4573. {PMLR}, 2021{\natexlab{a}}.

\bibitem[Immer et~al.(2021{\natexlab{b}})Immer, Korzepa, and Bauer]{ImmerKB21}
A.~Immer, M.~Korzepa, and M.~Bauer.
\newblock {Improving Predictions of Bayesian Neural Nets via Local
  Linearization}.
\newblock In \emph{Proceedings of the 24th International Conference on
  Artificial Intelligence and Statistics}, volume 130 of \emph{Proceedings of
  Machine Learning Research}, pages 703--711. {PMLR}, 2021{\natexlab{b}}.

\bibitem[Jacot et~al.(2018)Jacot, Gabriel, and Hongler]{jacot2018neural}
A.~Jacot, F.~Gabriel, and C.~Hongler.
\newblock {Neural Tangent Kernel: Convergence and Generalization in Neural
  Networks}.
\newblock In \emph{Advances in Neural Information Processing Systems},
  volume~31, pages 8571--8580. Curran Associates, Inc., 2018.

\bibitem[Jankowiak and Obermeyer(2018)]{jankowiak2018pathwise}
M.~Jankowiak and F.~Obermeyer.
\newblock {Pathwise Derivatives Beyond the Reparameterization Trick}.
\newblock In \emph{Proceedings of the 35th International Conference on Machine
  Learning}, volume~80 of \emph{Proceedings of Machine Learning Research},
  pages 2240--2249. {PMLR}, 2018.

\bibitem[Kantorovich(1942)]{Kantorovich1942}
L.~V. Kantorovich.
\newblock {On the transfer of masses}.
\newblock \emph{Doklady Akademii Nauk SSSR}, 37:\penalty0 227--229, 1942.

\bibitem[Kantorovich(1948)]{Kantorovich1948}
L.~V. Kantorovich.
\newblock {On a problem of Monge}.
\newblock \emph{Uspekhi Matematicheskikh Nauk}, 3:\penalty0 225--226, 1948.

\bibitem[Karaletsos and Bui(2019)]{Karaletsos2019}
T.~Karaletsos and T.~D. Bui.
\newblock {Gaussian Process Meta-Representations For Hierarchical Neural
  Network Weight Priors}.
\newblock In \emph{2nd Symposium on Advances in Approximate Bayesian
  Inference}, 2019.

\bibitem[Karaletsos and Bui(2020)]{Karaletsos2020}
T.~Karaletsos and T.~D. Bui.
\newblock {Hierarchical Gaussian Process Priors for Bayesian Neural Network
  Weights}.
\newblock In \emph{Advances in Neural Information Processing Systems},
  volume~33, 2020.

\bibitem[Kendall and Gal(2017)]{Kendall2017}
A.~Kendall and Y.~Gal.
\newblock {What Uncertainties Do We Need in Bayesian Deep Learning for Computer
  Vision?}
\newblock In \emph{Advances in Neural Information Processing Systems},
  volume~30, pages 5574--5584. Curran Associates, Inc., 2017.

\bibitem[Khan et~al.(2019)Khan, Immer, Abedi, and Korzepa]{KhanIAK19}
M.~E. Khan, A.~Immer, E.~Abedi, and M.~Korzepa.
\newblock {Approximate Inference Turns Deep Networks into Gaussian Processes}.
\newblock In \emph{Advances in Neural Information Processing Systems}, pages
  3088--3098, 2019.

\bibitem[Kingma and Ba(2015)]{jlb2015adam}
D.~P. Kingma and J.~Ba.
\newblock {Adam: {A} Method for Stochastic Optimization}.
\newblock In \emph{International Conference on Learning Representations}, 2015.

\bibitem[Kingma and Welling(2014)]{Kingma14}
D.~P. Kingma and M.~Welling.
\newblock {Auto-Encoding Variational Bayes}.
\newblock In \emph{International Conference on Learning Representations}, 2014.

\bibitem[Kingma et~al.(2016)Kingma, Salimans, Jozefowicz, Chen, Sutskever, and
  Welling]{Kingma2016}
D.~P. Kingma, T.~Salimans, R.~Jozefowicz, X.~Chen, I.~Sutskever, and
  M.~Welling.
\newblock {Improved Variational Inference with Inverse Autoregressive Flow}.
\newblock In \emph{Advances in Neural Information Processing Systems},
  volume~29, pages 4743--4751. Curran Associates, Inc., 2016.

\bibitem[Krizhevsky and Hinton(2009)]{krizhevsky2009learning}
A.~Krizhevsky and G.~Hinton.
\newblock {Learning Multiple Layers of Features from Tiny Images}.
\newblock \emph{Master's thesis, Department of Computer Science, University of
  Toronto}, 2009.

\bibitem[Lakshminarayanan et~al.(2017)Lakshminarayanan, Pritzel, and
  Blundell]{lakshminarayanan2017simple}
B.~Lakshminarayanan, A.~Pritzel, and C.~Blundell.
\newblock {Simple and Scalable Predictive Uncertainty Estimation using Deep
  Ensembles}.
\newblock In \emph{Advances in Neural Information Processing Systems},
  volume~30, pages 6402--6413. Curran Associates, Inc., 2017.

\bibitem[LeCun et~al.(1998)LeCun, Bottou, Bengio, and
  Haffner]{lecun1998gradient}
Y.~LeCun, L.~Bottou, Y.~Bengio, and P.~Haffner.
\newblock Gradient-based learning applied to document recognition.
\newblock \emph{Proceedings of the IEEE}, 86\penalty0 (11):\penalty0
  2278--2324, 1998.

\bibitem[LeCun et~al.(2015)LeCun, Bengio, and Hinton]{LeCun2015}
Y.~LeCun, Y.~Bengio, and G.~Hinton.
\newblock Deep learning.
\newblock \emph{Nature}, 521\penalty0 (7553):\penalty0 436--444, May 2015.

\bibitem[Lee et~al.(2020)Lee, Schoenholz, Pennington, Adlam, Xiao, Novak, and
  Sohl-Dickstein]{lee2020finite}
J.~Lee, S.~S. Schoenholz, J.~Pennington, B.~Adlam, L.~Xiao, R.~Novak, and
  J.~Sohl-Dickstein.
\newblock {Finite Versus Infinite Neural Networks: an Empirical Study}.
\newblock In \emph{Advances in Neural Information Processing Systems},
  volume~33, 2020.

\bibitem[{Liu} et~al.(2020){Liu}, {Ong}, {Shen}, and {Cai}]{Liu2020}
H.~{Liu}, Y.~S. {Ong}, X.~{Shen}, and J.~{Cai}.
\newblock {When Gaussian Process Meets Big Data: A Review of Scalable GPs}.
\newblock \emph{IEEE Transactions on Neural Networks and Learning Systems},
  31\penalty0 (11):\penalty0 4405--4423, 2020.

\bibitem[Liu and Wang(2016)]{Liu2016}
Q.~Liu and D.~Wang.
\newblock {Stein Variational Gradient Descent: A General Purpose Bayesian
  Inference Algorithm}.
\newblock In \emph{Advances in Neural Information Processing Systems},
  volume~29, pages 2378--2386. Curran Associates, Inc., 2016.

\bibitem[Louizos and Welling(2017)]{Louizos2017}
C.~Louizos and M.~Welling.
\newblock {Multiplicative Normalizing Flows for Variational {B}ayesian Neural
  Networks}.
\newblock In \emph{Proceedings of the 34th International Conference on Machine
  Learning}, volume~70 of \emph{Proceedings of Machine Learning Research},
  pages 2218--2227. PMLR, 2017.

\bibitem[Ma et~al.(2019)Ma, Li, and Hern{\'{a}}ndez{-}Lobato]{MaLH19}
C.~Ma, Y.~Li, and J.~M. Hern{\'{a}}ndez{-}Lobato.
\newblock {Variational Implicit Processes}.
\newblock In \emph{Proceedings of the 36th International Conference on Machine
  Learning}, volume~97 of \emph{Proceedings of Machine Learning Research},
  pages 4222--4233. {PMLR}, 2019.

\bibitem[MacKay(1992)]{mackay1992information}
D.~J. MacKay.
\newblock Information-based objective functions for active data selection.
\newblock \emph{Neural computation}, 4\penalty0 (4):\penalty0 590--604, 1992.

\bibitem[MacKay(1995)]{mackay1995probable}
D.~J. MacKay.
\newblock {Probable Networks and Plausible Predictions - a Review of Practical
  Bayesian Methods for Supervised Neural Networks}.
\newblock \emph{Network: {C}omputation in {N}eural {S}ystems}, 6\penalty0
  (3):\penalty0 469--505, 1995.

\bibitem[MacKay(1996)]{mackay1996bayesian}
D.~J. MacKay.
\newblock Bayesian non-linear modeling for the prediction competition.
\newblock In \emph{Maximum Entropy and Bayesian Methods}, pages 221--234.
  Springer, 1996.

\bibitem[Mackay(2003)]{MacKay03}
D.~J.~C. Mackay.
\newblock \emph{{Information Theory, Inference and Learning Algorithms}}.
\newblock Cambridge University Press, 1st edition, 2003.

\bibitem[Matsubara et~al.(2021)Matsubara, Oates, and Briol]{Matsubara2020}
T.~Matsubara, C.~J. Oates, and F.~Briol.
\newblock {The Ridgelet Prior: {A} Covariance Function Approach to Prior
  Specification for Bayesian Neural Networks}.
\newblock \emph{Journal of Machine Learning Research}, 22:\penalty0 1--57,
  2021.

\bibitem[Matthews et~al.(2018)Matthews, Hron, Rowland, Turner, and
  Ghahramani]{Matthews2018}
A.~Matthews, J.~Hron, M.~Rowland, R.~E. Turner, and Z.~Ghahramani.
\newblock {Gaussian Process Behaviour in Wide Deep Neural Networks}.
\newblock In \emph{International Conference on Learning Representations}, 2018.

\bibitem[Mo{\v{c}}kus(1975)]{marchuk1975}
J.~Mo{\v{c}}kus.
\newblock {On Bayesian Methods for Seeking the Extremum}.
\newblock In \emph{Optimization Techniques IFIP Technical Conference
  Novosibirsk}, pages 400--404. Springer Berlin Heidelberg, 1975.

\bibitem[M{\"u}ller(1997)]{muller1997integral}
A.~M{\"u}ller.
\newblock {Integral Probability Metrics and Their Generating Classes of
  Functions}.
\newblock \emph{Advances in Applied Probability}, 29\penalty0 (2):\penalty0
  429--443, 1997.

\bibitem[Nalisnick et~al.(2021)Nalisnick, Gordon, and
  Hern{\'{a}}ndez{-}Lobato]{nalisnick2020}
E.~T. Nalisnick, J.~Gordon, and J.~M. Hern{\'{a}}ndez{-}Lobato.
\newblock {Predictive Complexity Priors}.
\newblock In \emph{Proceedings of the 24th International Conference on
  Artificial Intelligence and Statistics}, volume 130 of \emph{Proceedings of
  Machine Learning Research}, pages 694--702. {PMLR}, 2021.

\bibitem[Neal(1996)]{neal1996bayesian}
R.~M. Neal.
\newblock \emph{{Bayesian Learning for Neural Networks (Lecture Notes in
  Statistics)}}.
\newblock Springer, 1st edition, Aug. 1996.

\bibitem[Nogueira(2014)]{fernando2014}
F.~Nogueira.
\newblock {Bayesian Optimization}: Open source constrained global optimization
  tool for {Python}, 2014.
\newblock URL \url{https://github.com/fmfn/BayesianOptimization}.

\bibitem[O'Hagan(1991)]{Hagan1991}
A.~O'Hagan.
\newblock {Bayes–Hermite quadrature}.
\newblock \emph{Journal of Statistical Planning and Inference}, 29\penalty0
  (3):\penalty0 245 -- 260, 1991.

\bibitem[Osawa et~al.(2019)Osawa, Swaroop, Khan, Jain, Eschenhagen, Turner, and
  Yokota]{Osawa2019}
K.~Osawa, S.~Swaroop, M.~E.~E. Khan, A.~Jain, R.~Eschenhagen, R.~E. Turner, and
  R.~Yokota.
\newblock {Practical Deep Learning with Bayesian Principles}.
\newblock In \emph{Advances in Neural Information Processing Systems},
  volume~32, pages 4287--4299. Curran Associates, Inc., 2019.

\bibitem[Ovadia et~al.(2019)Ovadia, Fertig, Ren, Nado, Sculley, Nowozin,
  Dillon, Lakshminarayanan, and Snoek]{ovadia2019can}
Y.~Ovadia, E.~Fertig, J.~Ren, Z.~Nado, D.~Sculley, S.~Nowozin, J.~Dillon,
  B.~Lakshminarayanan, and J.~Snoek.
\newblock {Can You Trust Your Model's Uncertainty? Evaluating Predictive
  Uncertainty Under Dataset Shift}.
\newblock In \emph{Advances in Neural Information Processing Systems},
  volume~32, pages 13991--14002. Curran Associates, Inc., 2019.

\bibitem[Paszke et~al.(2019)Paszke, Gross, Massa, Lerer, Bradbury, Chanan,
  Killeen, Lin, Gimelshein, Antiga, Desmaison, Kopf, Yang, DeVito, Raison,
  Tejani, Chilamkurthy, Steiner, Fang, Bai, and Chintala]{paszke2019pytorch}
A.~Paszke, S.~Gross, F.~Massa, A.~Lerer, J.~Bradbury, G.~Chanan, T.~Killeen,
  Z.~Lin, N.~Gimelshein, L.~Antiga, A.~Desmaison, A.~Kopf, E.~Yang, Z.~DeVito,
  M.~Raison, A.~Tejani, S.~Chilamkurthy, B.~Steiner, L.~Fang, J.~Bai, and
  S.~Chintala.
\newblock {PyTorch: An Imperative Style, High-Performance Deep Learning
  Library}.
\newblock In \emph{Advances in Neural Information Processing Systems},
  volume~32, pages 8026--8037. Curran Associates, Inc., 2019.

\bibitem[Pearce et~al.(2019)Pearce, Tsuchida, Zaki, Brintrup, and
  Neely]{Pearce2019}
T.~Pearce, R.~Tsuchida, M.~Zaki, A.~Brintrup, and A.~Neely.
\newblock {Expressive Priors in Bayesian Neural Networks: Kernel Combinations
  and Periodic Functions}.
\newblock In \emph{Proceedings of the 35th Conference on Uncertainty in
  Artificial Intelligence}, page~25. {AUAI} Press, 2019.

\bibitem[Phan et~al.(2019)Phan, Pradhan, and Jankowiak]{phan2019composable}
D.~Phan, N.~Pradhan, and M.~Jankowiak.
\newblock {Composable Effects for Flexible and Accelerated Probabilistic
  Programming in NumPyro}.
\newblock arXiv:1912.11554, 2019.

\bibitem[Rasmussen and Ghahramani(2002)]{Ghahramani2003}
C.~E. Rasmussen and Z.~Ghahramani.
\newblock {Bayesian Monte Carlo}.
\newblock In \emph{Advances in Neural Information Processing Systems},
  volume~15, pages 489--496. {MIT} Press, 2002.

\bibitem[Rasmussen and Williams(2006)]{Rasmussen06}
C.~E. Rasmussen and C.~Williams.
\newblock \emph{{Gaussian Processes for Machine Learning}}.
\newblock MIT Press, 2006.

\bibitem[Rezende and Mohamed(2015)]{Rezende2015}
D.~Rezende and S.~Mohamed.
\newblock {Variational Inference with Normalizing Flows}.
\newblock In \emph{Proceedings of the 32nd International Conference on Machine
  Learning}, volume~37 of \emph{Proceedings of Machine Learning Research},
  pages 1530--1538, Lille, France, 07--09 Jul 2015. PMLR.

\bibitem[Rezende et~al.(2014)Rezende, Mohamed, and
  Wierstra]{rezende2014stochastic}
D.~J. Rezende, S.~Mohamed, and D.~Wierstra.
\newblock {Stochastic Backpropagation and Approximate Inference in Deep
  Generative Models}.
\newblock In \emph{Proceedings of the 31th International Conference on Machine
  Learning}, volume~32 of \emph{Proceeding of Machine Learning Research}, pages
  1278--1286, Beijing, China, 21-26 June 2014. PMLR.

\bibitem[Rossi et~al.(2019)Rossi, Michiardi, and Filippone]{Rossi2018}
S.~Rossi, P.~Michiardi, and M.~Filippone.
\newblock {Good Initializations of Variational {B}ayes for Deep Models}.
\newblock In \emph{Proceedings of the 36th International Conference on Machine
  Learning}, volume~97 of \emph{Proceedings of Machine Learning Research},
  pages 5487--5497, Long Beach, California, USA, 09--15 Jun 2019. PMLR.

\bibitem[Rossi et~al.(2020)Rossi, Marmin, and Filippone]{Rossi2020}
S.~Rossi, S.~Marmin, and M.~Filippone.
\newblock {Walsh-Hadamard Variational Inference for Bayesian Deep Learning}.
\newblock In \emph{Advances in Neural Information Processing Systems},
  volume~33, 2020.

\bibitem[Settles(2009)]{settles2009active}
B.~Settles.
\newblock {Active Learning Literature Survey}.
\newblock Technical report, University of Wisconsin-Madison Department of
  Computer Sciences, 2009.

\bibitem[Shi et~al.(2018)Shi, Sun, and Zhu]{ShiS018}
J.~Shi, S.~Sun, and J.~Zhu.
\newblock {A Spectral Approach to Gradient Estimation for Implicit
  Distributions}.
\newblock In \emph{Proceedings of the 35th International Conference on Machine
  Learning}, volume~80 of \emph{Proceedings of Machine Learning Research},
  pages 4651--4660. {PMLR}, 2018.

\bibitem[Shi et~al.(2019)Shi, Khan, and Zhu]{ShiK019}
J.~Shi, M.~E. Khan, and J.~Zhu.
\newblock {Scalable Training of Inference Networks for Gaussian-Process
  Models}.
\newblock In \emph{Proceedings of the 36th International Conference on Machine
  Learning}, volume~97 of \emph{Proceedings of Machine Learning Research},
  pages 5758--5768. {PMLR}, 2019.

\bibitem[Simonyan and Zisserman(2015)]{simonyan2014very}
K.~Simonyan and A.~Zisserman.
\newblock {Very Deep Convolutional Networks for Large-Scale Image Recognition}.
\newblock In \emph{International Conference on Learning Representations}, 2015.

\bibitem[Skafte et~al.(2019)Skafte, Jorgensen, and Hauberg]{skafte2019reliable}
N.~Skafte, M.~Jorgensen, and S.~Hauberg.
\newblock {Reliable Training and Estimation of Variance Networks}.
\newblock In \emph{Advances in Neural Information Processing Systems},
  volume~32, pages 6326--6336. Curran Associates, Inc., 2019.

\bibitem[Snoek et~al.(2012)Snoek, Larochelle, and Adams]{Snoek2012}
J.~Snoek, H.~Larochelle, and R.~P. Adams.
\newblock {Practical Bayesian Optimization of Machine Learning Algorithms}.
\newblock In \emph{Advances in Neural Information Processing Systems},
  volume~25. Curran Associates, Inc., 2012.

\bibitem[Springenberg et~al.(2016)Springenberg, Klein, Falkner, and
  Hutter]{Springenberg2016}
J.~T. Springenberg, A.~Klein, S.~Falkner, and F.~Hutter.
\newblock {Bayesian Optimization with Robust Bayesian Neural Networks}.
\newblock In \emph{Advances in Neural Information Processing Systems},
  volume~29, pages 4134--4142. Curran Associates, Inc., 2016.

\bibitem[Srinivas et~al.(2010)Srinivas, Krause, Kakade, and
  Seeger]{Srinivas2010}
N.~Srinivas, A.~Krause, S.~M. Kakade, and M.~W. Seeger.
\newblock {Gaussian Process Optimization in the Bandit Setting: No Regret and
  Experimental Design}.
\newblock In \emph{Proceedings of the 27th International Conference on Machine
  Learning}, pages 1015--1022. Omnipress, 2010.

\bibitem[Sun et~al.(2019)Sun, Zhang, Shi, and Grosse]{Sun2019}
S.~Sun, G.~Zhang, J.~Shi, and R.~Grosse.
\newblock {Functional Variational Bayesian Neural Networks}.
\newblock In \emph{International Conference on Learning Representations}, 2019.

\bibitem[Tieleman and Hinton(2012)]{Tieleman2012}
T.~Tieleman and G.~Hinton.
\newblock {Lecture 6.5---RmsProp: Divide the Gradient by a Running Average of
  Its Recent Magnitude}.
\newblock COURSERA: Neural Networks for Machine Learning, 2012.

\bibitem[Tishby et~al.(1989)Tishby, Levin, and Solla]{Tishby1989}
Tishby, Levin, and Solla.
\newblock Consistent inference of probabilities in layered networks:
  predictions and generalizations.
\newblock In \emph{International 1989 Joint Conference on Neural Networks},
  pages 403--409 vol.2, 1989.

\bibitem[Tran et~al.(2021)Tran, Rossi, Milios, Michiardi, Bonilla, and
  Filippone]{Tran2021}
B.-H. Tran, S.~Rossi, D.~Milios, P.~Michiardi, E.~V. Bonilla, and M.~Filippone.
\newblock {Model Selection for Bayesian Autoencoders}.
\newblock In \emph{Advances in Neural Information Processing Systems},
  volume~34, pages 19730--19742. Curran Associates, Inc., 2021.

\bibitem[Villani(2003)]{Villani2003}
C.~Villani.
\newblock \emph{{Topics in Optimal Transportation}}.
\newblock Graduate studies in mathematics. American Mathematical Society, 2003.

\bibitem[Wenzel et~al.(2020)Wenzel, Roth, Veeling, {\'{S}}wi{\c{a}}tkowski,
  Tran, Mandt, Snoek, Salimans, Jenatton, and Nowozin]{Wenzel2020}
F.~Wenzel, K.~Roth, B.~S. Veeling, J.~{\'{S}}wi{\c{a}}tkowski, L.~Tran,
  S.~Mandt, J.~Snoek, T.~Salimans, R.~Jenatton, and S.~Nowozin.
\newblock {How Good is the Bayes Posterior in Deep Neural Networks Really?}
\newblock In \emph{Proceeding of the 37th International Conference on Machine
  Learning}, 2020.

\bibitem[Yang et~al.(2019)Yang, Lorch, Graule, Srinivasan, Suresh, Yao,
  Pradier, and Doshi-velez]{Yang2019}
W.~Yang, L.~Lorch, M.~A. Graule, S.~Srinivasan, A.~Suresh, J.~Yao, M.~F.
  Pradier, and F.~Doshi-velez.
\newblock {Output-Constrained Bayesian Neural Networks}.
\newblock In \emph{ICML workshop on Uncertainty {\&} Robustness in Deep
  Learning}, 2019.

\bibitem[Yao et~al.(2007)Yao, Rosasco, and Caponnetto]{rosasco2007early}
Y.~Yao, L.~Rosasco, and A.~Caponnetto.
\newblock {On Early Stopping in Gradient Descent Learning}.
\newblock \emph{Constructive Approximation}, 26\penalty0 (2):\penalty0
  289--315, 2007.

\bibitem[Zhang et~al.(2020)Zhang, Li, Zhang, Chen, and
  Wilson]{zhang2020csgmcmc}
R.~Zhang, C.~Li, J.~Zhang, C.~Chen, and A.~G. Wilson.
\newblock {Cyclical Stochastic Gradient MCMC for Bayesian Deep Learning}.
\newblock In \emph{International Conference on Learning Representations}, 2020.

\end{thebibliography}
\end{document}